\documentclass[runningheads]{llncs}

\makeatletter
\def\thanks#1{\protected@xdef\@thanks{\@thanks
        \protect\footnotetext{#1}}}
\makeatother

\usepackage{eccv}

\usepackage{amsmath}
\usepackage{amssymb}
\usepackage{pifont}
\usepackage{booktabs}
\usepackage{makecell}
\usepackage{multirow}
\usepackage{multicol}
\usepackage{colortbl}
\usepackage{wrapfig}

\newcommand{\tabincell}[2]{\begin{tabular}{@{}#1@{}}#2\end{tabular}}

\newcommand{\PreserveBackslash}[1]{\let\temp=\\#1\let\\=\temp}
\newcolumntype{C}[1]{>{\PreserveBackslash\centering}p{#1}}
\newcolumntype{R}[1]{>{\PreserveBackslash\raggedleft}p{#1}}
\newcolumntype{L}[1]{>{\PreserveBackslash\raggedright}p{#1}}

\usepackage{eccvabbrv}

\usepackage{graphicx}
\usepackage{booktabs}

\usepackage[accsupp]{axessibility}  

\usepackage[pagebackref,breaklinks,colorlinks,citecolor=eccvblue]{hyperref}

\usepackage{orcidlink}

\begin{document}

\title{\emph{VastTrack}: Vast Category Visual Object Tracking} 

\titlerunning{\emph{VastTrack}: Vast Category Visual Object Tracking}

\author{Liang Peng\inst{1*} \and
Junyuan Gao\inst{1*} \and
Xinran Liu\inst{1*} \and
Weihong Li\inst{1*} \and
Shaohua Dong\inst{2*} \and
Zhipeng Zhang\inst{3} \and
Heng Fan\inst{2\dag} \and
Libo Zhang\inst{1\dag\sharp}
\thanks{$^{*}$Equal contributions and co-first authors\; $^{\dag}$Equal advising \; $^{\sharp}$Corr. author}
}

\authorrunning{L.~Peng et al.}

\institute{
$^{1\;\;}$Institute of Software Chinese Academy of Sciences \; \\
$^{2\;\;}$Department of CSE, University of North Texas \;
$^{3\;\;}$KargoBot \\
\email{E-mail: hengfan@unt.edu; libo@iscas.ac.cn} 
\\
\href{https://github.com/HengLan/VastTrack}{\email{https://github.com/HengLan/VastTrack}}
}

\maketitle

\begin{abstract}
  In this paper, we introduce a novel benchmark, dubbed \textbf{\emph{VastTrack}}, towards facilitating the development of more general visual tracking via encompassing abundant classes and videos. VastTrack possesses several attractive properties: \textbf{(1) Vast Object Category.} In particular, it covers target objects from 2,115 classes, largely surpassing object categories of existing popular benchmarks (\eg, GOT-10k with 563 classes and LaSOT with 70 categories). With such vast object classes, we expect to learn more general object tracking. \textbf{(2) Larger scale.} Compared with current benchmarks, VastTrack offers 50,610 sequences with 4.2 million frames, which makes it to date the largest benchmark regarding the number of videos, and thus could benefit training even more powerful visual trackers in the deep learning era. \textbf{(3) Rich Annotation.} Besides conventional bounding box annotations, VastTrack also provides linguistic descriptions for the videos. The rich annotations of VastTrack enables development of both the vision-only and the vision-language tracking. To ensure precise annotation, all videos are manually labeled with multiple rounds of careful inspection and refinement. To understand performance of existing trackers and to provide baselines for future comparison, we extensively assess 25 representative trackers. The results, not surprisingly, show significant drops compared to those on current datasets due to lack of abundant categories and videos from diverse scenarios for training, and more efforts are required to improve general tracking. Our VastTrack and all the evaluation results will be made publicly available \href{https://github.com/HengLan/VastTrack}{\email{here}}.
  
\end{abstract}

\section{Introduction}
\label{sec:intro}

Visual tracking is one of the most fundamental problems in computer vision, and has many important applications such as video surveillance, robotics, intelligent vehicle, human-machine interaction, and augmented reality. The ultimate goal for visual tacking is to localize the target of an \emph{arbitrary} category in an \emph{arbitrary} scenario from a sequence, given its initial position (\eg, usually a bounding box) from the first image frame, which we term \emph{universal visual tracking}. For such a goal, numerous tracking algorithms have been proposed in recent decades~\cite{yilmaz2006object,smeulders2013visual,li2018deep,javed2022visual,marvasti2021deep}. In particular, with the introduction of several large-scale tracking benchmarks (\eg,~\cite{muller2018trackingnet,fan2019lasot,huang2019got}) in deep learning era, considerable advancements (\eg,~\cite{chen2021transformer,cui2022mixformer,ye2022joint,chen2022backbone,lin2022swintrack,chen2023seqtrack,wei2023autoregressive}) have been seen in the visual tracking community. Despite this, it remains challenging to achieve universal tracking.

\setlength{\columnsep}{10pt}
\begin{wrapfigure}{r}{0.5\textwidth}
\vspace{-6mm}
\centering
\includegraphics[width=0.5\textwidth]{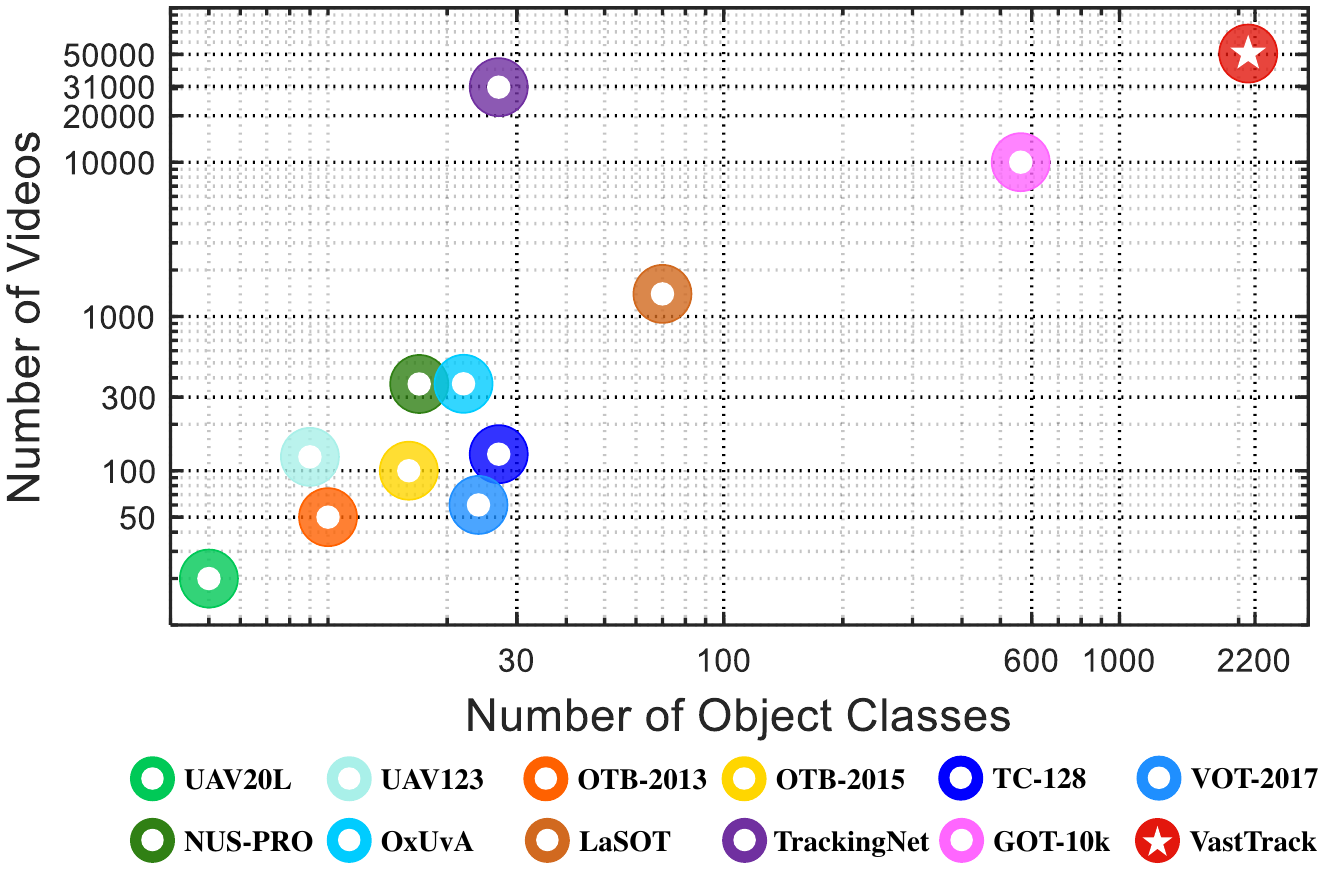}
\caption{Summary of representative benchmarks, comprising OTB-2013/2015~\cite{wu2013online,wu2015object}, TC-128~\cite{liang2015encoding}, UAV123~\cite{mueller2016benchmark}, NUS-PRO~\cite{li2016nus}, UAV20L~\cite{mueller2016benchmark}, VOT-2017~\cite{kristan2016novel}, OxUvA~\cite{valmadre2018long}, GOT-10k~\cite{huang2019got}, TrackingNet~\cite{muller2018trackingnet}, and VastTrack. We can clearly see that VastTrack is \emph{larger} than all other datasets by containing 2,115 object categories and 50,803 videos. \emph{Best viewed in color for all figures in paper.}
}
\vspace{-15pt}
\label{fig:1}
\end{wrapfigure}
One important reason is relatively \emph{restricted} number of object categories in current tracking datasets. The objects in the real world are from countless categories. To achieve general visual tracking like humans, the tracker is desired to ``see'' various  sequences from an extremely large set of object categories during training to acquire the ability of generalization. However, the categories in existing large-scale datasets are rather \emph{limited}. For example, the popular TrackingNet~\cite{muller2018trackingnet} and LaSOT~\cite{fan2019lasot} comprise respectively 27 and 70 categories (see Fig.~\ref{fig:1}), which fall short for training universally generalizable trackers. Another popular dataset GOT-10k~\cite{huang2019got} aims to handle this by largely expanding the number of object categories to 563. Despite its success in advancing generic-purpose tracking, the 563 object categories are still insufficient to represent massive diversity of categories present in the real world. Besides training, a real general tracking system requires evaluation on videos of vast object categories, which can help mitigate biases to certain classes for more faithful assessment in real applications. However, the test sets of existing large-scale benchmarks (\eg,~\cite{muller2018trackingnet,fan2019lasot,huang2019got}) all consist of \emph{less than} 100 object categories, which may not be enough for faithful assessment of general tracking.

\begin{figure}[!t]
	\centering
\includegraphics[width=0.96\linewidth]{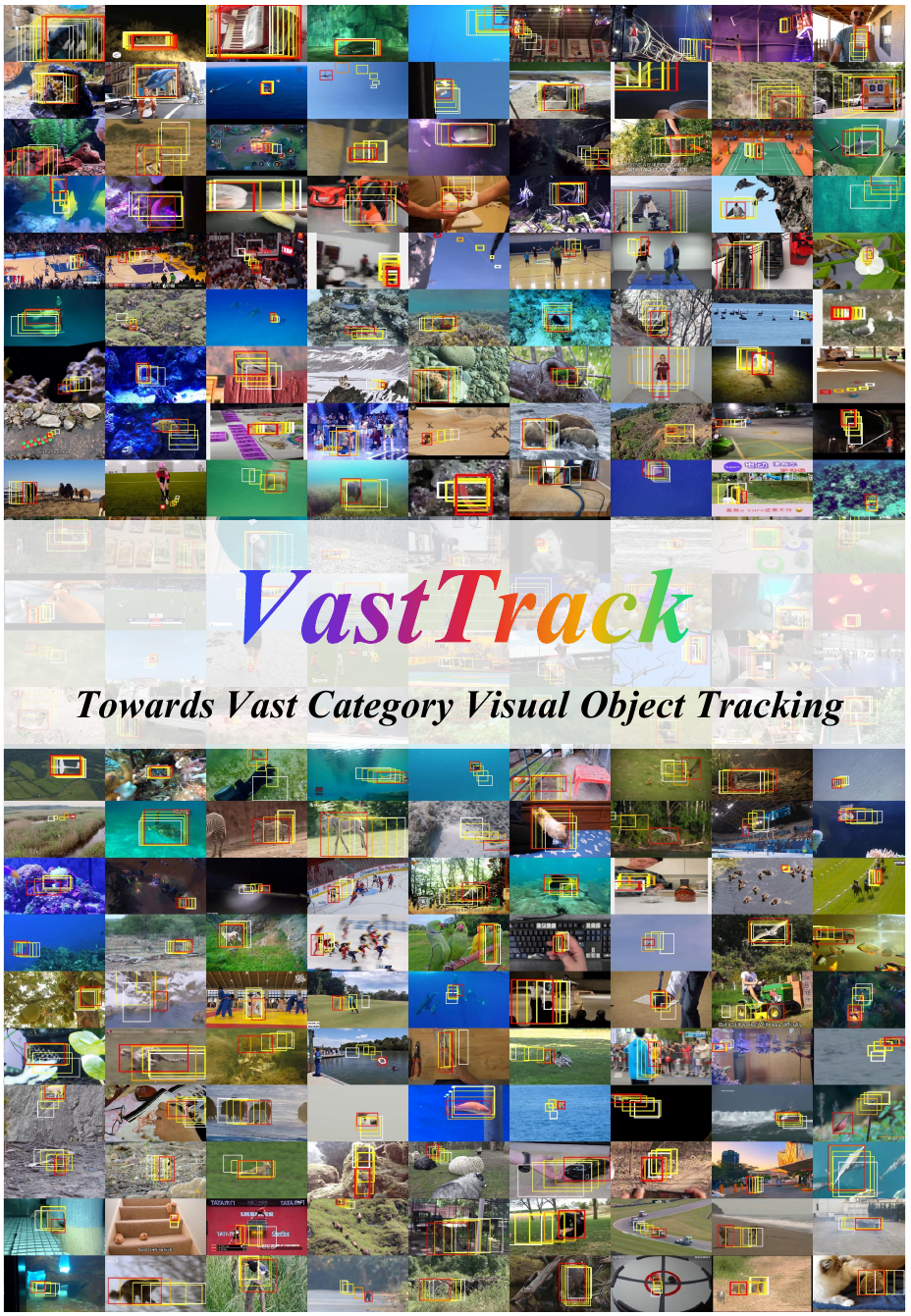}
	\caption{\textbf{VastTrack}, a new large-scale benchmark for facilitating general single object tracking with abundant object categories and videos. Here we show the partial target trajectory in a video. \emph{Note, only a very small part of categories and videos are displayed.}}
	\label{fig:bigimg}
\end{figure}

In addition to rich object categories, abundant sequences with high-quality annotations are crucial for learning robust general visual tracking. Particularly, as the tracking model becomes larger and more complicated, \eg, from convolutional neural networks (CNNs)~\cite{krizhevsky2012imagenet,he2016deep} to Transformer model~\cite{vaswani2017attention,dosovitskiy2020image}, more and more training sequences are required to unleash the power of deep network for achieving robustness and generality. While there have been extensive efforts to develop tracking benchmarks, they are comparatively small in scale or limited in annotation quality. For example, currently the largest (in term of video number) tracking benchmark~\cite{huang2019got} with \emph{precise} annotations has only 10K video sequences, which may still be inadequate for training generalized trackers, as evidenced by enhanced performance~\cite{cui2022mixformer} on it when employing extra training videos. Although another benchmark~\cite{muller2018trackingnet} offers more training data of around 30K sequences, its annotations are not precise, which may degrade the tracking performance. 

More recently, natural language has demonstrated great potential for enhancing the robustness of general tracking, and the resulted paradigm, the so-called vision-language tracking (\eg,~\cite{li2017tracking,guo2022divert,feng2021siamese,zhou2023joint}), has drawn increasing attention in tracking community. To learn a robust and general vision-language tracker, it is crucial to provide ample videos with visual and linguistic annotations. Although there exist a few datasets (\eg,~\cite{fan2019lasot,wang2021towards}) developed for this goal, the number of linguistic sentences are limited in scale (\eg, 1.4K in \cite{fan2019lasot} and 2K in~\cite{wang2021towards}), which may impede the exploration of more general vision-language tracking.

In order to alleviate the aforementioned limitations in existing datasets for developing more general visual tracking, we present \textbf{\emph{VastTrack}} (Fig.~\ref{fig:bigimg}), a novel large-scale dataset towards \textbf{Vast} visual object \textbf{Track}ing by comprising abundant categories and video from diverse scenarios in the wild. Particularly, our VastTrack makes the following efforts for developing general object tracking:

\textbf{(1) \emph{Vast Object Category}:} To enrich the diversity in object categories for general tracking, VastTrack consists of video sequences from 2,115 classes, which largely surpasses the number of categories in the popular benchmarks such as GOT-10k~\cite{huang2019got} with 563 classes and LaSOT~\cite{fan2019lasot} with 70 classes, as can be seen in Fig.~\ref{fig:1}. To the best of our knowledge, VastTrack is the richest tracking benchmark with the largest number of object categories. With such vast object classes, we expect to accelerate the exploration towards more general tracking.

\textbf{(2) \emph{Larger Scale}:} For learning a robust universal tracking algorithm, our VastTrack offers 50,610 video sequences with 4.2 million frames, which makes it so far the largest and the most diverse tracking dataset in terms of the numbers of video sequences and targets compared to existing benchmarks (\eg~\cite{fan2019lasot,huang2019got,muller2018trackingnet}), as displayed in Fig.~\ref{fig:1}. Such larger scale and diversity of the proposed VastTrack in videos and targets can potentially benefit training more powerful trackers, particularly Transformer-based models, in the deep learning era.

\textbf{(3) \emph{Rich and Precise Annotations}:} Considering the benefits of language for enhancing general object tracking, VastTrack offers both standard bounding box annotations and rich linguistic specifications for the sequences, and thus enables exploration of both the vision-only and vision-language universal tracking. Compared with current benchmarks (\eg,~\cite{fan2019lasot} with 1.4K sentences and \cite{wang2021towards} with 2K texts) for vision-language tracking, the proposed VastTrack provides over 50K descriptions, a magnitude order larger than~\cite{fan2019lasot,wang2021towards}, of more and diverse targets for better vision-language tracking. In addition, to ensure precise annotations, each video in VastTrack is manually labeled with multi-round refinements. 

By releasing VastTrack, we aim at providing a new large-scale platform with abundant video sequences from vast object categories for facilitating the development of more general and universal tracking and its applications.

In order to understand the performance of existing tracking algorithms on VastTrack and to provide baseline for future comparison, we extensively evaluate 25 recent representative visual trackers in a hybrid protocol in which the test videos have partial overlap with the training sequences (as described later) and conduct in-depth analysis. The evaluation demonstrates that, not surprisingly, the top-performing object trackers significantly degenerate on more challenging VastTrack. For example, the success scores of existing state-of-the-art trackers, \eg, SeqTrack~\cite{chen2023seqtrack}, MixFormer~\cite{cui2022mixformer}, and  OSTrack~\cite{ye2022joint}, degrade from 0.725, 0.724, and 0.711 on LaSOT~\cite{fan2019lasot} to 0.396 (with a drop of 0.329), 0.395 (with a drop of 0.329), and 0.336 (with a drop of 0.375) on VastTrack. This exhibits the challenge in achieving universal tracking for current visual trackers, and more efforts are thus desired to improve general-purpose object tracking.

In summary, our main \textbf{contributions} are as follows: \ding{171} We introduce a new benchmark VastTrack that covers 2,115 object categories to facilitate more general object tracking; \ding{170} VastTrack provides a large scale of 50,610 videos which could benefit developing more powerful deep trackers; \ding{168} Rich annotations of bounding boxes and language in VastTrack enable the exploration of both vision-only and vision-language tracking; \ding{169} Extensive evaluation of 25 trackers are conducted to understand VastTrack and provide baselines for future comparison.

\section{Related Work}
\label{related}
In this section, we will briefly discuss works that are closely relevant to ours from the following three directions, including visual tracking datasets, visual tracking algorithms, and other vision benchmarks with vast categories.

\subsection{Visual Tracking Benchmarks}

Benchmarks have been crucial for development of tracking. Early tracking benchmarks are usually in small scale and mainly aim at the evaluation purpose for fairly comparing different algorithms. OTB-2013~\cite{wu2013online} is the first object tracking benchmark by introducing 51 videos and later extend to OTB-2015~\cite{wu2015object} by introducing extra videos. VOT~\cite{kristan2016novel} presents a series of challenges to compare trackers in different aspects. TC-128~\cite{liang2015encoding} consists of 128 colorful videos with the goal of studying the impact of color information on tracking models. NfS~\cite{galoogahi2017need} assesses tracking performance by providing 100 sequences with high frame rate. UAV123 and UAV20L~\cite{mueller2016benchmark} respectively consist of 123 and 20 video sequences captured by unmanned aerial vehicle for tracking performance evaluation. NUS-PRO~\cite{li2016nus} offers 365 videos to assess trackers on rigid target objects. OxUvA~\cite{valmadre2018long} contains 366 video sequences for evaluating the long-term tracking performance of different algorithms. From a different perspective than opaque tracking benchmarks, TOTB~\cite{fan2021transparent} collects 225 videos for investigating transparent object tracking. 

Despite facilitating tracking, early datasets are limited in scale and cannot provide training videos for deep tracking. To alleviate this, several large-scale benchmarks have been introduced in recent years. TrackingNet~\cite{muller2018trackingnet} presents a large-scale datasets with around 30K videos for training deep tracking. However, its annotations are generated using a tracker, which may be inaccurate and thus degrade the training of deep tracking. LaSOT~\cite{fan2019lasot} comprises 1,400 long-term videos with precise dense annotations, and is later extended in~\cite{fan2021lasot} by adding more video sequences. Notably, it provides both the bounding box and language annotation to enable both vision-only and vision-language tracking. GOT-10~\cite{huang2019got} contributes a large benchmark with around 10K video sequences from 563 classes. Despite advancing the deep tracking, 563 object classes may still be insufficient to represent the massive categories in the real world.

Our VastTrack is related to the aforementioned large-scale datasets but provides a more diverse and larger platform with more than 50K videos from 2,115 categories, which aims to accelerate the exploration towards universal and general tracking. Tab.~\ref{tab:l}  shows the comparison of VastTrack with other datasets.

\renewcommand\arraystretch{1.11}
\begin{table*}[t]\scriptsize
	\centering
	\caption{Comparison of VastTrack with representative generic tracking benchmarks. ``Tra.'' and ``Eav.'' indicate training and evaluation, respectively. ${\textbf{\dag}}$: TrackingNet is semi-automatically, instead of manually,  annotated using a correlation filter tracker.}
 \scalebox{0.89}{
	\begin{tabular}{rcccccccccc}
		\Xhline{2\arrayrulewidth}
		\textbf{Benchmark} & \multicolumn{1}{c}{\textbf{Year}} & \multicolumn{1}{c}{\textbf{Classes}} & \multicolumn{1}{c}{\textbf{Videos}} & \multicolumn{1}{l}{\textbf{\tabincell{c}{Mean \\Frames}}} & \textbf{\tabincell{c}{Total\\ Frames}} & \textbf{\tabincell{c}{Total\\ Duration}} & \multicolumn{1}{c}{\textbf{\tabincell{c}{Absent \\label}}} & \multicolumn{1}{c}{\textbf{\tabincell{c}{Num. of \\Att.}}} & \multicolumn{1}{c}{\textbf{\tabincell{c}{Lang. \\ Anno.}}}  & \textbf{\tabincell{c}{Dataset \\Goal}} \\
		\hline\hline
		OTB-2013~\cite{wu2013online} & 2013  & 10    & 50    & 578   & 29$\textbf{K}$   & 16.4 $\textbf{min}$ & \ding{55} & 11      & \ding{55} & Eva. \\
		OTB-2015~\cite{wu2015object} & 2015  & 16    & 100   & 590   & 59$\textbf{K}$   & 32.8 $\textbf{min}$ & \ding{55} & 11      & \ding{55} & Eva. \\
		TC-128~\cite{liang2015encoding} & 2015  & 27    & 128   & 429   & 55$\textbf{K}$   & 30.7 $\textbf{min}$ & \ding{55} & 11      & \ding{55} & Eva. \\
		NUS-PRO~\cite{li2016nus} & 2016  & 17    & 365   & 371   & 135$\textbf{K}$  & 75.2 $\textbf{min}$ & \ding{55} & 12      & \ding{55} & Eva. \\
		UAV123~\cite{mueller2016benchmark} & 2016  & 9     & 123   & 915   & 113$\textbf{K}$  & 62.5 $\textbf{min}$ & \ding{55} & 12     & \ding{55} & Eva. \\
		UAV20L~\cite{mueller2016benchmark} & 2016  & 5     & 20    & 2,934 & 59$\textbf{K}$   & 32.6 $\textbf{min}$ & \ding{55} & 12     & \ding{55} & Eva. \\
		NfS~\cite{galoogahi2017need}   & 2017  & 17    & 100   & 3,830 & 383$\textbf{K}$  & 26.6 $\textbf{min}$ & \ding{55} & 9       & \ding{55} & Eva. \\
		VOT-2017~\cite{kristan2016novel} & 2017  & 24    & 60    & 356   & 21$\textbf{K}$   & 11.9 $\textbf{min}$ & \ding{55} & 24      & \ding{55} & Eva. \\
		OxUvA~\cite{valmadre2018long} & 2018  & 22    & 366   & 4235  & 1.55$\textbf{M}$ & 14.4 $\textbf{hours}$ & \ding{55} & 6       & \ding{55} & Eva. \\
		TrackingNet~\cite{muller2018trackingnet} & 2018  & 27    & 30,643 & 471   & 14.43$\textbf{M}^{\textbf{\dag}}$ & 140.0 $\textbf{hours}$ & \ding{55} & 15      & \ding{55} & Tra./Eva. \\
		LaSOT~\cite{fan2019lasot} & 2019  & 70    & 1,400  & 2,053 & 3.52$\textbf{M}$ & 32.5 $\textbf{hours}$ & \ding{51}  & 14      & \ding{51}  & Tra./Eva. \\
		GOT-10k~\cite{huang2019got} & 2021  & 563   & 9,935 & 149   & 1.45$\textbf{M}$ & 40.0 $\textbf{hours}$ & \ding{51}  & 6       & \ding{55} & Tra./Eva. \\
		TNL2K~\cite{wang2021towards} & 2021  & -   & 2,000  & 622   & 1.24$\textbf{M}$ & 11.5 $\textbf{hours}$ & \ding{51}  & 17    & \ding{51}   & Tra./Eva. \\
		\hline
		\rowcolor{gray!15} \textbf{VastTrack} & 2023  & 2,115  & 50,610 & 83    & 4.20$\textbf{M}$ & 194.4 $\textbf{hours}$ & \ding{51}  & 10      & \ding{51}  & Tra./Eva. \\
		\Xhline{2\arrayrulewidth}
	\end{tabular}}%
	\label{tab:l}%
\end{table*}%

\subsection{Visual Tracking Algorithms}

Visual object tracking has witnessed considerable progress in recent years by embracing deep neural network. In particular, the Siamese tracking framework~\cite{bertinetto2016fully}, owing to its balanced accuracy and running efficiency, has attracted extensive attention in the tracking community with numerous extensions for improvements (\eg,~\cite{li2018high,zhu2018distractor,li2019siamrpn++,fan2019siamese,zhang2019deeper,wang2019spm,guo2020siamcar,fan2021cract,fu2021stmtrack,zhang2021learn}). More recently, Transformer model~\cite{vaswani2017attention,dosovitskiy2020image} has been introduced for tracking because of excellent ability in context modeling in images. The seminal methods of~\cite{chen2021transformer,wang2021transformer} propose to integrate Transformer into convolution neural network (CNN) architecture, displaying promising improvement. The approach of~\cite{guo2022learning} proposes inserting the attention in Transformer into multiple stages to enhance template and search region interaction and effectively enhances performance. The method of~\cite{yan2021learning} introduces a spatio-temporal Transformer network to improve tracking. In order to further improve the Transformer tracking, pure vision Transformer architecture has been employed to replace the CNN network for tracking. The algorithm of~\cite{lin2022swintrack} proposes to adopt the powerful Swin Transformer~\cite{liu2021swin} and achieves promising performance. The work of~\cite{ye2022joint} introduces a simple yet effective one-stream architecture for Transformer tracking. The work of~\cite{cui2022mixformer} presents a mix-attention module to achieve Transformer tracking, and shows excellent results. Besides these works, there are more Transformer trackers proposed recently (\eg~\cite{chen2023seqtrack,xie2023videotrack,gao2023generalized,wei2023autoregressive,zhao2023representation,cai2023robust}).  

\subsection{Other Vision Benchmarks with Vast Categories}

Benchmarks with vast object categories are desired in learning general vision systems. Many such benchmarks have been introduced for various vision tasks. For example, the well-known ImageNet~\cite{deng2009imagenet} contains 1,000 classes for image recognition. Open Image~\cite{kuznetsova2020open} covers 600 categories for object detection. LVIS~\cite{gupta2019lvis} comprises 1,203 classes for the tasks of object detection and instance segmentation. TAO~\cite{dave2020tao} contains 833 categories for general multi-object tracking. The recently proposed V3Det~\cite{wang2023v3det} contributes a new dataset with 13,204 object classes with the goal of facilitating the general detection system development.

In the similar spirit with the aforementioned vast category benchmarks, we introduce the novel VastTrack that contains 2,115 object classes and more than 50K video sequences for visual tracking. To the best of our knowledge, VastTrack is by far the largest tracking benchmark regarding categories and videos, and we hope it can serve as a cornerstone dataset for developing more general tracking systems and foster its various applications in the real world.

\section{The Proposed VastTrack}
\label{unisot}

\subsection{Construction Principle}

The goal of VastTrack to offer a unique platform with abundant object categories and video sequences with rich as well as precision annotations for facilitating the development of more general tracking. For this purpose, we follow principles below in constructing our VastTrack:

\vspace{-0.3em}
\begin{itemize}
	\setlength{\itemsep}{1.5pt}
	\setlength{\parsep}{1.5pt}
	\setlength{\parskip}{1.5pt}
	
	\item \emph{Vast Object Category.} One key motivation of VastTrack is to facilitate more universal object tracking with a rich class diversity. To this end, we hope that the new benchmark covers at least 2,000 object classes, containing common target objects suitable for visual tracking in our life.
	
	\item \emph{Larger Scale.}  Abundant sequences are crucial for training deep trackers. We expect VastTrack to include at least 50K videos with an average video length of at least 80 frames. Such a scale, greatly larger than current benchmarks, can potentially benefit training more powerful deep trackers.
	
	\item \emph{Rich Annotation.} One of important goals of VastTrack is to design a comprehensive platform that supports both vision-only and vision-language tracking. Considering this, both bounding boxes and language specifications will be provided to boost tracking in different directions.
	
	\item \emph{High Quality.} The quality of annotation is very important for both the training and evaluation of trackers. To ensure high quality of VastTrack, we manually label each video with multiple rounds of inspections and refinements.
	
\end{itemize}

\subsection{Data Acquisition}


Our benchmark aims to cover abundant object categories for tracking. For this purpose, 2,115 categories are selected for building VastTrack. These object categories are chosen from different sources, including classes in ImageNet~\cite{deng2009imagenet} and V3Det~\cite{wang2023v3det}, WordNet~\cite{miller1995wordnet}, and Wikipedia, and organized in a hierarchical tree structure. It is worth noticing that, each selected category is verified by an expert (\eg, a PhD or MS student working on the related topic) to ensure that it is suitable for the tracking task. Compared with existing datasets, the object classes of VastTrack are more diverse and more desired for universal tracking as discussed before. The details of object categories in VastTrack can be found in the \textbf{supplementary material} because of limited space.

\setlength{\columnsep}{10pt}%
\begin{wrapfigure}{r}{0.5\textwidth}
\centering
\vspace{-5mm}
\includegraphics[width=0.5\textwidth]{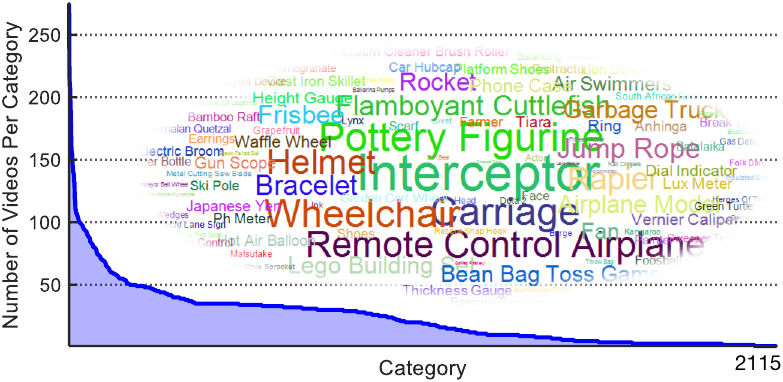}
\caption{The number of videos in each object class forms a long-tail distribution, which is common and universal in our real world.}
\vspace{-10pt}
\label{fig:2}
\end{wrapfigure}
After determining all object classes of VastTrack, we then search for the sequences of each class from YouTube. The reason to use YouTube for sourcing videos is because it is currently the largest the video platform and many videos come from the real world. Initially, we gather over 66K sequences. Then, we carefully inspect each video for the availability for tracking, and finally pick out 50,610 sequences. For each qualified video, we remove the irrelevant content from it, and only retain an usable clip for tracking. Note, unlike LaSOT~\cite{fan2019lasot} in which each category has the same number of videos, the sequence number of each class is not equal, forming a long-tail distribution (see Fig.~\ref{fig:2}) that is more universal in real world and could encourage learning more practical and general visual trackers~\cite{huang2019got}. 

Eventually, we create a new large-scale tracking dataset, VastTrack, by covering 2,115 categories. It consists of 50,610 video sequences with 4.2 million frames, and has an average video length of 83 frames. Due to space limitation, we show the detailed distribution of sequence length of VastTrack in the \textbf{supplementary material}. Please \textbf{note}, VastTrack is focused on short-term tracking by offering abundant object classes and sequences. Despite this, it could still be utilized for training long-term temporal trackers, as evidenced by the effectiveness of short-term videos in~\cite{huang2019got,muller2018trackingnet} for learning robust trackers on both long-/short scenarios. In order words, diversity and quantity of objects and videos may be more crucial and beneficial for deep tracker training.

\subsection{Annotation}

We follow the similar principle as in~\cite{fan2019lasot,fan2021lasot} for the bounding box annotation of a video: given the initial target object, in each frame, if the object shows up in the view, a labeler manually draws its (axis-aligned) bounding box as the tightest one to fit any visible part of the target; otherwise an absence label, either \emph{out-of-view} or \emph{full occlusion}, will be given to the frame. Note that, for some categories such as ``\emph{Kite}'' and ``\emph{Yo-Yo}'', the string does not belong to the target object to track, and thus will not be included in the annotated bounding box.

\begin{figure}[!t]
	\centering
		\includegraphics[width=0.19\textwidth]{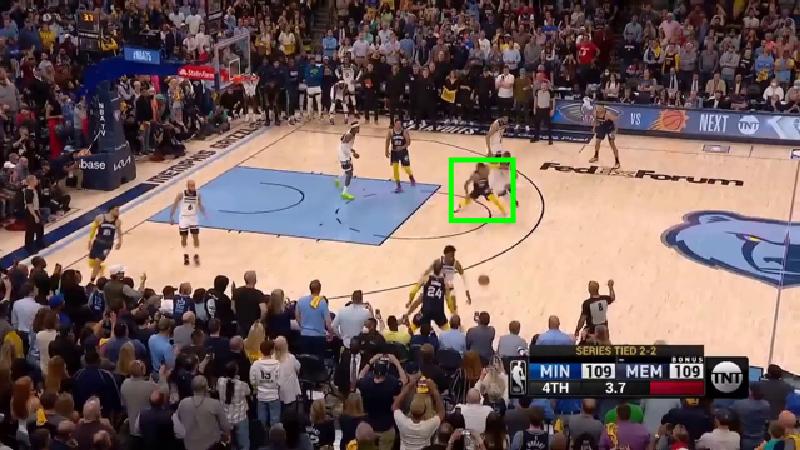}
        \includegraphics[width=0.19\textwidth]{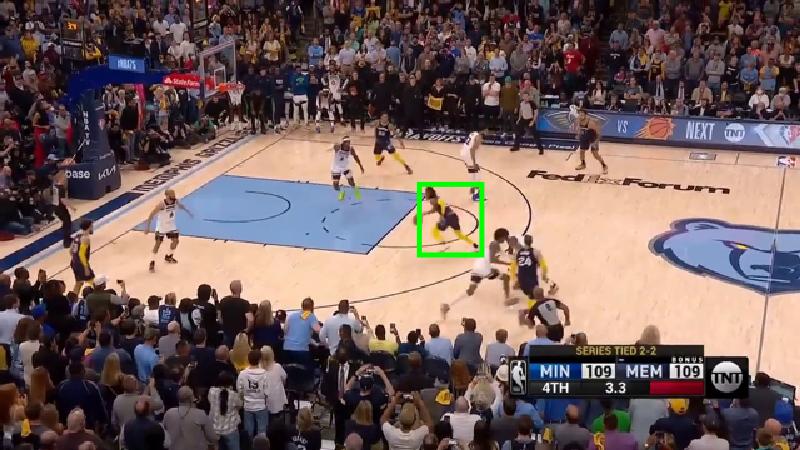}
		\includegraphics[width=0.19\textwidth]{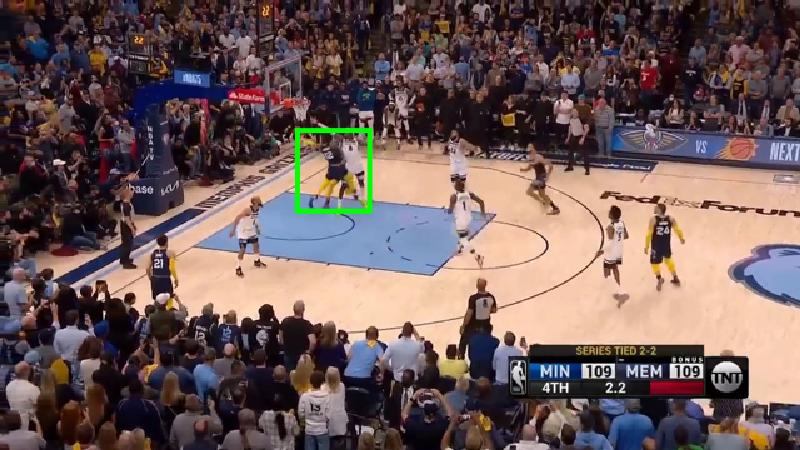}
        \includegraphics[width=0.19\textwidth]{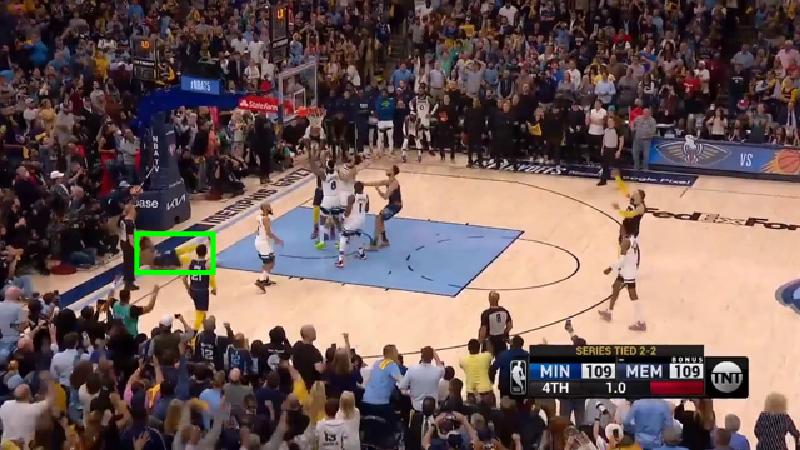}
		\includegraphics[width=0.19\textwidth]{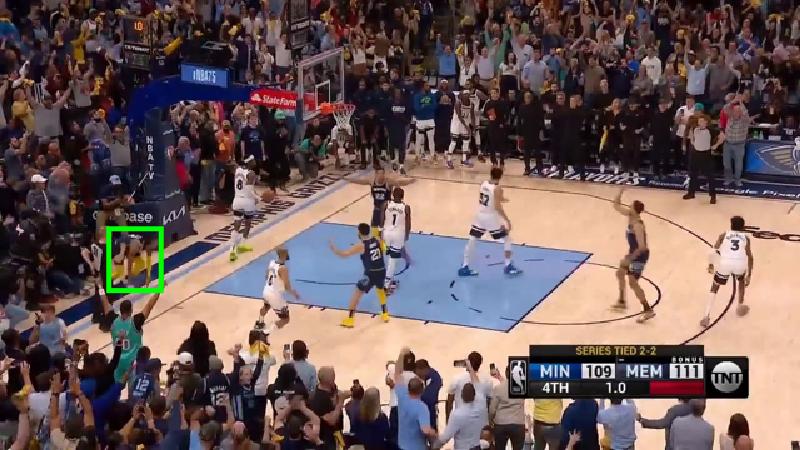}\\
		\scriptsize{Language: ``\emph{a basketball player in purple and yellow shooting the basketball}''}
		\\
		\includegraphics[width=0.19\textwidth]{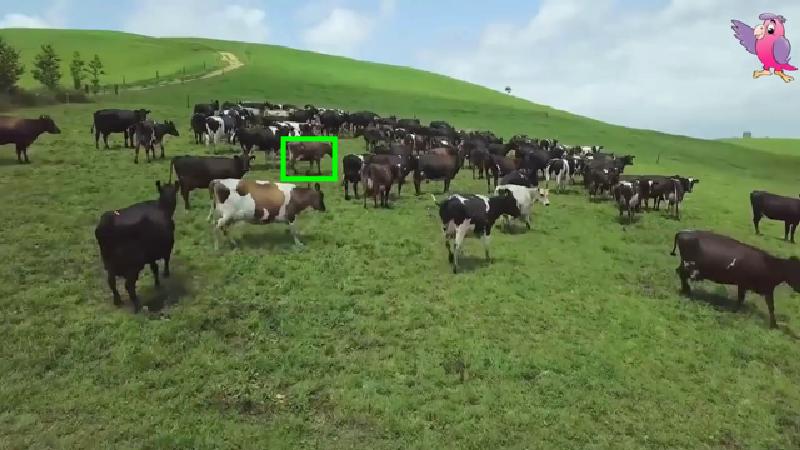}
        \includegraphics[width=0.19\textwidth]{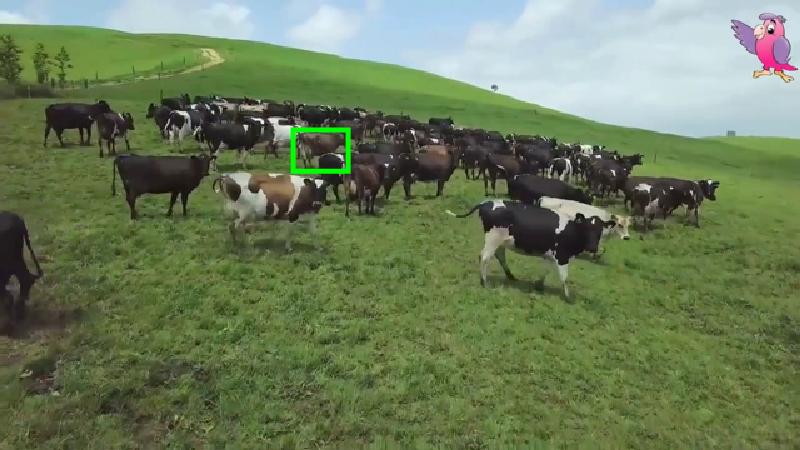} 
		\includegraphics[width=0.19\textwidth]{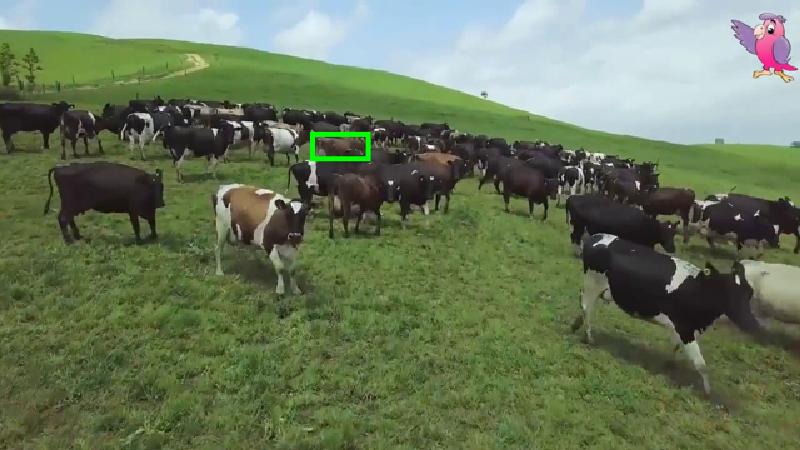}
        \includegraphics[width=0.19\textwidth]{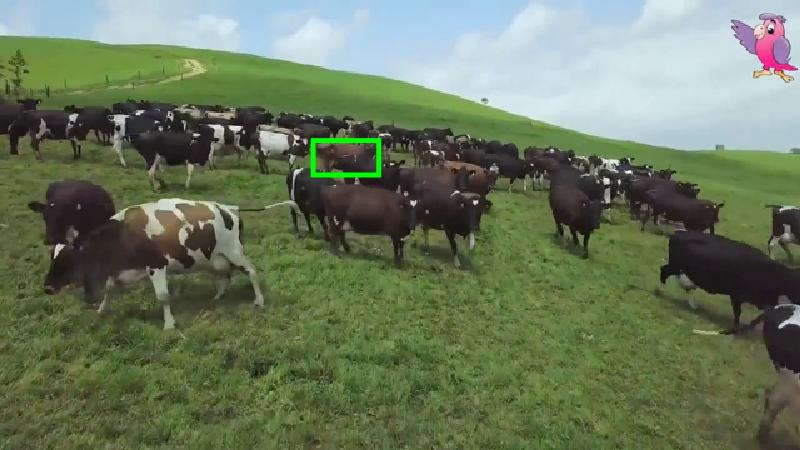}
		\includegraphics[width=0.19\textwidth]{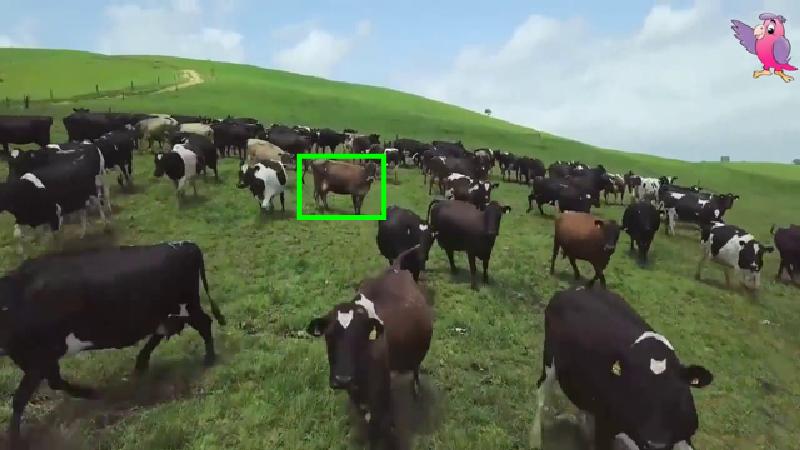}\\
		\scriptsize{Language: ``\emph{a brown cow walking on the pasture among other cows}''} \\
		\includegraphics[width=0.19\textwidth]{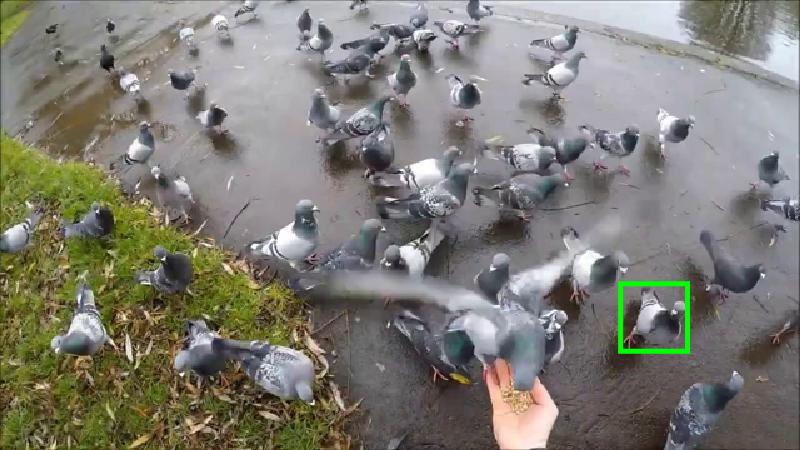} 
        \includegraphics[width=0.19\textwidth]{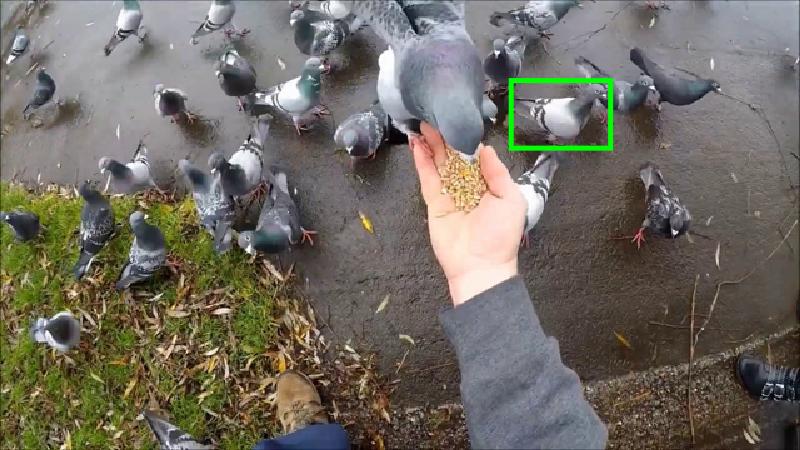}
		\includegraphics[width=0.19\textwidth]{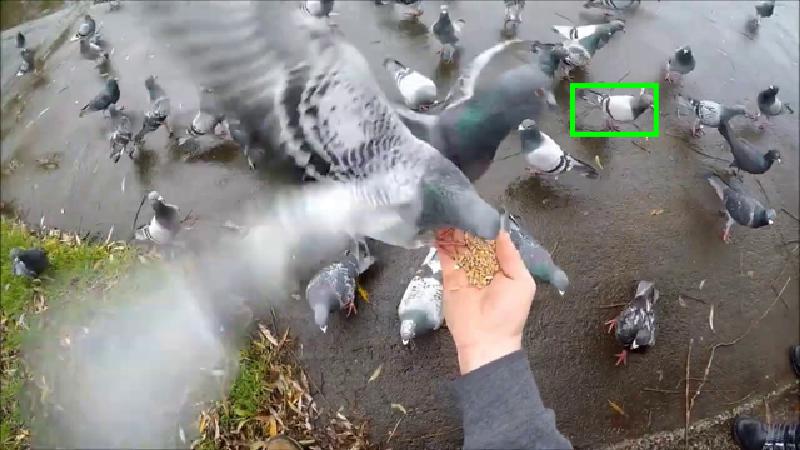}
        \includegraphics[width=0.19\textwidth]{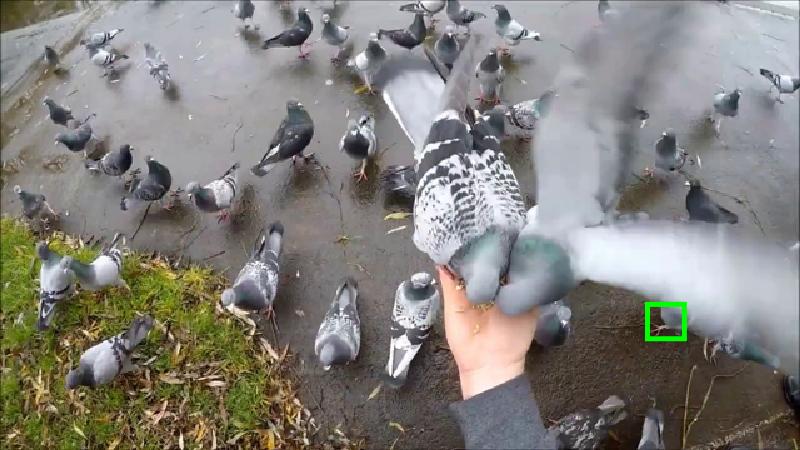}
		\includegraphics[width=0.19\textwidth]{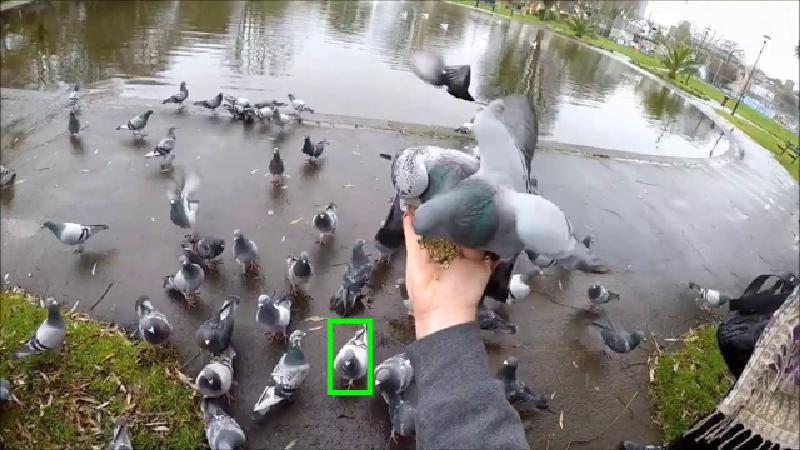} \\
		\scriptsize{Language: ``\emph{a feral pigeon in white and gray walking around a small puddle}''} \\
		\includegraphics[width=0.19\textwidth]{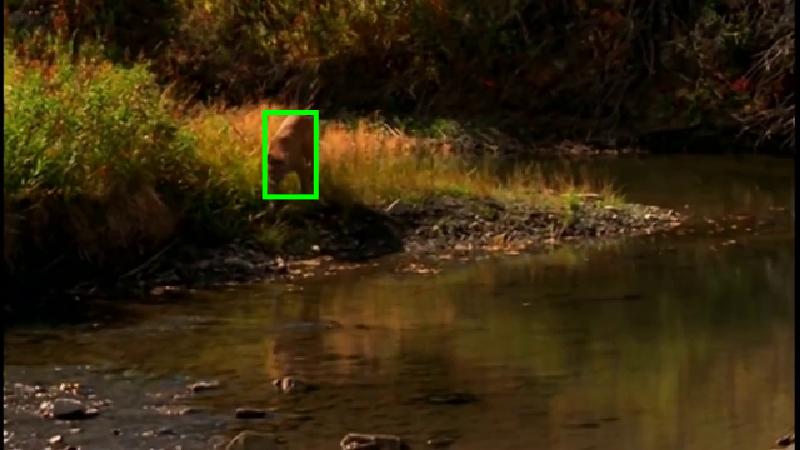} 
		\includegraphics[width=0.19\textwidth]{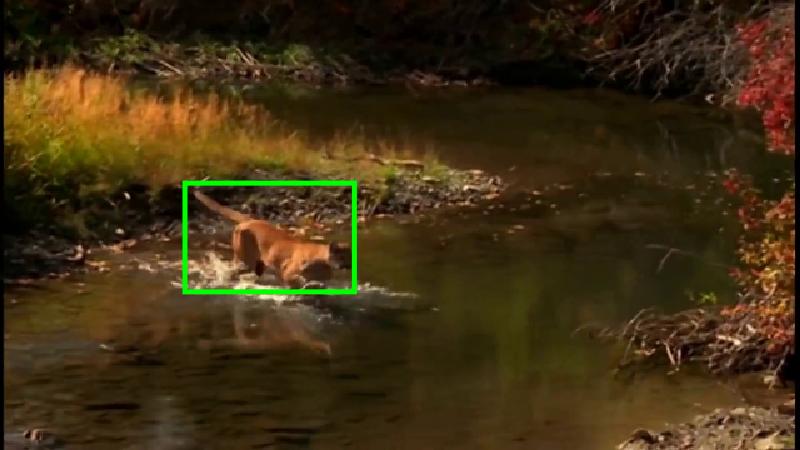}
        \includegraphics[width=0.19\textwidth]{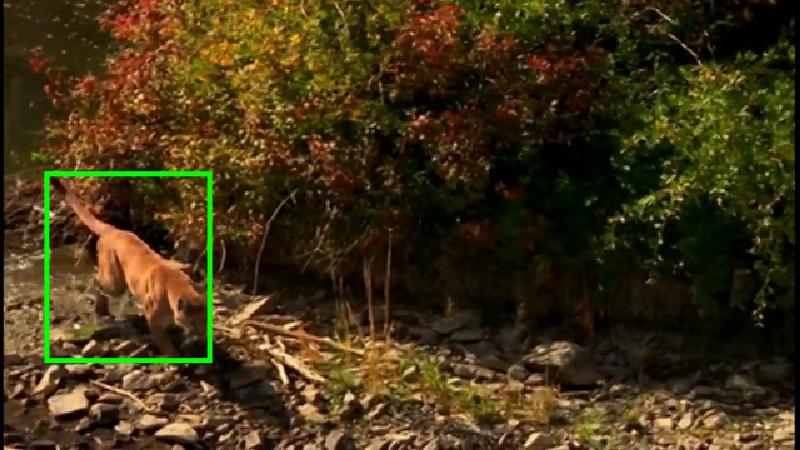}
        \includegraphics[width=0.19\textwidth]{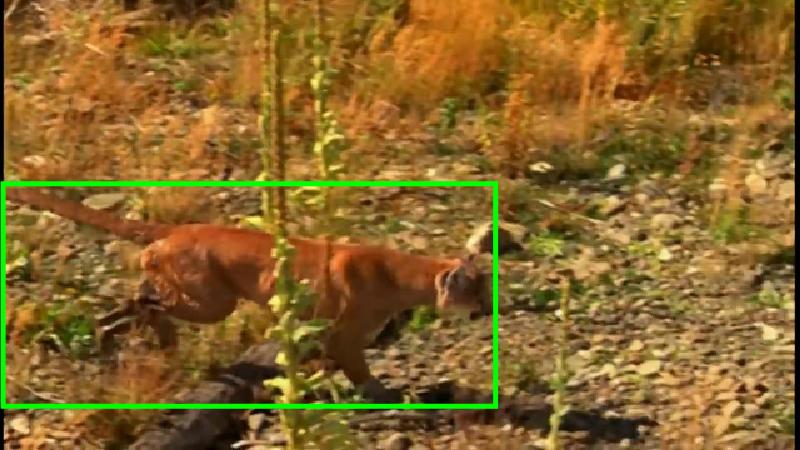}
		\includegraphics[width=0.19\textwidth]{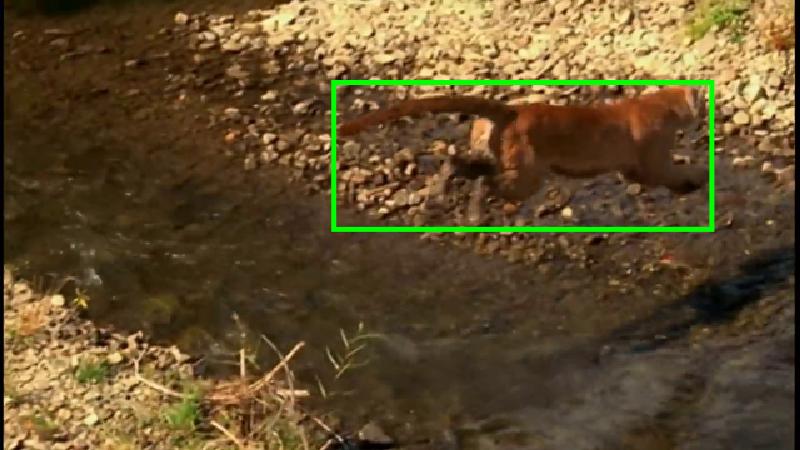} \\
		\scriptsize{Language: ``\emph{a brown cougar crossing the river and running on the ground}''} 
	\caption{Visualization of several annotation examples in the proposed VastTrack.
	}
	\label{anno_exa}
	\vspace{-2mm}
\end{figure}

Guided by the above principle, we compile an annotation team, which contains a few experts and a qualified labeling group, and adopt a multi-step mechanism, including manual labeling, visual inspection, and refinement. In the first step, after the experts label the initial target in the first frame, the annotation group starts to work on labeling the target in all other frames in the video. Notice that, to ensure consistency, each video is labeled (and refined if necessary) by the same annotator. After this, in the second step, the experts verify the completed annotations in the first step. If the annotation is not unanimously agreed by a validation team (formed by two or three experts), it will be returned back to the original labeler for refinement in the third step. Throughout the annotation process, the second and third steps are repeated for multiple rounds, which ensures high-quality annotations of VastTrack. Fig.~\ref{anno_exa} displays several annotation examples, and we show more statistics in the \textbf{supplementary material}. 

Considering great benefits of natural language in improving tracking robustness (\eg~\cite{li2017tracking,guo2022divert,feng2021siamese,zhou2023joint}), we provide language specifications, in addition to the bounding box annotations, for sequences in VastTrack, aiming to facilitate the development of vision-language trackers. In specific, a sentence of natural language that describes color information, behavior, and surroundings of the object as well as optionally its interaction with other objects is given as the linguistic annotation for the sequence (see Fig.~\ref{anno_exa} for examples). Although there have been benchmarks proposed for similar goal (\eg~\cite{fan2019lasot,wang2021towards} as in Tab.~\ref{tab:l}), their scale is limited by containing 1.4K~\cite{fan2019lasot} and 2K sentences~\cite{wang2021towards}. Differently, VastTrack offers over 50K videos with richer linguistic specifications for different objects, and thus may benefit learning more general and powerful vision-language trackers.

\subsection{Attributes}

To enable further in-depth analysis of trackers, we offer ten attributes for \emph{test} videos in VastTrack, including (1) invisibility (INV), assigned when object is partially or fully invisible due to occlusion or out of view, (2) deformation (DEF), assigned when target is deformable, (3) rotation (ROT), assigned when object rotates in the view, (4) aspect ratio change (ARC), assigned when ratio of bounding box aspect ratio is outside [0.5, 2], (5) illumination variation (IV), assigned when illumination in object region heavily varies, (6) scale variation (SV), assigned when ratio of bounding box is outside [0.5, 2], (7) fast motion (FM), assigned when target center moves larger than its size in last frame, (8) motion blur (MB), assigned when blur in object regions occurs (9) background clutter (BC), assigned when the similar appearance (not necessarily the same class of target) as target appears, and (10) low resolution (LR), assigned when target region is less than 1,000 pixels. For each sequence, a 10D binary vector is adopted to indicate the presence of an attribute, \ie, ``1'' for presence , ``0'' otherwise.

\begin{wrapfigure}{r}{0.5\textwidth}
\centering
\vspace{-5mm}
\includegraphics[width=0.5\textwidth]{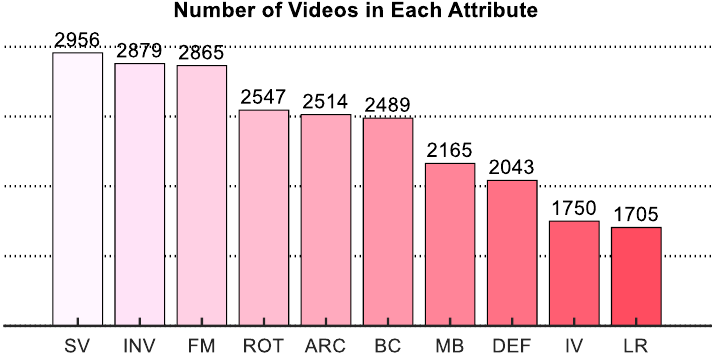}
\caption{Distribution of videos per attribute.}
\vspace{-10pt}
\label{fig:4}
\end{wrapfigure}
The distribution of attributes for the test videos of VastTrack is exhibited in Fig.~\ref{fig:4}. As shown in Fig.~\ref{fig:4}, we can observe that the most common challenge is scale variation, which is involved with 2,956 videos in the test set. In addition, invisibility that may be caused by partial or full occlusion or out-of-view and fast motion also frequently occur with 2,879 and 2,865 videos, respectively.

\subsection{Dataset Split and Evaluation Protocol}

\setlength{\columnsep}{10pt}%
\begin{wraptable}{r}{0.5\textwidth}\scriptsize
\vspace{-15pt}
\centering
\vspace{-5mm}
\caption{Comparison between the \emph{training} and \emph{testing} sets of VastTrack.}
    \scalebox{0.95}{
	\begin{tabular}{@{}L{1.5cm}C{1cm}@{}C{1.2cm}@{}C{1.2cm}@{}C{1.3cm}@{}}
		\specialrule{.1em}{.05em}{.05em} 
		& \textbf{Classes} & \textbf{Videos} & \tabincell{c}{\textbf{Mean} \\\textbf{frames}} & \tabincell{c}{\textbf{Total} \\\textbf{frames}}  \\
		\hline\hline
		VastTrack$_\mathrm{Tst}$ &702     & 3,500    &  106.3   &  372$\textbf{K}$   \\
		VastTrack$_\mathrm{Tra}$ & 1,974    & 47,110    & 81.2    &   3.82$\textbf{M}$  \\
		\specialrule{.1em}{.05em}{.05em} 
        \end{tabular}}%
\label{tab2}%
\vspace{-10pt}
\end{wraptable}

\textbf{Dataset Split.} VastTrack comprises 50,610 video sequences. Among them, 47,110 videos are used in the training set named VastTrack$_{\text{Tra}}$, and the rest 3,500 sequences for testing set named VastTrack$_{\text{Tst}}$. Tab.~\ref{tab2} displays comparison of training and testing sets. In dataset split, we try to keep distributions of training and testing sets similar. Please notice that, the reason to use 3,500 videos ($\sim$7\% of total) in VastTrack$_{\text{Tst}}$ is to keep it relatively compact so that evaluation of trackers can be fast, similar to the popular GOT-10k~\cite{huang2019got} in which 420 sequences out of around 10K are for testing ($\sim$4.2\% of the total). Although VastTrack$_{\text{Tst}}$ contains only 3,500 sequences, it is representative by including rich categories and various scenarios for tracking assessment, and much larger and more diverse compared with other testing sets regarding video number and classes, making evaluation reliable.

\vspace{0.3em}
\noindent
\textbf{Evaluation Protocol.} Unlike the \emph{full overlap}~\cite{fan2019lasot,muller2018trackingnet} or \emph{one-shot}~\cite{huang2019got}, we utilize a \emph{hybrid} evaluation protocol in which the training and testing sets have overlap in classes. The reason is that, in real world, humans usually track both frequently seen and unseen objects. To develop human-like trackers, we divide VastTrack into training and testing sets with partial overlap of 561 categories.

\section{Experiments}
\label{exp}

\subsection{Evaluation Methodology}

Following existing popular  benchmarks~\cite{wu2013online,fan2019lasot,muller2018trackingnet}, we employ the \emph{one-pass evaluation} (OPE) and compare different visual trackers using three common metrics, including \emph{precision} (PRE) and normalized precision (NPRE) as well as success (SUC). In specific, PRE measures the center position distance between  tracking results and groundtruth in pixels, and trackers are ranked by PRE on a preset threshold, \eg, 20 pixels. In order to mitigate the influence of video resolutions, NPRE is calculated by normalizing PRE using target region. Different from PRE and NPRE, SUC measures Intersection over Union (IoU) between the tracking results and groundtruth boxes, and is computed by the percentage of frames in which the IoU is larger than a threshold, \eg, 0.5.

\subsection{Evaluated Trackers}

\setlength{\columnsep}{10pt}%
\renewcommand\arraystretch{1.35}
\begin{wraptable}{r}{0.5\textwidth}\scriptsize
\vspace{-20pt}
\centering
\vspace{-5mm}
\caption{Summary of algorithms. ``CNN'': CNN-based; ``CNN-T'': CNN-Transformer-based, ``Trans.'': Transformer-based. TP: ``\ding{51}'' for trackers leveraging temporal information, and ``\ding{55}'' for trackers using only the information from initial frame for tracking.}
\scalebox{0.77}{
 \tabcolsep=0.08cm
	\begin{tabular}{rrcccc}
		\Xhline{2\arrayrulewidth}
		\textbf{Tracker} & \textbf{Where} & \textbf{Backbone} & \textbf{\textbf{Type}} & \textbf{TP}  \\
		\hline\hline
		SiamFC~\cite{bertinetto2016fully} & ECCVW'16 & AlexNet & CNN & \ding{55}  \\
		ATOM~\cite{danelljan2019atom} & CVPR'19 & ResNet-18 & CNN & \ding{51}  \\
		SiamRPN++~\cite{li2019siamrpn++} & CVPR'19 & ResNet-50 & CNN & \ding{55} \\
		 DiMP~\cite{bhat2019learning} & ICCV'19 & ResNet-50 & CNN & \ding{51}  \\
		SiamBAN~\cite{chen2020siamese} & CVPR'20 & ResNet-50 & CNN & \ding{55}  \\
		SiamCAR~\cite{guo2020siamcar} & CVPR'20 & ResNet-50 & CNN & \ding{55}  \\
		PrDiMP~\cite{danelljan2020probabilistic} & CVPR'20 & ResNet-50 & CNN & \ding{51}  \\
		Ocean~\cite{zhang2020ocean} & ECCV'20 & ResNet-50 & CNN & \ding{51}  \\
		 STMTrack~\cite{fu2021stmtrack} & CVPR'21 & GoogLeNet & CNN & \ding{51}  \\
		TrSiam~\cite{wang2021transformer} & CVPR'21 & ResNet-50 & CNN-T & \ding{51}  \\
		 TransT~\cite{chen2021transformer} & CVPR'21 & ResNet-50 & CNN-T & \ding{55}  \\
		STARK~\cite{yan2021learning} & ICCV'21 & ResNet-101 & CNN-T & \ding{51}  \\
		 AutoMatch~\cite{zhang2021learn} & ICCV'21 & ResNet-50 & CNN & \ding{55}  \\
		 ToMP~\cite{mayer2022transforming} & CVPR'22 & ResNet-101 & CNN-T & \ding{51}  \\
		 MixFormer (L)~\cite{cui2022mixformer} & CVPR'22 & CVT24-W & Trans. & \ding{51}  \\
		 OSTrack (384)~\cite{ye2022joint} & ECCV'22 & ViT-Base & Trans. & \ding{55}  \\
		 RTS~\cite{paul2022robust} & ECCV'22 & ResNet-50 & CNN & \ding{51}  \\
		SimTrack (224)~\cite{chen2022backbone} & ECCV'22 & ViT-Large & Trans. & \ding{55}  \\
		SwinTrack (224)~\cite{lin2022swintrack} & NeurIPS'22 & SwinT & Trans. & \ding{51}  \\
		SeqTrack (L384)~\cite{chen2023seqtrack} & CVPR'23 & ViT-Large & Trans. & \ding{51}  \\
		 ARTrack (256)~\cite{wei2023autoregressive} & CVPR'23 & ViT-Large & Trans. & \ding{51}  \\
		DropMAE~\cite{wu2023dropmae} & CVPR'23 & ViT & Trans. & \ding{51}  \\
		 GRM (256)~\cite{gao2023generalized} & CVPR'23 & ViT-Base & Trans. & \ding{55}  \\
		ROMTrack (384)~\cite{cai2023robust} & ICCV'23 & ViT-Base & Trans. & \ding{51}  \\
		MixFormerV2 (B)~\cite{cui2023mixformerv2} & NeurIPS’23 & ViT-Base & Trans. & \ding{51} \\
		\Xhline{2\arrayrulewidth}
	\end{tabular}}
	\label{tab:3}
\vspace{-10pt}
\end{wraptable}
In order to understand performance of existing approaches and also to provide baselines for future comparison on VastTrack, we extensively evaluate 25 representative trackers from different periods, which could be classified into three categories:

\vspace{0.3em}
\noindent
\textbf{(i) CNN-based} that achieves object tracking using only CNN architecture consisting of SiamFC~\cite{bertinetto2016fully}, ATOM~\cite{danelljan2019atom}, 
SiamRPN++~\cite{li2019siamrpn++}, 
SiamBAN~\cite{chen2020siamese},
DiMP~\cite{bhat2019learning},  SiamCAR~\cite{guo2020siamcar}, PrDiMP~\cite{danelljan2020probabilistic}, STMTrack~\cite{fu2021stmtrack}, Ocean~\cite{zhang2020ocean}, RTS~\cite{paul2022robust}, and AutoMatch~\cite{zhang2021learn};

\vspace{0.3em}
\noindent
\textbf{(ii) CNN-Transformer-based} that implements visual tracking via hybrid CNN and Transformer architectures, including STARK~\cite{yan2021learning}, TrSiam~\cite{wang2021transformer}, TransT~\cite{chen2021transformer}, and ToMP~\cite{mayer2022transforming}; 

\vspace{0.3em}
\noindent
\textbf{(iii) Transformer-based} that tracks the target objects through leveraging a pure Transformer architecture. The visual object tracking models in this category consist of OSTrack~\cite{ye2022joint}, SwinTrack~\cite{lin2022swintrack}, MixFormer~\cite{cui2022mixformer} and MixFormerV2~\cite{cui2023mixformerv2}, SimTrack~\cite{chen2022backbone}, SeqTrack~\cite{chen2023seqtrack}, ARTrack~\cite{wei2023autoregressive}, DropMAE~\cite{wu2023dropmae}, ROMTrack~\cite{cai2023robust}, and GRM~\cite{gao2023generalized}. 

It is worth noting that, all the aforementioned evaluated trackers are assessed as they are, without modifications. Tab.~\ref{tab:3} summarizes these tracking algorithms.

\subsection{Evaluation Results}

\begin{figure*}[!t]
	\centering
	\includegraphics[width=\linewidth]{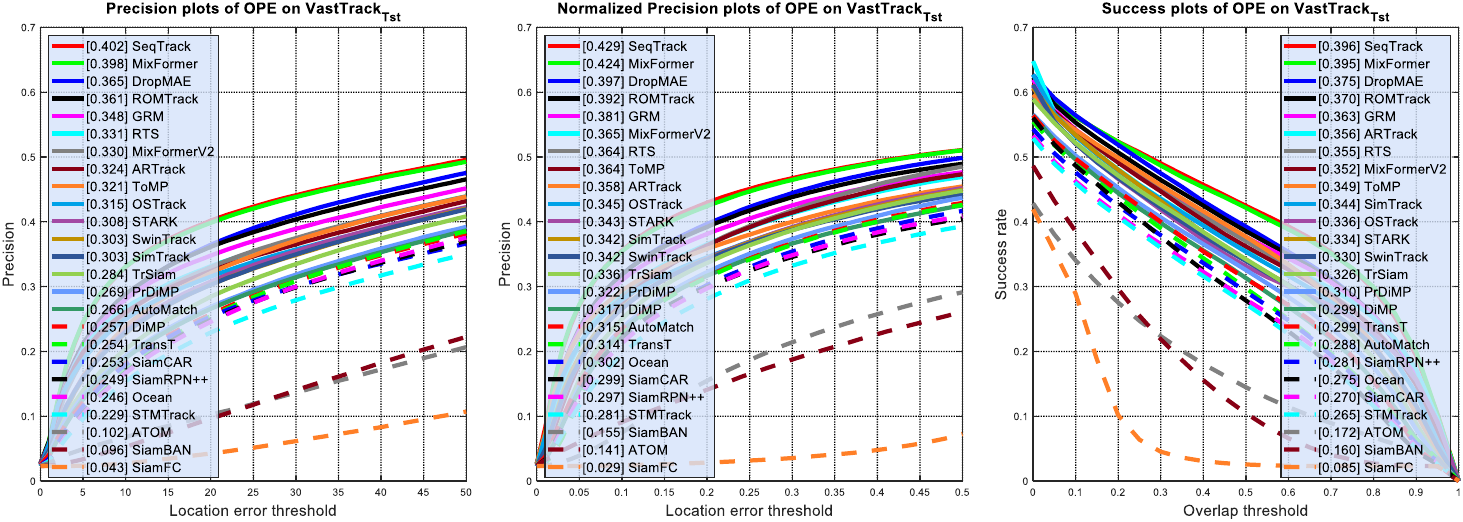}
	\caption{Evaluation results of 25 trackers on VastTrack$_{\text{Tst}}$ using PRE, NPRE, and SUC.}
	\label{fig:overall}
	\vspace{-3mm}
\end{figure*}

\textbf{Overall Performance.} We evaluate 25 trackers on VastTrack, including many recently proposed Transformer-based methods. Note that, during evaluation, each tracker is evaluated as it is, without any modification. The evaluation results are reported in OPE using PRE, NPRE, and SUC scores, as in Fig.~\ref{fig:overall}. We can see that, SeqTrack achieves the best performance on all three metrics with 0.402 PRE, 0.429 NPRE, and 0.396 SUC scores, MixFormer displays the second best results with 0.398 PRE, 0.424, and 0.395 SUC scores, and DropMAE obtains the third best results with 0.365 PRE, 0.397 NPRE, and 0.375 SUC scores. All these three trackers are developed based on vision Transformer architecture, showing its power in feature learning for tracking. Notably, although RTS does not employ Transformer architecture for tracking, it still achieves promising results with 0.331 PRE, 0.364 NPRE, and 0.355 SUC scores, even better than a few Transformer trackers like OSTrack with 0.315 PRE, 0.345 NPRE, and 0.336 SUC scores and SwinTrack with 0.303 PRE, 0.342 NPRE, and 0.330 SUC scores. We argue this is because RTS adopts tracking-by-segmentation which is beneficial for tracking object with extreme aspect ratio. Note that, the recent MixFormerV2 with 0.330 PRE, 0.365 NPRE, and 0.352 SUC scores performs worse than its previous version MixFormer, because it leverages much lighter network for efficiency. An interesting observation is that, SiamRPM++, one of the seminal Siamese trackers, surprisingly outperforms many its extensions such as SiamCAR, Ocean, and SiamBAN, which shows its generality to some extent.

\textbf{Discussion.} In addition, the evaluation also shows some useful observations: (1) Feature network. From the overall evaluation in Fig.~\ref{fig:overall}, we observe that, the top five tracking approaches are based on Vision Transformer architecture, which reveals that the exploration of more powerful for tracking is still an important direction for improving tracking performance. This is consistent with finds in other benchmarks. Despite adopting powerful feature network, the performance is still far from satisfaction, compared to that on other existing benchmarks (as shown later in the comparison to other benchmarks). We argue this is caused by the lack of universal large-scale training of more general object categories for tracking. (2) Temporal information. Video sequences contain abundant temporal information which is important for visual tracking. However, this is largely ignored to some extend owing to the great success of Siamese tracking in recent years. Especially, even without using temporal information, many trackers still achieve state-of-the-art performance. However, from the evaluation results of existing trackers, we can see that, the top three trackers all leverage temporal information for tracking, which indicates the crucial role of temporal cue for tracking. Particularly, when target suffers from complicated challenges, temporal information provides an useful source to infer target position. We hope, through evaluation results on VastTrack, researchers can pay more attention in how to develop robust tracking by incorporating temporal cues.

\begin{figure*}[!t]
	\centering
	\includegraphics[width=\linewidth]{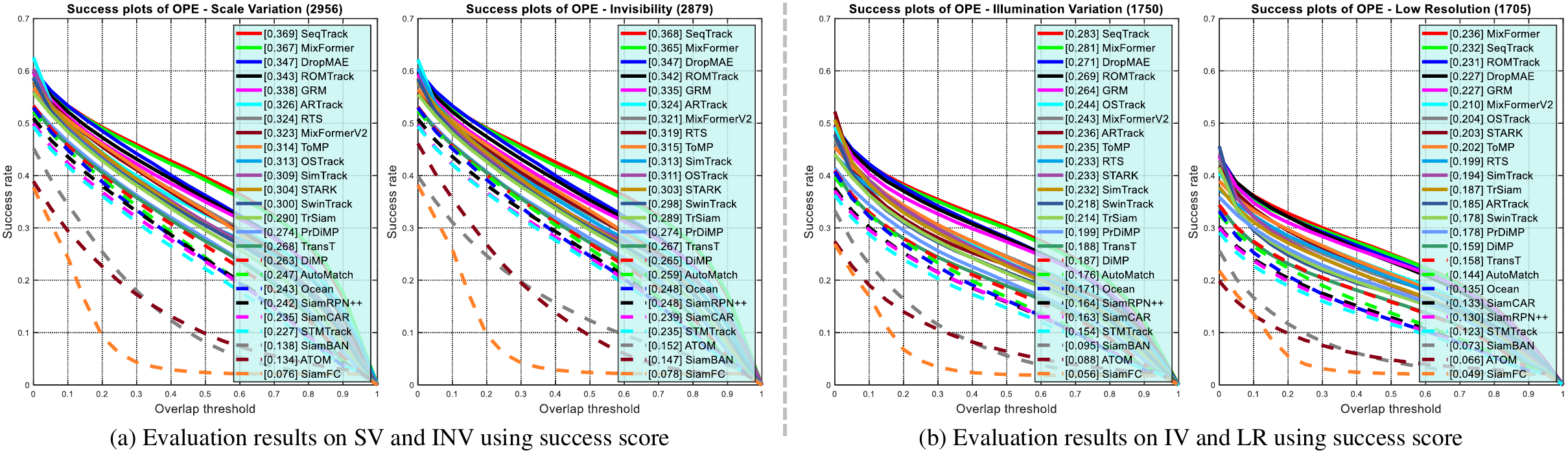}
	\caption{Evaluation results on the most two common attributes (see image (a)) and on the two difficult attributes (see image (b)) using the metric of success score.}
	\label{fig:att}
\end{figure*}

\begin{figure*}[!t]
	\centering
	\begin{tabular}{c@{\hspace{1.8mm}}c}
		\includegraphics[width=0.155\textwidth]{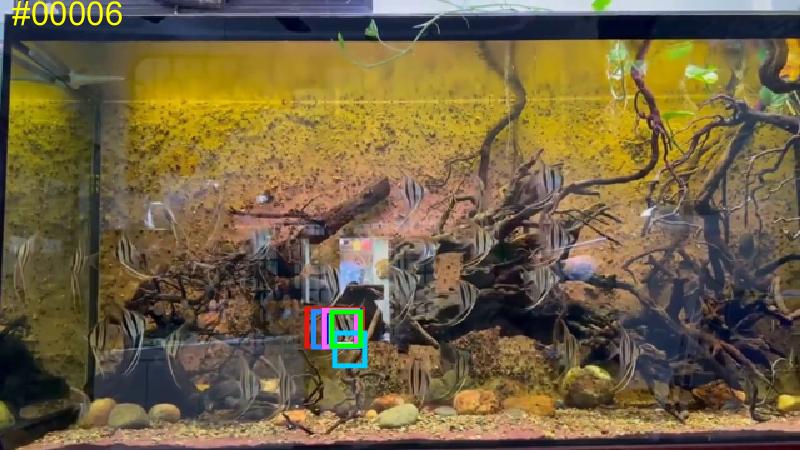} \includegraphics[width=0.155\textwidth]{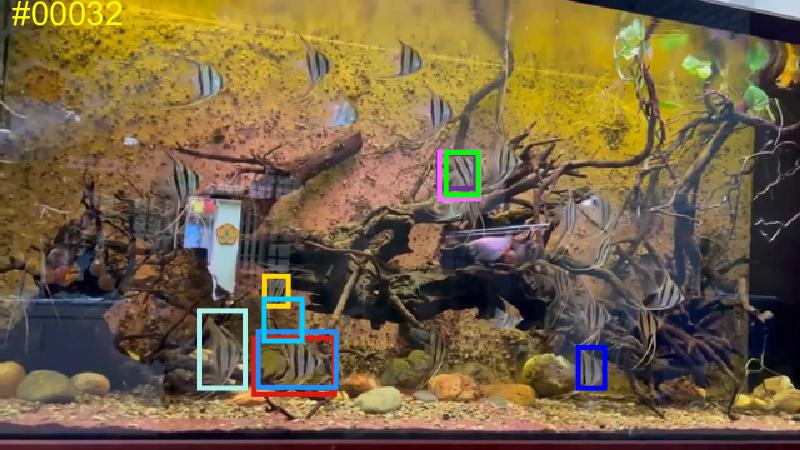} \includegraphics[width=0.155\textwidth]{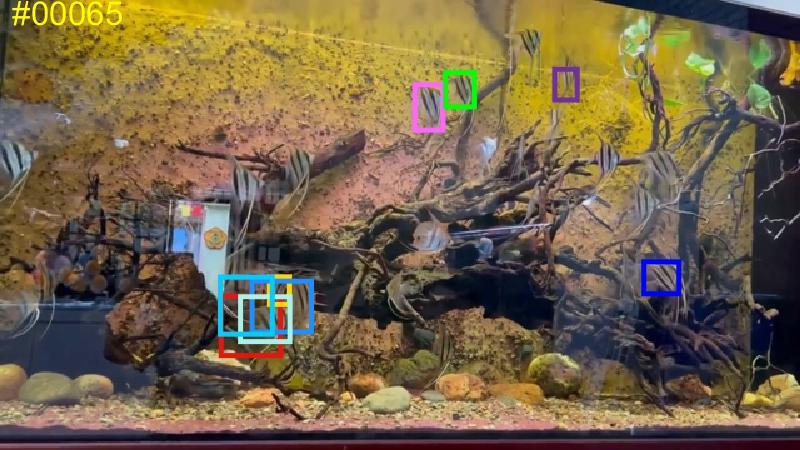} & \includegraphics[width=0.155\textwidth]{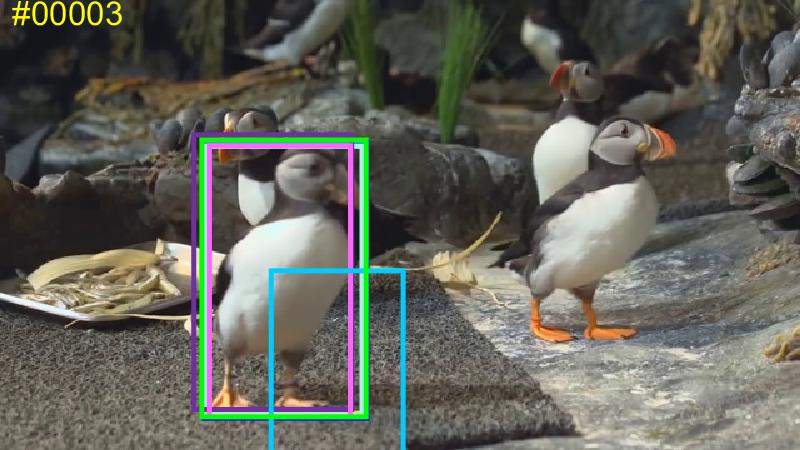}  \includegraphics[width=0.155\textwidth]{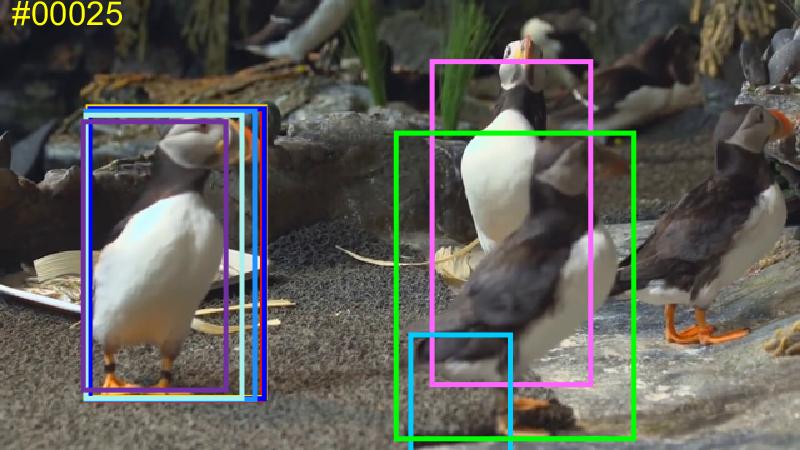} \includegraphics[width=0.155\textwidth]{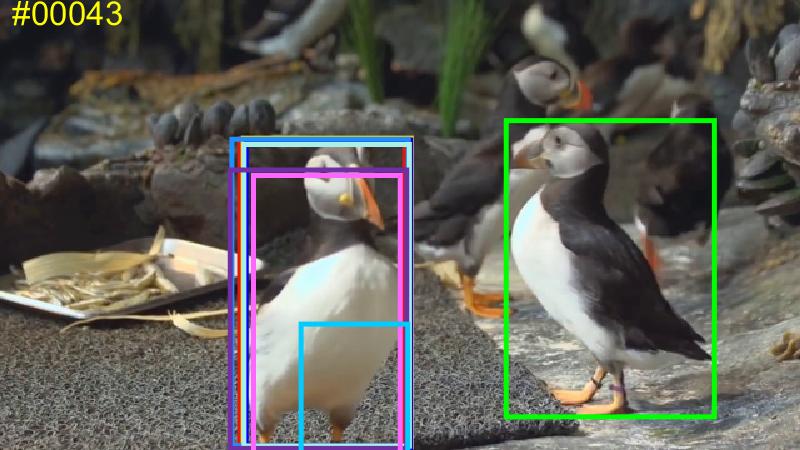} \\
		{\scriptsize (a) \emph{Angelfish-18} with BC and SV}& {\scriptsize (b) \emph{Atlantic Puffin-35} with INV and BC} \\
		\includegraphics[width=0.155\textwidth]{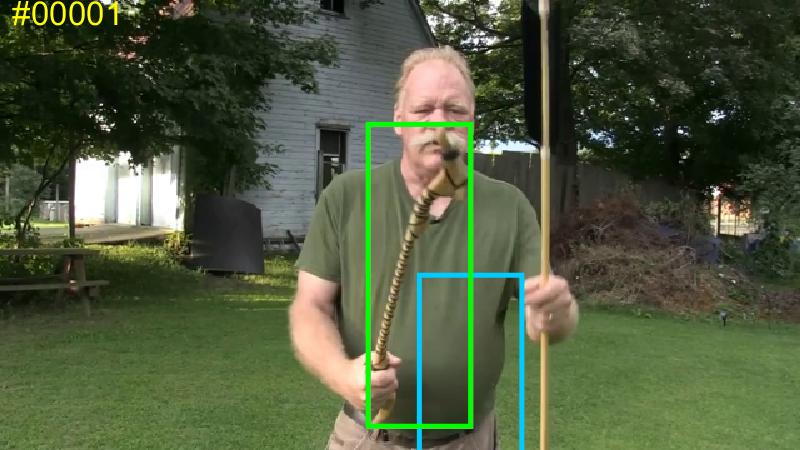} \includegraphics[width=0.155\textwidth]{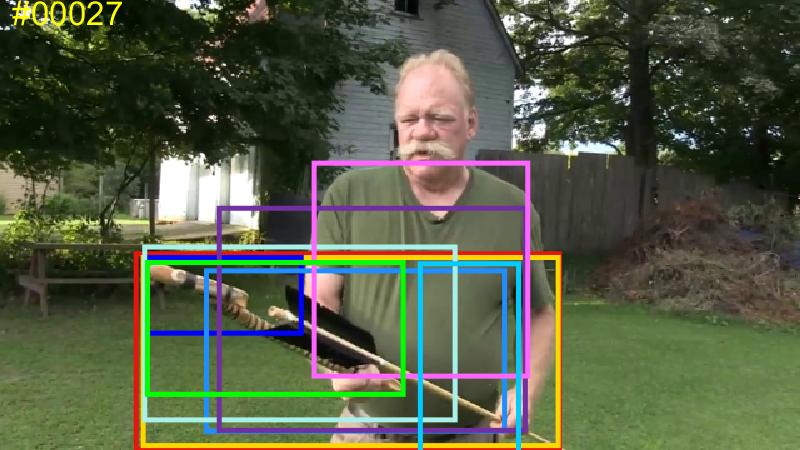} \includegraphics[width=0.155\textwidth]{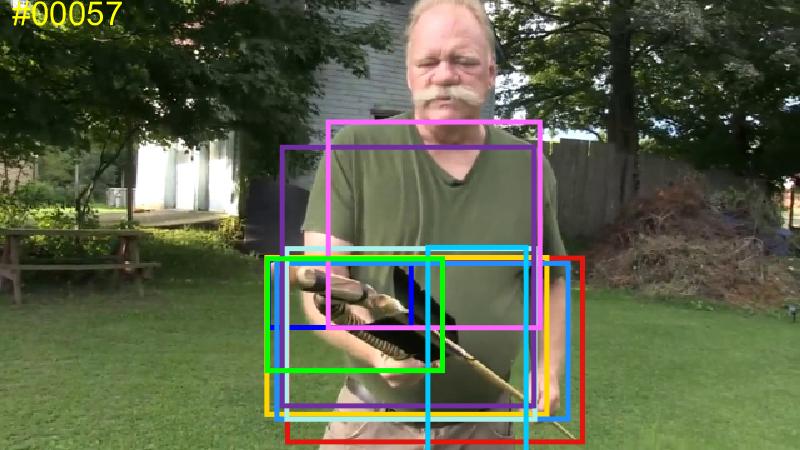} & \includegraphics[width=0.155\textwidth]{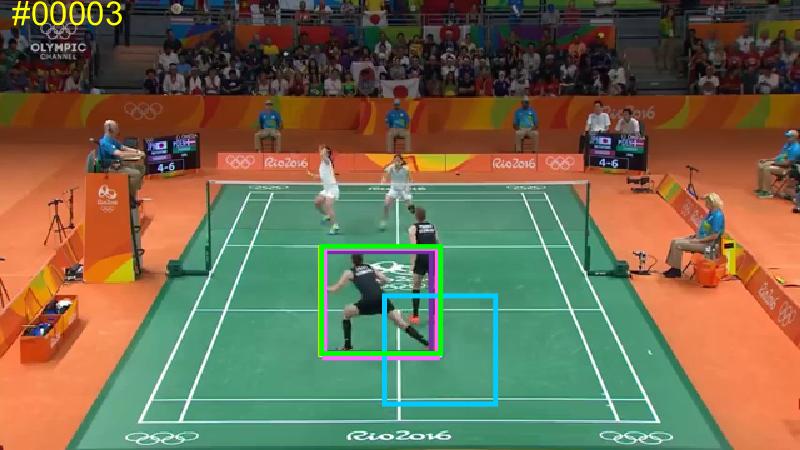}  \includegraphics[width=0.155\textwidth]{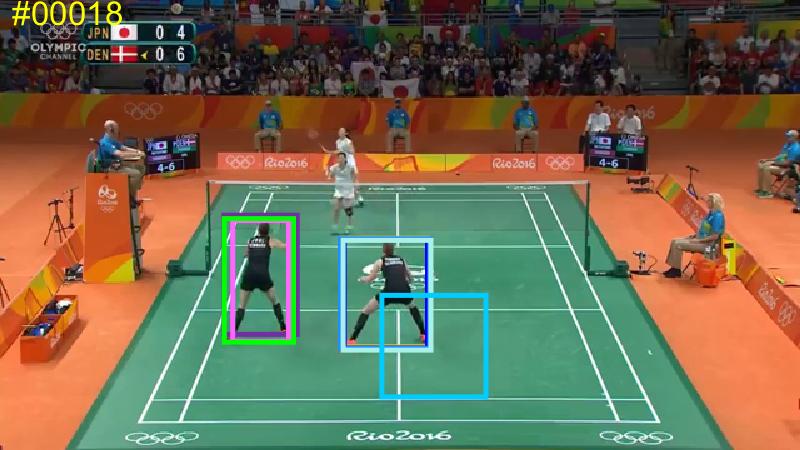} \includegraphics[width=0.155\textwidth]{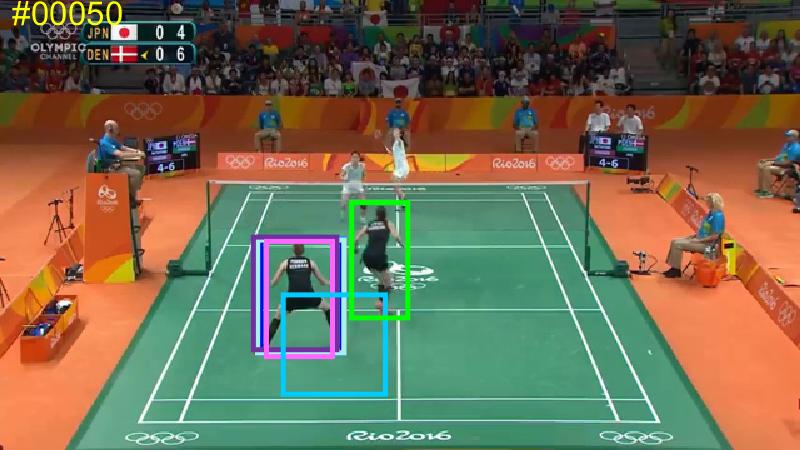} \\
		{\scriptsize (c) \emph{Atlatl-27} with SV and ARC}& {\scriptsize (d) \emph{Badminton player-17} with INV and DEF} \\
		\includegraphics[width=0.155\textwidth]{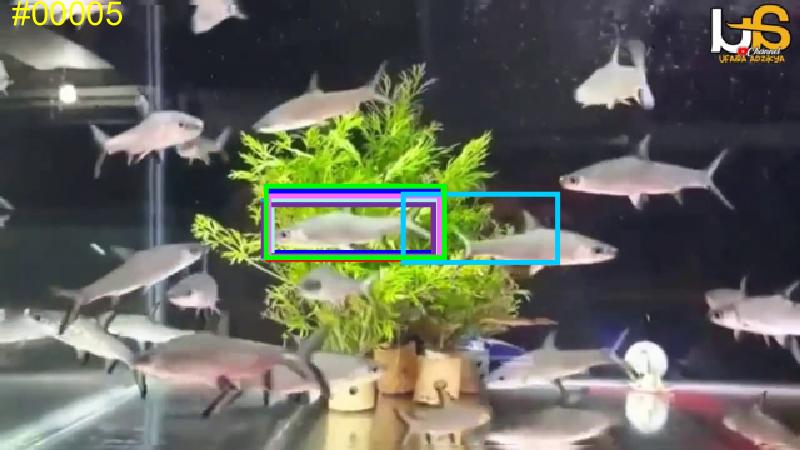} \includegraphics[width=0.155\textwidth]{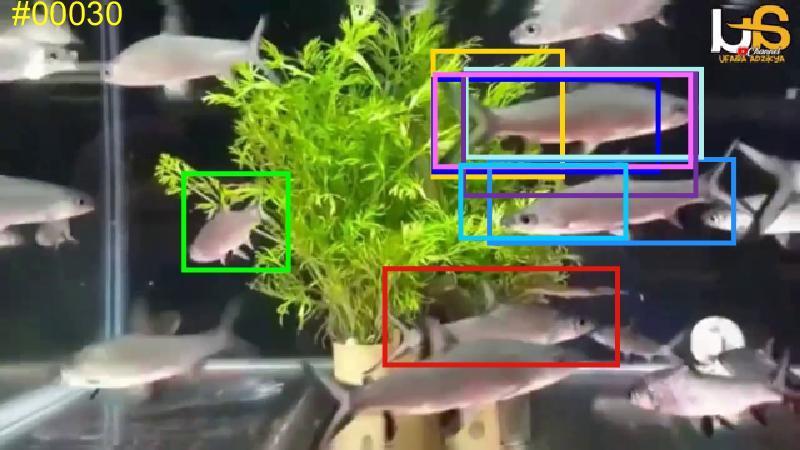} \includegraphics[width=0.155\textwidth]{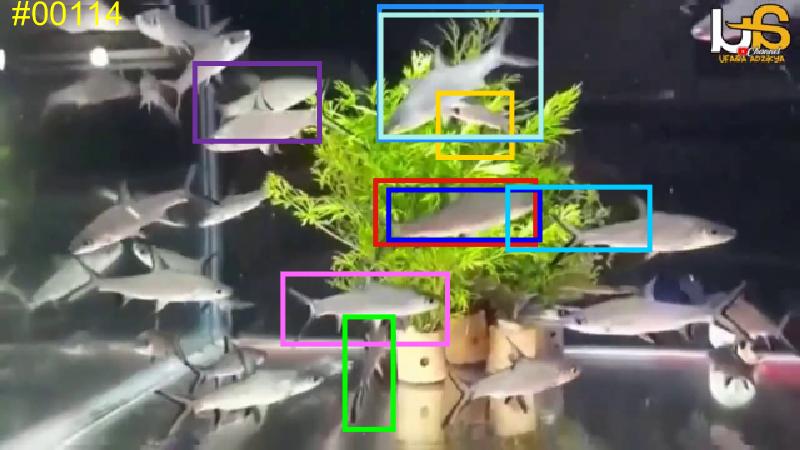} & \includegraphics[width=0.155\textwidth]{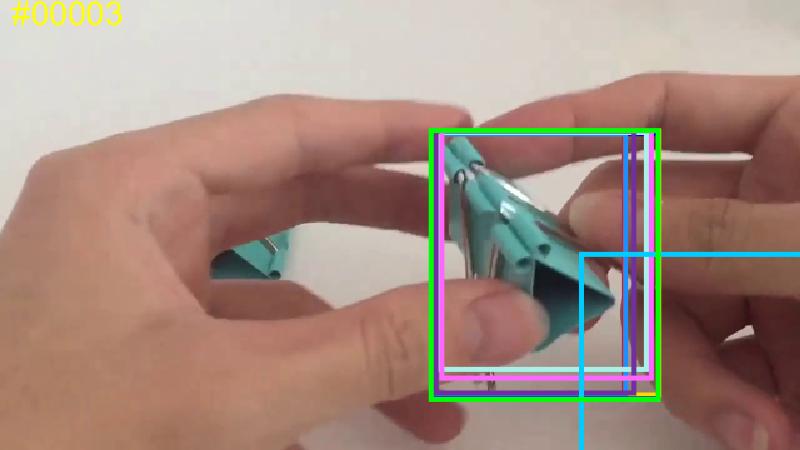}  \includegraphics[width=0.155\textwidth]{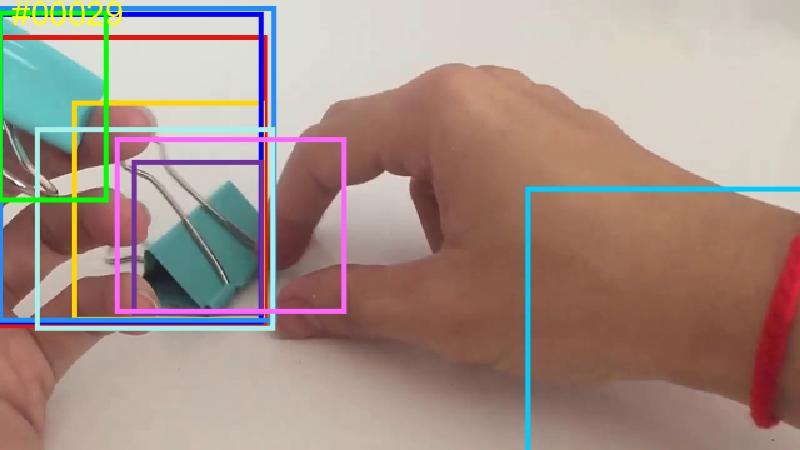} \includegraphics[width=0.155\textwidth]{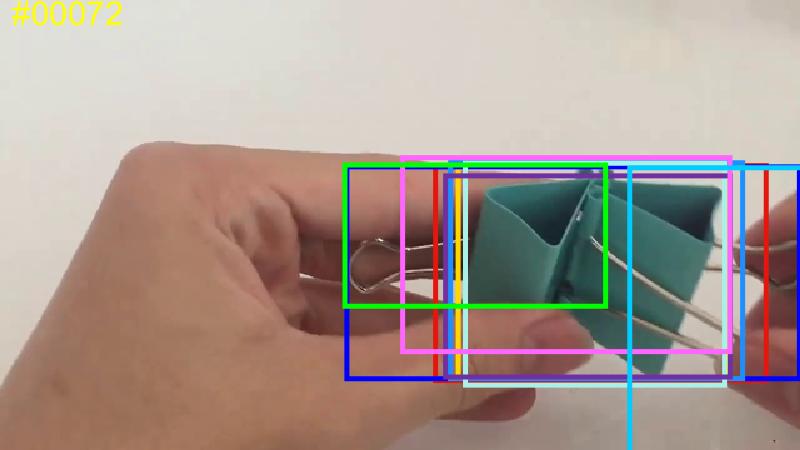} \\
		{\scriptsize (e) \emph{Bala Shark-31} with ROT and BC}& {\scriptsize (f) \emph{Clip-28} with INV and ROT} \\
		\includegraphics[width=0.155\textwidth]{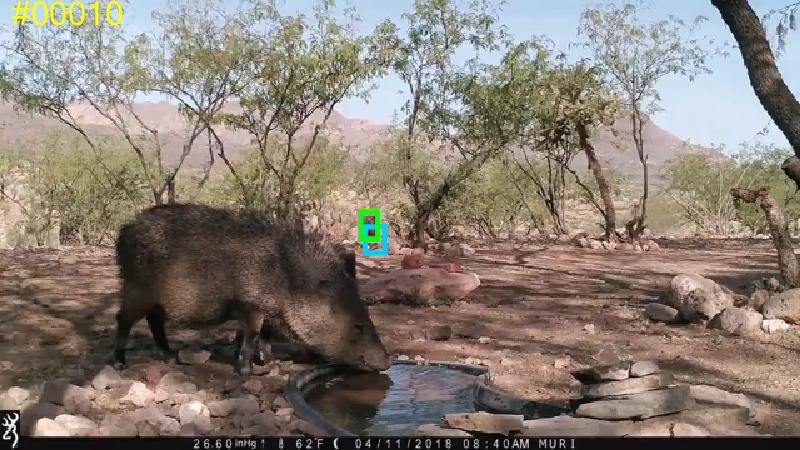} \includegraphics[width=0.155\textwidth]{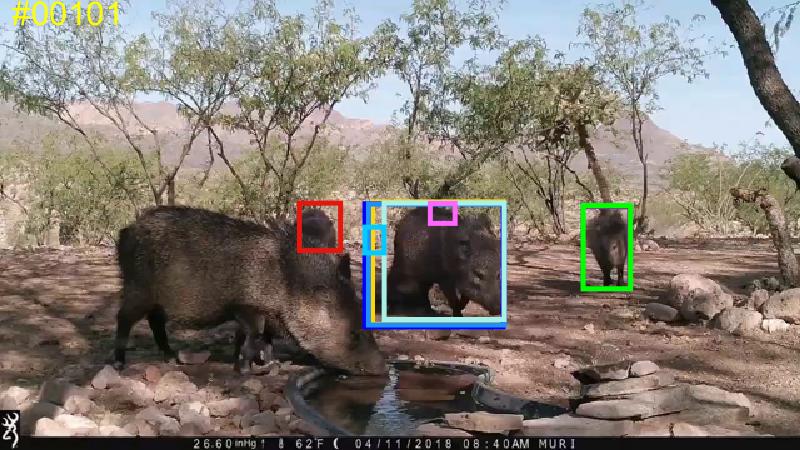} \includegraphics[width=0.155\textwidth]{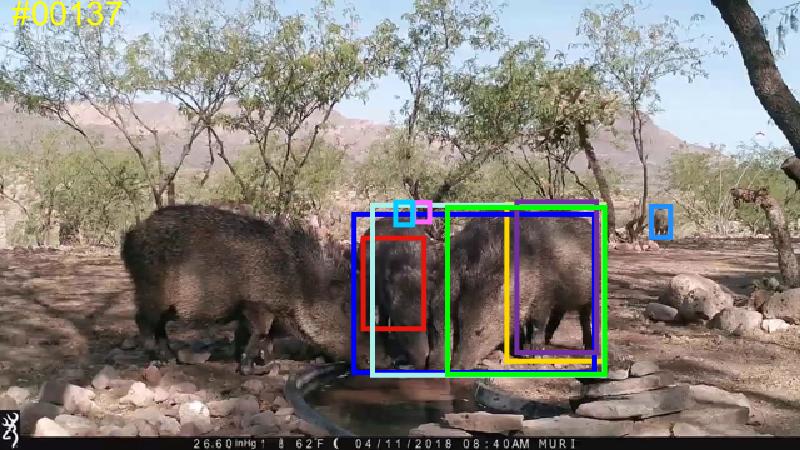} & \includegraphics[width=0.155\textwidth]{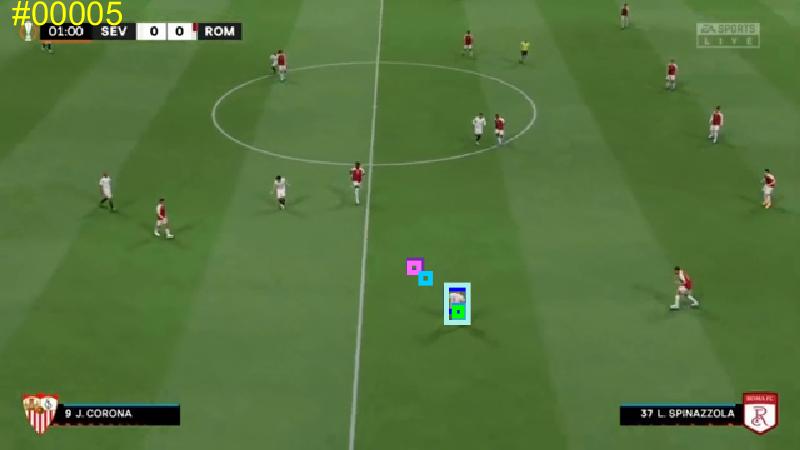}  \includegraphics[width=0.155\textwidth]{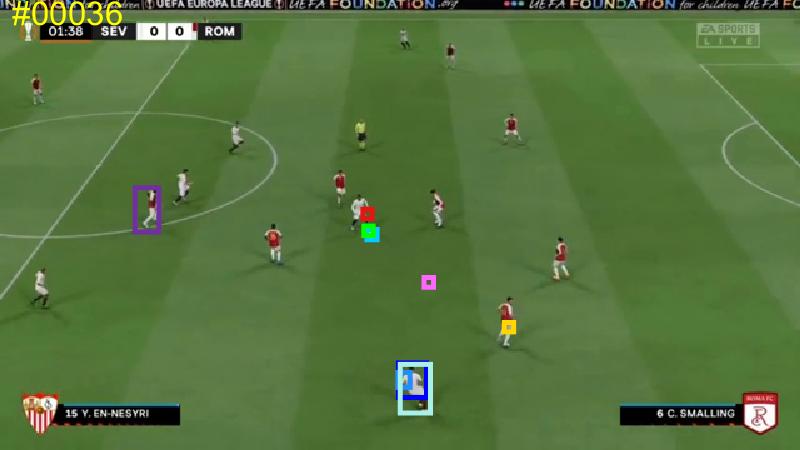} \includegraphics[width=0.155\textwidth]{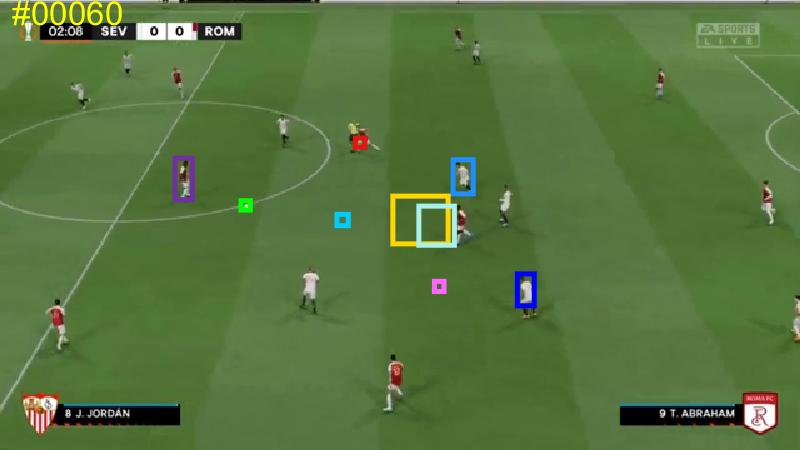} \\
		{\scriptsize (g) \emph{Collared peccary-24} with INV and SV}& {\scriptsize (h) \emph{Football-sq30} with LR, FM, and MB} \\
		\multicolumn{2}{c}{\includegraphics[width=0.9\textwidth]{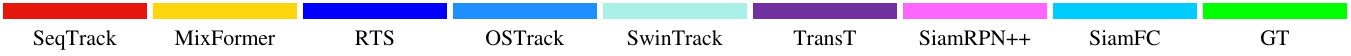}}\\
	\end{tabular}
	\caption{Qualitative results of eight representative trackers on different sequences. We observe that these trackers drift to the background region or even lose the target object due to different challenging factors in the videos such as background clutter, scale variation, deformation, invisibility, motion blur, rotation, and low resolution. More efforts are desired to improve tracking robustness. 
	}
	\label{tracking_res}
 \vspace{-3mm}
\end{figure*}

\vspace{0.5em}
\noindent
\textbf{Attribute-based Performance.} To better analyze different trackers in handling different challenges, we further perform attribute-based evaluation on the ten  challenges. Fig.~\ref{fig:att} demonstrates the attribute-based evaluation results for the two most common attributes of SV and INV (based on the number of videos in each attribute) as well as the two difficult attributes of LR and IV using SUC. As displayed in Fig.~\ref{fig:att} (a), SeqTrack, MixFormer, and DropMAE achieves the best three results on SV/INV with 0.369/0.368, 0.367/0.365, and 0.347/0.347 scores in SUC, which is consistent with their performance in overall evaluation. As in Fig.~\ref{fig:att} (b), SeqTrack, MixFormer, and DropMAT are the best three trackers on IV. An interesting finding here is that, IV has shown to be easy for tracking in other study~\cite{fan2021tracklinic}. Nevertheless, our result shows difference. We argue, this is because the IV in our datasets usually occurs in  low-light condition with complicated background, which degrades tracking performance. This also shows, extreme illumination change still need to be carefully dealt with. LR is the most difficult challenge in VastTrack, because it may result in low-quality feature extraction. On this attribute, MixFormer, SeqTrack, and ROMTrack achieve the best three results with 0.236, 0.232, and 0.231 SUC scores. 

Due to the space limitation, we display more evaluation results and analysis such as the full attribute-based results in the \textbf{supplementary material}. Please kindly refer to it for details.

\vspace{0.5em}
\noindent
\textbf{Qualitative Evaluation} In addition to the quantitative evaluation, we further show qualitative results on VastTrack. Specifically, we demonstrate visualizations of eight representative trackers, including SeqTrack, MixFormer, RTS, OSTrack, SwinTrack, TransT, SimaRPN++, and SiamFC in different attributes such as \emph{scale variation}, \emph{deformation}, \emph{rotation}, \emph{aspect ratio change}, \emph{background clutter}, \emph{invisibility}, \emph{blur}, \emph{fast motion}, and \emph{low resolution} in Fig.~\ref{tracking_res}. As displayed Fig.~\ref{tracking_res}, we can observe that, although the trackers can deal with some challenges in the video sequences, they may still fail in more complicated scenarios where multiple challenges occur simultaneously, which indicates that more efforts are desired to improve existing approaches towards universal visual tracking.


\subsection{Comparison to Other Tracking Benchmarks}

In comparison with existing tracking datasets, VastTrack is more challenging due to the requirement of tracking object from more classes (in test). We present a comparison of VastTrack and other tracking datasets including TrackingNet~\cite{muller2018trackingnet} and LaSOT~\cite{fan2019lasot}. Note that, GOT-10k~\cite{huang2019got} is not compared because it adopts different metrics. Tab.~\ref{tab:datasetcomp} reports the results of top 15 trackers on VastTrack and their results on TrackingNet and LaSOT using SUC. From the Tab.~\ref{tab:datasetcomp}, we can clearly observe that, all compared approaches have a heavy performance drop on VastTrack. For instance, SeqTrack, the best tracker on VastTrack, achieves high SUC scores of 0.855 and 0.725 on TrackingNet and LaSOT, while degrades with 0.396 on our VastTrack with 0.459 and 0.329 drops. OSTrack drops from

\renewcommand\arraystretch{1.2}
\setlength{\columnsep}{10pt}%
\begin{wraptable}{r}{0.5\textwidth}\scriptsize
\vspace{-4mm}
\centering
\caption{Comparison of VastTrack to existing datasets using success score.}
\scalebox{0.95}{
 \tabcolsep=0.05cm
\begin{tabular}{@{}rccc@{}}
		\Xhline{2\arrayrulewidth}
		\multirow{2}[2]{*}{\textbf{Methods}} & \multicolumn{3}{c}{\textbf{Success Score}} \\
		\cmidrule{2-4}          & \tabincell{c}{TrackingNet\\\cite{muller2018trackingnet}} & \tabincell{c}{LaSOT\\\cite{fan2019lasot}} & \tabincell{c}{VastTrack \\ (Ours)} \\
		\hline\hline
		SeqTrack~\cite{chen2023seqtrack} & 0.855  & 0.725  & 0.396  \\
		MixFormer~\cite{cui2022mixformer} & 0.854  & 0.724  & 0.395  \\
		DropMAE~\cite{wu2023dropmae} & 0.841  & 0.718  & 0.375  \\
		ROMTrack~\cite{cai2023robust} & 0.841  & 0.714  & 0.370  \\
		GRM~\cite{gao2023generalized}   & 0.840  & 0.699  & 0.363  \\
		ARTrack~\cite{wei2023autoregressive} & 0.843  & 0.708  & 0.356  \\
		RTS~\cite{paul2022robust}   & 0.816  & 0.697  & 0.355  \\
		MixFormerV2~\cite{cui2023mixformerv2} & 0.834  & 0.706  & 0.352  \\
		ToMP~\cite{mayer2022transforming}  & 0.815  & 0.685  & 0.349  \\
		SimTrack~\cite{chen2022backbone} & 0.834  & 0.705  & 0.344  \\
		OSTrack~\cite{ye2022joint} & 0.839  & 0.711  & 0.336  \\
		STARK~\cite{yan2021learning} & 0.820  & 0.671  & 0.334  \\
		SwinTrack~\cite{lin2022swintrack}& 0.811  & 0.672  & 0.330  \\
		TrSiam~\cite{wang2021transformer} & 0.781  & 0.624  & 0.326  \\
		PrDiMP~\cite{danelljan2020probabilistic} & 0.758  & 0.598  & 0.310  \\
		\Xhline{2\arrayrulewidth}
	\end{tabular}}
	\label{tab:datasetcomp}
\vspace{-10pt}
\end{wraptable}
\noindent
0.839 and 0.711 SCU scores on TrackingNet and LaSOT to 0.336 on VastTrack. Likewise, other trackers suffer similar drops, which shows the challenge for current trackers in localizing object from more broad categories and there is still a long way for universal object tracking. 

In addition, we observe an interesting observation about the relative performance of trackers from Tab.~\ref{tab:datasetcomp}. Specifically, from Tab.~\ref{tab:datasetcomp}, we see that a few tracker such as OSTrack and SimTrack that perform better on LaSOT may perform relatively worse than other approaches like GRM, RTS, and ToMP on VastTrack. We argue that the possible reason is the abilities of different trackers in dealing with overfitting for object categories, which shows the need of more diverse video sequences with different classes in learning more general tracking systems

\subsection{Retraining Experiments with VastTrack}

\renewcommand\arraystretch{1.2}
\setlength{\columnsep}{10pt}%
\begin{wraptable}{r}{0.54\textwidth}\scriptsize
\vspace{-15pt}
\centering
\vspace{-5mm}
\caption{Further training with VastTrack.}
\scalebox{0.87}{
 \tabcolsep=0.03cm
\begin{tabular}{@{}rcccc@{}}
    \Xhline{2\arrayrulewidth}
          & \multicolumn{2}{c}{SiamRPN++~\cite{li2019siamrpn++}} & \multicolumn{2}{c}{OSTrack~\cite{ye2022joint}} \\
          \cmidrule(l){2-3} \cmidrule(l){4-5}
          & \tabincell{c}{SUC w/o \\retraining} & \tabincell{c}{SUC w/\\retraining} & \tabincell{c}{SUC w/o\\retraining} & \tabincell{c}{SUC w/\\retraining} \\
    \hline\hline
    VastTrack & 0.281 & 0.298 ($\uparrow$1.7\%) & 0.336 & 0.362 ($\uparrow$2.6\%) \\
    LaSOT & 0.496 & 0.528 ($\uparrow$3.2\%) & 0.711 & 0.722 ($\uparrow$1.1\%) \\
    \Xhline{2\arrayrulewidth}
    \end{tabular}}
	\label{tab:train}
\vspace{-10pt}
\end{wraptable}
To reveal the effectiveness of VastTrack in improving existing methods, we further train two trackers, consisting of SiamRPN++~\cite{li2019siamrpn++} and OSTrack~\cite{ye2022joint}, on the training set of VastTrack. Tab.~\ref{tab:train} shows the results. As shown in Tab.~\ref{tab:train}, we can clearly see that, after further training, the SUC score of SiamRPN++ is clearly improved from 0.281 to 0.298 with performance gains of 1.7\% on VastTrack and from 0.496 to 0.528 with 3.2\% gains on LaSOT. Besides, for OSTrack, the SUC score is boosted from 0.336 to 0.362 with 2.6\% improvements on VastTrack and from 0.711 to 0.722 with 1.1\% gains on LaSOT. Note that, OSTrack is already strong on LaSOT but still enhanced using VastTrack. All these experiments validate the effectiveness of VastTrack in improving tracking performance.

\section{Conclusion}
\label{con}

In this paper, we propose a large-scale benchmark, VastTrack, to facilitate the development of more general and universal tracking systems. To this goal, VastTrack contains abundant object classes and videos. Specifically, it covers 2,115 object categories, and 50,610 video sequences with 4.2 million frames. To the best of our knowledge, VastTrack is, to date, the largest tracking benchmark in terms of class diversity and the video number, which could potentially benefit more robust and general visual tracking. Furthermore, VastTrack provides rich annotations of both bounding box and language specification, which enables learning of both vision-only and vision-language tracking. To ensure high quality, each video in VastTrack is manually annotated with multi-round careful inspection and refinement. We extensively evaluate 25 representative trackers to analyze VastTrack and to offer baselines for comparison. By releasing VastTrack, we expect to provide the community a cornerstone dataset for developing more general and universal visual object tracking and facilitating its applications.

\section*{Supplementary Material}

\noindent
In the supplementary material, we demonstrate additional details of VastTrack and more experimental results, as follows,

\vspace{-0.5em}
\begin{itemize}
	\setlength{\itemsep}{1pt}
	\setlength{\parsep}{1pt}
	\setlength{\parskip}{1pt}
	
	\item[] \textbf{S1. Details of Object Classes} \\ We show the detailed object classes in VastTrack and number of sequences in each category.
	
	\item[] \textbf{S2. Statistics of Sequence Length and Annotation Boxes} \\ We present important statistics on the sequence length and annotation boxes of VastTrack for better understanding its features.
	
	\item[] \textbf{S3. Full Results of Attribute-based Evaluation} \\ We show the full attribute-based evaluation results for all evaluated trackers in precision (PRE), normalized precision (NPRE), and success (SUC).

    \item[] \textbf{S4. Additional Discussion} \\ We include brief additional discussion on analysis based on classes.
	
\end{itemize}

\begin{figure}[!t]
	\centering
        \includegraphics[width=\linewidth]{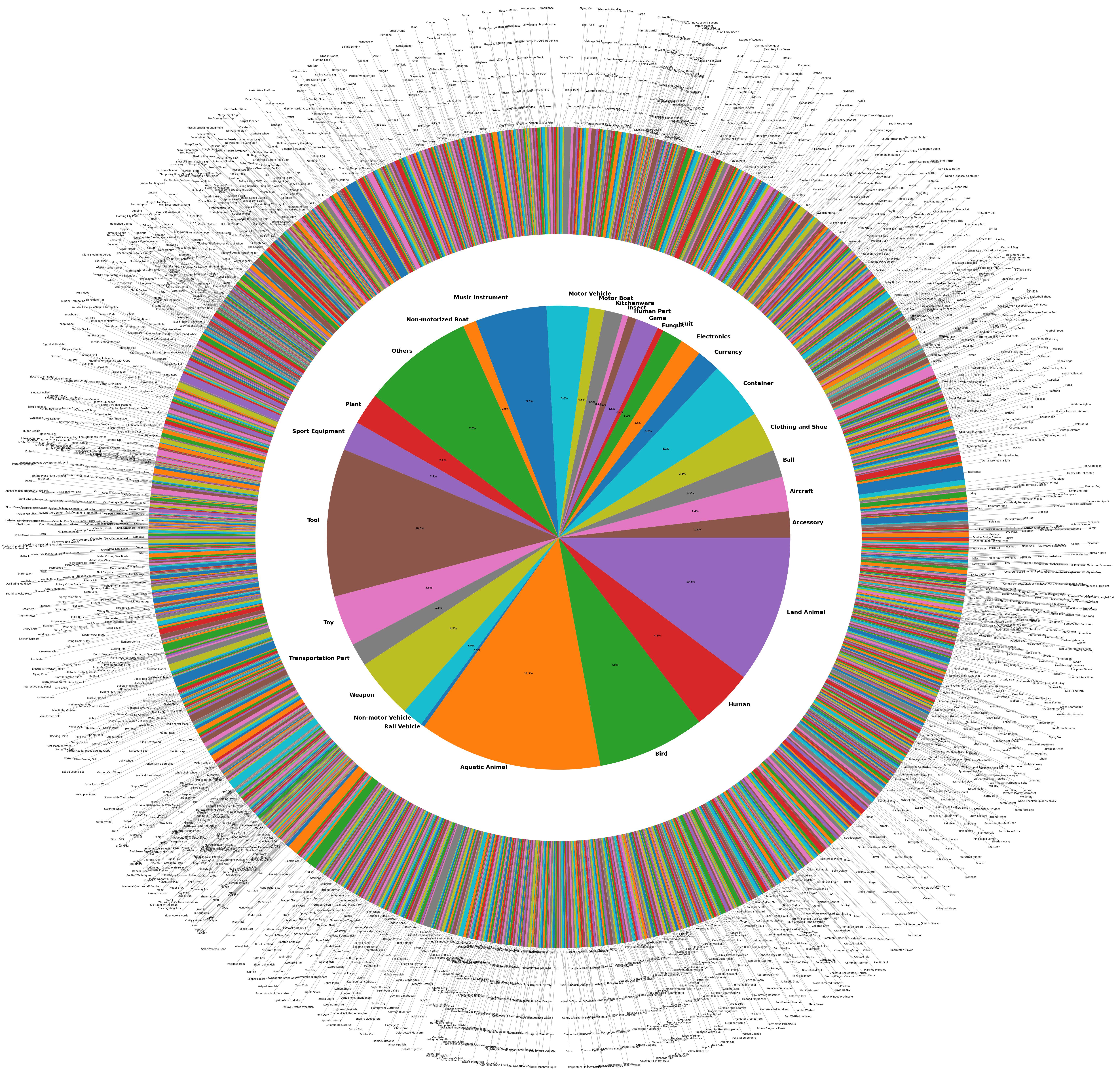}
	\caption{Category organization of VastTrack. Please zoom in.}
	\label{fig:class}
\end{figure}

\subsection*{S1 \;  Details of Object Classes}

VastTrack covers 2,115 classes, aiming to facilitate the development of universal and general object tracking. We collect these categories in a hierarchical way. In specific, we first collect 30 coarse classes, including ``\emph{Human}'', ``\emph{Human Part}'', ``\emph{Land Animal}'', ``\emph{Aquatic Animal}'', ``\emph{Bird}'', ``\emph{Motor Vehicle}'', ``\emph{Non-motor Vehicle}'', ``\emph{Motor Boat}'', ``\emph{Non-motorized Boat}'', ``\emph{Accessory}'', ``\emph{Aircraft}'', ``\emph{Ball}'', ``\emph{Clothing and Shoe}'', ``\emph{Container}'', ``\emph{Currency}'', ``\emph{Electronics}'', ``\emph{Fruit}'', ``\emph{Kitchenware}'', ``\emph{Game}'', ``\emph{Insect}'', ``\emph{Fungus}'', ``\emph{Music Instrument}'', ``\emph{Plant}'', ``\emph{Sport Equipment}'', ``\emph{Tool}'', ``\emph{Toy}'', ``\emph{Weapon}'', ``\emph{Transportation Part}'', ``\emph{Rail Vehicle}'',  and ``\emph{Others}''. Then, we further gather 2,115 fine object categories in each coarse classes. All these fine categories are verified by the expert to ensure that the sequences in this class are suitable for tracking. Fig.~\ref{fig:class} displays the category organization in VastTrack. In order to enable better understanding of VastTrack, below we present each object category and the number of sequences in the format of ``\emph{Class} (\# videos)'', \eg, ``\emph{Cow} (35)'', that means object category of ``Cow'' with 35 videos in alphabetical order:

\noindent\scriptsize{\centering{\textbf{\normalsize{A (107 classes)}}}\\\emph{Ardwolf} (18), \emph{Accessory Box} (21), \emph{Accordion} (33), \emph{Acipenser Sinensis} (13), \emph{Acrobat} (35), \emph{Actinomycetes} (4), \emph{Activity Wall} (3), \emph{Actor} (82), \emph{Adhesive Tape} (58), \emph{Adjustable Ladder} (3), \emph{Adjustable Wrench} (10), \emph{Aerial Drones In Flight} (50), \emph{Aerial Silk Performers} (48), \emph{Aerial Work Platform} (34), \emph{Afghan Hound} (20), \emph{Afm} (20), \emph{African Skimmer} (4), \emph{Aim} (3), \emph{Aims} (3), \emph{Aiptasia Eating Filefish} (35), \emph{Air Ambulance} (31), \emph{Air Hockey} (30), \emph{Air Swimmers} (119), \emph{Aircraft Carrier} (33), \emph{Airedale Terrier} (3), \emph{Airline Stewardess} (15), \emph{Airplane Model} (141), \emph{Airport Vehicle} (25), \emph{Airportshuttle} (27), \emph{Airship} (35), \emph{Ais74} (3), \emph{Ak100} (3), \emph{Ak102} (3), \emph{Ak103} (3), \emph{Ak104} (3), \emph{Ak105} (3), \emph{Ak47} (3), \emph{Ak74M} (3), \emph{Akm} (3), \emph{Akm63} (3), \emph{Akms} (3), \emph{Aks74} (3), \emph{Aks74U} (3), \emph{Alaskan Malamute} (3), \emph{Alligator} (31), \emph{Almond} (5), \emph{Aloe Vera} (2), \emph{Aloe Vera Cactus} (7), \emph{Alpaca} (20), \emph{Ambulance} (35), \emph{Amd65} (3), \emph{American Bulldog} (20), \emph{American Cocker Spaniel} (3), \emph{American Eskimo Dog} (3), \emph{Amphibious Vehicle} (21), \emph{Amulet} (26), \emph{An94} (3), \emph{Anchor Winch Wheel} (33), \emph{Andean Cock-Of-The-Rock} (18), \emph{Andinoacara Latifrons} (36), \emph{Angelfish} (35), \emph{Angle Gauge} (33), \emph{Angle Grinder} (2), \emph{Anhinga} (100), \emph{Ankle Socks} (3), \emph{Ankle Boots} (15), \emph{Ankylosaurus} (9), \emph{Annona} (6), \emph{Antarctic Shag} (5), \emph{Antarctic Tern} (10), \emph{Antelope} (44), \emph{Anti-Radiation Clothing} (3), \emph{Apistogramma Eremnopyge} (47), \emph{Apollo Shark} (8), \emph{Apothecary Box} (20), \emph{Ar15} (3), \emph{Arctic Hare} (24), \emph{Arctic Warbler} (31), \emph{Arctic Wolf} (21), \emph{Arena Of Valor} (32), \emph{Argentine Peso} (15), \emph{Arm Warmers} (3), \emph{Armadillo} (33), \emph{Armored Searobin} (7), \emph{Armoured Personnel Carrier} (11), \emph{Art Supply Box} (34), \emph{Arterial Line Kit} (1), \emph{As50} (1), \emph{Asian Arowana} (28), \emph{Asian Lady Beetle} (30), \emph{Atlantic Puffin} (44), \emph{Atlatl} (54), \emph{Audio} (26), \emph{Audio Box} (7), \emph{Audio Equipment Caster} (10), \emph{Auger} (3), \emph{Auricularia Auricula} (35), \emph{Australian Cattle Dog} (2), \emph{Australian Dollar} (15), \emph{Australian Pratincole} (10), \emph{Autoinjector} (3), \emph{Aviator Glasses} (3), \emph{Avocado} (14), \emph{Awaous Flavus} (14), \emph{Azaras Night Monkey} (3), \emph{Azarass Capuchin} (2), \emph{Azure-Winged Magpie} (31).} \\

\noindent\scriptsize{\centering{\textbf{\normalsize{B (226 classes)}}}\\\emph{Baboon} (57), \emph{Baby Bottle} (26), \emph{Backhoe Loader} (53), \emph{Backpack} (55), \emph{Badis Ruber} (8), \emph{Badminton} (30), \emph{Badminton Player} (69), \emph{Badminton Racket} (30), \emph{Bagh Nakh} (2), \emph{Baglama} (32), \emph{Bala Shark} (8), \emph{Balalaika} (93), \emph{Balance Pods} (10), \emph{Balance Wheel} (34), \emph{Balancing Machine} (88), \emph{Bald Uakari} (4), \emph{Ballerina Pumps} (74), \emph{Ballet Dancer} (39), \emph{Ballpoint Pen} (47), \emph{Ballpoint Pen Fish} (32), \emph{Bamboo Raft} (94), \emph{Bamboo Rat} (32), \emph{Banana} (29), \emph{Band Saw} (20), \emph{Bandage} (28), \emph{Banggai Cardinalfish} (35), \emph{Banjo} (29), \emph{Bank Vole} (35), \emph{Barbadian Dollar} (58), \emph{Barbat} (33), \emph{Bare-Eared Squirrel Monkey} (5), \emph{Barge} (77), \emph{Barley} (5), \emph{Barn Swallow} (30), \emph{Barred Cuckoo-Dove} (35), \emph{Barrel Cactus} (10), \emph{Barrel Jellyfish} (20), \emph{Barrel Wheel} (68), \emph{Baseball} (30), \emph{Baseball Bat Swinging} (14), \emph{Baseball Cap} (3), \emph{Basketball} (30), \emph{Basketball Player} (29), \emph{Basketball Shoes} (55), \emph{Basking Shark} (11), \emph{Bass Clarinet} (11), \emph{Bass Drum} (35), \emph{Bass Saxophone} (63), \emph{Basset Hound} (3), \emph{Bat} (58), \emph{Baton} (33), \emph{Baton Twirling} (18), \emph{Batteries Box} (53), \emph{Battleaxe} (6), \emph{Bayonet} (33), \emph{Bcm4 Recce-14 Mcmr} (3), \emph{Beach Pants} (4), \emph{Beach Volleyball} (33), \emph{Bean Bag Toss Game} (131), \emph{Beanie Hat} (30), \emph{Bearded Axe} (32), \emph{Bearded Collie} (3), \emph{Beaver} (40), \emph{Bedlington Terrier} (3), \emph{Bee} (29), \emph{Beer} (29), \emph{Beer Bottle} (93), \emph{Belgian Malinois} (3), \emph{Belizean Dollar} (11), \emph{Belly Dancer} (28), \emph{Belt} (4), \emph{Belt Bag} (61), \emph{Beluga Whale} (32), \emph{Bench Grinder} (3), \emph{Bench Swing} (3), \emph{Bench Vise} (20), \emph{Benelli Lupo} (3), \emph{Beretta Holding  M9A3} (11), \emph{Beretta Holding 90Two} (4), \emph{Beretta Holding 92F} (7), \emph{Beretta Holding Apx} (6), \emph{Bergara Bmr} (3), \emph{Berthier Mle 1890} (3), \emph{Betta Fish} (35), \emph{Bhutan Takin} (28), \emph{Bible} (63), \emph{Bichon Frise} (19), \emph{Bicycle Lane Sign} (83), \emph{Bicycle Race Bike} (1), \emph{Bifocal Glasses} (2), \emph{Bigfin Reef Squid} (35), \emph{Bike} (64), \emph{Billiards} (30), \emph{Binturong} (21), \emph{Bird Wrasse} (35), \emph{Birgus Latro} (11), \emph{Black Ant} (13), \emph{Black Bean} (8), \emph{Black Bearded Saki} (4), \emph{Black Caiman} (13), \emph{Black Guillemot} (5), \emph{Black Molly} (11), \emph{Black Muntjac} (33), \emph{Black Musk Deer} (4), \emph{Black Panther} (25), \emph{Black Skimmer} (5), \emph{Black Swan} (34), \emph{Black-Bellied Tern} (10), \emph{Black-Fronted Titi Monkey} (1), \emph{Black-Headed Gull} (60), \emph{Black-Legged Kittiwake} (4), \emph{Black-Necked Swan} (33), \emph{Black-Nest Swiftlet} (10), \emph{Black-Tailed Gull} (71), \emph{Black-Throated Bushtit} (33), \emph{Black-Winged Pratincole} (9), \emph{Blanket Octopus} (25), \emph{Bleach Bottle} (32), \emph{Blond Capuchin} (3), \emph{Blood Administration Set} (3), \emph{Blood Collection Needle} (3), \emph{Blood Collection Set} (10), \emph{Blood Collection Tube} (10), \emph{Blood Draw Syringe} (5), \emph{Blood Transfer Device} (7), \emph{Blood Tube Holder} (10), \emph{Blue Blubber Jellyfish} (34), \emph{Blue Crab} (20), \emph{Blue Picardy Spaniel} (3), \emph{Blue Rock Thrush} (34), \emph{Blue Sea Dragon} (16), \emph{Blue Sheep} (34), \emph{Blue Star Leopard Wrasse} (35), \emph{Blue Whale} (8), \emph{Blue-And-White Flycatcher} (71), \emph{Blueberry} (19), \emph{Blue-Crowned Hanging-Parrot} (30), \emph{Blue-Footed Booby} (10), \emph{Blue-Ringed Octopus} (30), \emph{Bluethroat} (28), \emph{Bluetooth Speaker} (74), \emph{Blunt Cannula} (10), \emph{Blunt Fill Needle} (3), \emph{Bo Staff} (30), \emph{Bo Staff Techniques} (44), \emph{Boat Shoes} (15), \emph{Bobcat} (35), \emph{Bobsledder} (20), \emph{Bobtail Squid} (18), \emph{Bocce Ball} (29), \emph{Bocce Ball Set} (34), \emph{Bodhran} (35), \emph{Bodianus Sepiacaudus} (3), \emph{Body Wash Bottle} (34), \emph{Bolas} (10), \emph{Bolero Jacket} (3), \emph{Bolt Cutter} (9), \emph{Bonapartes Gull} (6), \emph{Bongos} (31), \emph{Bonobo} (31), \emph{Book Bag} (24), \emph{Boomerang} (30), \emph{Boots} (4), \emph{Border Collie} (2), \emph{Boston Terrier} (3), \emph{Bottle Cap} (38), \emph{Bottle Opener} (29), \emph{Bounce And Spin} (4), \emph{Bouncing Bumpers} (10), \emph{Bouncing Platforms} (1), \emph{Bow And Arrow} (6), \emph{Bowed Psaltery} (70), \emph{Bowl} (35), \emph{Boxer} (50), \emph{Boxer Dog} (3), \emph{Boxfish} (35), \emph{Bra} (8), \emph{Bracelet} (155), \emph{Brad Nailer} (20), \emph{Brahminy Blind Snake} (3), \emph{Bralette Top} (3), \emph{Brazil Nut} (1), \emph{Bread Machine} (29), \emph{Break Dancer} (95), \emph{Brick Tongs} (20), \emph{Bridge Ices Before Road Sign} (6), \emph{Briefcase} (16), \emph{British Shorthair Cat} (5), \emph{Brl} (28), \emph{Bronze-Winged Courser} (10), \emph{Broom} (33), \emph{Brothers In Arms} (30), \emph{Brown Bear} (32), \emph{Brown Booby} (10), \emph{Brown Howler} (5), \emph{Brown Noddy} (10), \emph{Brown Spider Monkey} (5), \emph{Brown-Flanked Bush Warbler} (19), \emph{Browning X-Bolt} (3), \emph{Brown-Mantled Tamarin} (3), \emph{Brush} (26), \emph{Brussels Griffon} (2), \emph{Bubble Machine} (22), \emph{Bubble Play Area} (1), \emph{Bubble Snail} (18), \emph{Bucket} (34), \emph{Bucket Backpack} (13), \emph{Buckwheat} (8), \emph{Buffy Saki} (3), \emph{Buffy-Headed Marmoset} (17), \emph{Bugle} (34), \emph{Bull Terrier} (3), \emph{Bulldozer} (34), \emph{Bullock Cart} (35), \emph{Bumper Boats} (10), \emph{Bumper Car} (6), \emph{Bungee Trampoline} (10), \emph{Bunker Tanker} (53), \emph{Bunny Ears Cactus} (6), \emph{Burmese Ferret-Badger} (20), \emph{Butis Butis Fish} (9), \emph{Butterfly} (29), \emph{Butterfly Needle} (3), \emph{Butterfly Sword} (16), \emph{Butterflyfish} (28), \emph{Buzuq} (29).}\\

\noindent\scriptsize{\centering{\textbf{\normalsize{C (208 classes)}}}\\\emph{Cable Cutter} (3), \emph{Cajon} (35), \emph{Cake Pan} (11), \emph{Calendar} (4), \emph{California Spangled Cat} (4), \emph{Call Of Duty} (10), \emph{Camel} (31), \emph{Camera Backpack} (21), \emph{Camera Wheel} (16), \emph{Camisole} (4), \emph{Camogie} (32), \emph{Can Opener} (20), \emph{Candy Box} (36), \emph{Candy Crab} (13), \emph{Canik Tp9} (9), \emph{Cannonball Jellyfish} (39), \emph{Cannula} (3), \emph{Cape} (3), \emph{Car Hubcap} (99), \emph{Carcano M1891} (2), \emph{Cardigan} (4), \emph{Cargo Pants} (2), \emph{Cargo Plane} (4), \emph{Cargo Ship} (52), \emph{Cargo Truck} (34), \emph{Carnation} (26), \emph{Carp} (32), \emph{Carpenters Flasher Wrasse} (11), \emph{Carpet Cleaner} (29), \emph{Carpet Shark} (17), \emph{Carriage} (209), \emph{Carrot} (3), \emph{Cart Caster Wheel} (42), \emph{Cashew} (14), \emph{Caspian Tern} (10), \emph{Cassins Auklet} (2), \emph{Cast Iron Skillet} (100), \emph{Castor Bean} (7), \emph{Cat} (45), \emph{Catamaran} (45), \emph{Catheter Insertion Tray} (1), \emph{Catheter Introducer} (2), \emph{Catheter Securement Device} (3), \emph{Catheter Tip Syringe} (3), \emph{Cattle Egret} (26), \emph{Caulking Gun} (10), \emph{Cavaquinho} (28), \emph{C-Clamp} (22), \emph{Celesta} (27), \emph{Cell} (6), \emph{Central American Spider Monkey} (4), \emph{Central Venous Catheter} (3), \emph{Cereal Box} (32), \emph{Chain Drive Sprocket} (79), \emph{Chain Sickle} (3), \emph{Chakram} (1), \emph{Chalk} (11), \emph{Chalk Line} (10), \emph{Chalkboard Eraser} (12), \emph{Chambered Nautilus} (24), \emph{Chandelier} (29), \emph{Channa Argus} (31), \emph{Channa Striata} (49), \emph{Charger Loading Lee-Metford} (3), \emph{Checkers} (30), \emph{Cheese Box} (15), \emph{Cheetah} (32), \emph{Chef Bag} (11), \emph{Cherry Shrimp} (35), \emph{Chestnut} (8), \emph{Chestnut-Bellied Rock Thrush} (21), \emph{Cheytacm200} (3), \emph{Chicken} (30), \emph{Chickpea} (7), \emph{Chiffon Skirt} (3), \emph{Chigiriki} (4), \emph{Chilean Skua} (5), \emph{Chimaera} (30), \emph{Chimpanzee} (35), \emph{Chinese Algae Eater} (35), \emph{Chinese Army Chess} (1), \emph{Chinese Bulbul} (65), \emph{Chinese Chess} (15), \emph{Chinese Chongqing Dog} (20), \emph{Chinese Cobra} (3), \emph{Chinese High-Fin Banded Shark} (34), \emph{Chinese Li Hua Cat} (5), \emph{Chinese White-Browed Bird Warbler} (21), \emph{Chitarra Battente} (35), \emph{Chive} (2), \emph{Chocolate Box} (24), \emph{Chop Saw} (1), \emph{Chopping Board} (18), \emph{Chow Chow} (16), \emph{Christmas Cactus} (8), \emph{Chrysanthemum} (22), \emph{Cicada} (20), \emph{Cicada Killer Wasp} (30), \emph{Cigar Box} (32), \emph{Cirrhilabrus Naokoae} (10), \emph{Civet} (80), \emph{Claret Cup Cactus} (1), \emph{Clarinet} (40), \emph{Clavichord} (1), \emph{Cleaning Cloth} (32), \emph{Cleaning Gloves} (45), \emph{Cleaning Paste} (34), \emph{Clear Tote} (40), \emph{Cleistocactus} (29), \emph{Clerk} (1), \emph{Climbing Boulders} (4), \emph{Climbing Dome} (3), \emph{Climbing Rope} (16), \emph{Clip} (6), \emph{Cloth} (47), \emph{Clothing Package Box} (31), \emph{Clown Killifish} (35), \emph{Clown Loach} (14), \emph{Clownfish} (32), \emph{Coast Guard Cutter} (21), \emph{Coat} (3), \emph{Cockatoo Squid} (2), \emph{Cocktails} (30), \emph{Cocoa Bean} (7), \emph{Coconut} (3), \emph{Coconut Octopus} (35), \emph{Coffee Bean} (7), \emph{Coffee Box} (12), \emph{Coffee Press} (14), \emph{Coffinfish} (18), \emph{Colander} (18), \emph{Cold Planer} (3), \emph{Collared Crow} (70), \emph{Collared Peccary} (49), \emph{Colombian Red Howler} (4), \emph{Colombian White-Faced Capuchin} (3), \emph{Colored Socks} (3), \emph{Colorimeter} (71), \emph{Command Conquer} (31), \emph{Common Goldeneye} (35), \emph{Common Kingfisher} (32), \emph{Common Moorhen} (44), \emph{Common Murre} (3), \emph{Common Redstart} (31), \emph{Common Treeshrew} (33), \emph{Common Woolly Monkey} (3), \emph{Commuter Bag} (40), \emph{Compass} (35), \emph{Computer Chair Caster Wheel} (29), \emph{Concrete Mixer Truck} (40), \emph{Concrete Pump Truck} (55), \emph{Concrete Spreader} (3), \emph{Conditioner Bottle} (5), \emph{Congas} (31), \emph{Congo Tetra} (35), \emph{Connector Caps} (3), \emph{Construction Ahead Sign} (16), \emph{Construction Workers} (48), \emph{Contrabassoon} (35), \emph{Convertible} (29), \emph{Conveyor Belt Wheel} (72), \emph{Coordinate Measuring Machine} (9), \emph{Coracle} (48), \emph{Coral Catshark} (31), \emph{Cordless Handheld Power Scrubber} (35), \emph{Cordless Screwdriver} (20), \emph{Cordless Skipping Rope Reviews} (4), \emph{Corn} (1), \emph{Cornet} (32), \emph{Cornflower} (33), \emph{Corsac Fox} (35), \emph{Corydoras} (35), \emph{Cosmetic Gift Box} (36), \emph{Cosmetics Case} (29), \emph{Cotton-Top Tamarin} (3), \emph{Cougar} (32), \emph{Courteney Stalking Rifle} (3), \emph{Cow} (35), \emph{Cow Shark} (5), \emph{Cownose Ray} (10), \emph{Crab-Plover} (10), \emph{Crane} (32), \emph{Crane Barge} (47), \emph{Crane Wheel} (10), \emph{Crayon} (27), \emph{Crested Auklet} (14), \emph{Crested Ibis} (5), \emph{Cricket} (4), \emph{Cricket Bat} (3), \emph{Croquet Set} (35), \emph{Cross Line Laser} (15), \emph{Crossbody Backpack} (49), \emph{Crossbow} (49), \emph{Crossbow Pistol} (32), \emph{Crowbar} (10), \emph{Cruise Ship} (47), \emph{Ctenopoma Acutirostre} (49), \emph{Cucumber} (2), \emph{Cufflinks} (18), \emph{Cup} (32), \emph{Cupping} (35), \emph{Curler} (20), \emph{Curling} (39), \emph{Curling Iron} (22), \emph{Cyclist} (50), \emph{Cz P11C} (10), \emph{Cz76} (10), \emph{Cz-Usa Model 557 Eclipse} (3).}\\

\noindent\scriptsize{\centering{\textbf{\normalsize{D (74 classes)}}}\\\emph{Dagger} (5), \emph{Dahlia} (33), \emph{Daisy} (31), \emph{Dalmatian} (3), \emph{Damselfly} (3), \emph{Danbau} (32), \emph{Dandelion Siphonophore} (7), \emph{Daniel Defense Delta 5 Pro} (3), \emph{Dartboard Set} (34), \emph{Daurian Hedgehog} (34), \emph{Deer Crossing Sign} (23), \emph{Depth Gauge} (35), \emph{Desk Lamp} (30), \emph{Detour Sign} (27), \emph{Dhole} (30), \emph{Dial Indicator} (105), \emph{Dialysis Needle} (1), \emph{Diamond Drill} (3), \emph{Diamond Tail Flasher Wrasse} (33), \emph{Dice} (7), \emph{Didgeridoo} (35), \emph{Digging Toys} (3), \emph{Digital Multi-Meter} (34), \emph{Digital Piano} (27), \emph{Dilruba} (35), \emph{Dinnerware} (4), \emph{Dione Ratsnake} (3), \emph{Disc Swing} (10), \emph{Discus Fish} (35), \emph{Disinfecting Cotton Balls} (31), \emph{Diver} (33), \emph{Diving Support Vessel} (36), \emph{Djembe} (30), \emph{Doberman Pinscher} (3), \emph{Doctor} (1), \emph{Document Box} (8), \emph{Dogfish Shark} (15), \emph{Dolly Wheel} (58), \emph{Dolphin Gull} (5), \emph{Dominican Peso} (27), \emph{Domra} (35), \emph{Dota 2} (81), \emph{Double Bass} (35), \emph{Double Bridge Glasses} (10), \emph{Doweling Jig} (19), \emph{Down Jacket} (3), \emph{Dragon Dance} (35), \emph{Dragon Dance Staff} (30), \emph{Dragon Wrasse} (64), \emph{Drainage Truck} (35), \emph{Dress} (4), \emph{Drift Boat} (47), \emph{Drillship} (46), \emph{Drop Slide} (3), \emph{Drum Set} (13), \emph{Drum Spinner} (7), \emph{Drum Stick} (30), \emph{Drywall Stilts} (20), \emph{Dsc} (11), \emph{Duck} (49), \emph{Duct Tape} (1), \emph{Duduk} (35), \emph{Duffle Backpack} (4), \emph{Dulcimer} (5), \emph{Dumbo Octopus} (35), \emph{Dune} (10), \emph{Durian} (47), \emph{Dusky Shark} (10), \emph{Dust Mitt} (51), \emph{Dust Mop} (29), \emph{Duster} (30), \emph{Dustpan} (33), \emph{Dwarf Chin Cactus} (4), \emph{Dwarf Gourami} (44).}\\

\noindent\scriptsize{\centering{\textbf{\normalsize{E (65 classes)}}}\\\emph{Ear} (40), \emph{Earrings} (94), \emph{Eastern Caribbean Dollar} (12), \emph{Ecuadorian Sucre} (32), \emph{Edamame} (10), \emph{Edelweiss} (5), \emph{Egg} (15), \emph{Egg Slicer} (10), \emph{Eggbeater} (9), \emph{Eggplant} (3), \emph{Elden Ring} (35), \emph{Electric Air Blower} (48), \emph{Electric Air Hockey Table} (8), \emph{Electric Air Purifier} (30), \emph{Electric Animal Rides} (10), \emph{Electric Broom} (98), \emph{Electric Car} (38), \emph{Electric Drill Driver} (35), \emph{Electric Hedge Trimmer} (68), \emph{Electric Lawn Edger} (29), \emph{Electric Mixer} (6), \emph{Electric Piano} (34), \emph{Electric Power Scrubber Brush} (30), \emph{Electric Power Washer Foam Cannon} (32), \emph{Electric Ray} (10), \emph{Electric Scooters} (10), \emph{Electric Scrubber Machine} (28), \emph{Electric Squeegee} (41), \emph{Electric Ultrasonic Toothbrush} (69), \emph{Electronic Product Box} (61), \emph{Electronic Scale} (30), \emph{Elegant Tern} (10), \emph{Elephant} (37), \emph{Elevator Pulley} (7), \emph{Elliptical Machine Flywheel} (37), \emph{Emperor Tamarin} (2), \emph{Endlers Livebearers} (8), \emph{English Horn} (31), \emph{Enterprise} (19), \emph{Epidural Kit} (4), \emph{Epinephelus Marginatus} (44), \emph{Eraser} (21), \emph{Escrima Sticks} (8), \emph{Espadrilles} (34), \emph{Euphonium} (20), \emph{Eurasian Badger} (19), \emph{Eurasian Curlew} (33), \emph{Eurasian Eagle-Owl} (33), \emph{Eurasian Hoopoe} (5), \emph{Eurasian Sparrowhawk} (30), \emph{Eurasian Tree Sparrow} (33), \emph{Euro} (39), \emph{European Bee-Eaters} (28), \emph{European Otter} (44), \emph{European Polecat} (18), \emph{European Robin} (21), \emph{Exercise Resistance Band Wheel} (93), \emph{Exercise Wheel} (48), \emph{Exit Sign} (30), \emph{Exotic Shorthair Cat} (5), \emph{Extension Set} (3), \emph{Extension Tubing} (3), \emph{Eye Mask} (3), \emph{Eyebrow} (19), \emph{Eyes} (45).}\\

\noindent\scriptsize{\centering{\textbf{\normalsize{F (97 classes)}}}\\\emph{F90} (3), \emph{Face} (94), \emph{Fake Collar} (4), \emph{Falcata} (16), \emph{Falcated Duck} (31), \emph{Falling Rocks Sign} (8), \emph{Fallow Deer} (30), \emph{Famas} (3), \emph{Fan} (132), \emph{Fan Wh} (7), \emph{Farm Tractor Wheel} (50), \emph{Farmer} (99), \emph{Fashion Glasses} (2), \emph{Feder} (30), \emph{Fedora Hat} (46), \emph{Fencer} (67), \emph{Fennec Fox} (49), \emph{Feral Pigeons} (24), \emph{Ferris Wheel} (32), \emph{Ferris Wheel Axle} (55), \emph{Ferris Wheel Support Structure} (51), \emph{Ferrule Fitting} (75), \emph{Ferry} (11), \emph{Fiddler Crab} (4), \emph{Fiddler Ray} (60), \emph{Fidget Spinner} (9), \emph{Field Hockey} (29), \emph{Fighter Jet} (56), \emph{Figt} (6), \emph{Filipino Martial Arts Stick And Knife Techniques} (30), \emph{Finless Porpoise} (13), \emph{Finnish Mark} (2), \emph{Fire Station Sign} (9), \emph{Fire Truck} (40), \emph{Fireboat} (20), \emph{Firefighters} (35), \emph{Firefighting Aircraft} (34), \emph{Firemouth Cichlid} (21), \emph{Fish Tank} (20), \emph{Fishermen} (33), \emph{Fishing Reel Spool} (74), \emph{Fishing Vessel} (46), \emph{Fishnet Stockings} (3), \emph{Fistball} (32), \emph{Fistula Needle} (2), \emph{Fitness Roller} (73), \emph{Flamboyant Cuttlefish} (158), \emph{Flame Jelly} (9), \emph{Flammulina Velutipes} (45), \emph{Flapjack Octopus} (17), \emph{Flea} (4), \emph{Floating Board} (35), \emph{Floating Crane} (32), \emph{Floating Lily Pads} (8), \emph{Floating Logs} (5), \emph{Floating Stepping Stones} (1), \emph{Floatplane} (59), \emph{Floor Lamp} (6), \emph{Floor Squeegee} (5), \emph{Floral Pants} (4), \emph{Flounder} (32), \emph{Fluid Warming Set} (9), \emph{Flush Syringe} (1), \emph{Flute} (35), \emph{Flying Ball} (25), \emph{Flying Car} (54), \emph{Flying Fish} (8), \emph{Flying Fox} (15), \emph{Flying Gurnard} (35), \emph{Flying Kites} (36), \emph{Flying Lemurs} (14), \emph{Fn M1935} (10), \emph{Fn509} (10), \emph{Fn57} (9), \emph{Folk Dancer} (79), \emph{Food Tongs} (29), \emph{Food-Print Shirt} (3), \emph{Foosball} (96), \emph{Foot} (42), \emph{Football} (15), \emph{Football Boots} (48), \emph{Force Gauge} (14), \emph{Fork} (10), \emph{Fork-Tailed Sunbird} (40), \emph{Formula Two} (49), \emph{Forsters Tern} (9), \emph{Foxtail Millet} (3), \emph{Franchi} (3), \emph{Fried Egg Jellyfish} (35), \emph{Frisbee} (148), \emph{Frog} (35), \emph{Fruit Bat} (35), \emph{Fruit Box} (1), \emph{Fruit Fly} (5), \emph{Frying Pan} (10), \emph{Fur Coat} (4), \emph{Futsal} (32).}\\

\noindent\scriptsize{\centering{\textbf{\normalsize{G (108 classes)}}}\\\emph{G3} (3), \emph{G36} (2), \emph{Gaff Rig} (35), \emph{Game Box} (1), \emph{Ganoderma} (35), \emph{Gansu Zokor} (6), \emph{Garbage Bag} (25), \emph{Garbage Can} (33), \emph{Garbage Truck} (140), \emph{Garden Cart Wheel} (96), \emph{Garden Spider} (5), \emph{Garden Warbler} (31), \emph{Garlic} (4), \emph{Garlic Press} (10), \emph{Garment Bag} (8), \emph{Gas Detector} (83), \emph{Gastraphetes} (9), \emph{Gaudy Clown Crab} (2), \emph{Gavialis Gangeticus} (10), \emph{Gc} (18), \emph{Gemore} (29), \emph{Geoffroys Tamarin} (2), \emph{German Blue Ram} (35), \emph{Ghost Crab} (35), \emph{Ghost Pipefish} (38), \emph{Giant Anteater} (48), \emph{Giant Armadillo} (45), \emph{Giant Australian Cuttlefish} (55), \emph{Giant Inflatable Slides} (4), \emph{Giant Otter} (33), \emph{Giant Panda} (34), \emph{Giant Saguaro Cactus} (35), \emph{Giant Twister Game} (2), \emph{Gibbon} (33), \emph{Gift Box} (40), \emph{Giraffe} (29), \emph{Giri-Giri} (40), \emph{Glaive} (2), \emph{Glass Catfish} (16), \emph{Glasses} (4), \emph{Glassware} (10), \emph{Glider} (43), \emph{Glock G17} (6), \emph{Glock G19X} (58), \emph{Glock G23} (50), \emph{Glock G45} (50), \emph{Glockenspiel} (15), \emph{Gloomy Nudibranch} (34), \emph{Gloomy Octopus} (38), \emph{Glue} (27), \emph{Go} (2), \emph{Goatfish} (35), \emph{Goblin Shark} (2), \emph{Goeldis Marmoset} (3), \emph{Go-Karts} (10), \emph{Gold-Dotted Flatworm} (26), \emph{Golden Barrel Cactus} (32), \emph{Golden Bush-Robin} (14), \emph{Golden Lion Tamarin} (3), \emph{Golden Pheasant} (35), \emph{Golden Eagle} (19), \emph{Golden-Bellied Capuchin} (5), \emph{Golden-Handed Tamarin} (2), \emph{Golden-Mantled Tamarin} (3), \emph{Golf} (30), \emph{Golf Player} (1), \emph{Goliath Tigerfish} (64), \emph{Googly-Eyed Stubby Squid} (10), \emph{Gorilla} (39), \emph{Grainsorghum} (3), \emph{Grapefruit} (88), \emph{Grapsus Grapsus} (45), \emph{Grater} (9), \emph{Gray Fox} (34), \emph{Gray Leaf Monkey} (7), \emph{Gray Whale} (25), \emph{Great Bustard} (28), \emph{Great White Shark} (16), \emph{Great Egret} (25), \emph{Greater Crested Tern} (9), \emph{Green Cochoa} (28), \emph{Green Leafhopper} (5), \emph{Green Turtle} (88), \emph{Greenland Shark} (8), \emph{Grevys Zebra} (47), \emph{Grey Jay} (32), \emph{Grey Seal} (18), \emph{Grey Treepie} (34), \emph{Grey-Capped Greenfinch} (35), \emph{Grey-Crowned Warbler} (12), \emph{Grill Pan} (23), \emph{Grizzly Bear} (30), \emph{Ground Trampoline} (2), \emph{Grouper} (49), \emph{Grubfish} (18), \emph{Guatemalan Quetzal} (89), \emph{Guianan Squirrel Monkey} (4), \emph{Guinea Pig} (26), \emph{Guisarme} (3), \emph{Guitar} (41), \emph{Gull-Billed Tern} (10), \emph{Gulper Eel} (13), \emph{Gun Scope} (110), \emph{Gymnast} (68), \emph{Gymnocalycium} (49), \emph{Gypsy Moth} (2), \emph{Gyro Spinner} (10), \emph{Gyroscope} (79).}\\

\noindent\scriptsize{\centering{\textbf{\normalsize{H (101 classes)}}}\\\emph{Hacksaw} (6), \emph{Hair Accessory Box} (11), \emph{Hair Dryer} (16), \emph{Hair Gel Bottle} (27), \emph{Hairpin} (20), \emph{Haitian Gourde} (53), \emph{Half Banded Flasher Wrasse} (5), \emph{Half-Life} (5), \emph{Half-Moon Spear} (2), \emph{Halichoeres Leucoxanthus} (15), \emph{Halloween Crab} (18), \emph{Halo} (10), \emph{Hamidashi} (7), \emph{Hammer Drill} (2), \emph{Hammock Swing} (3), \emph{Hand} (4), \emph{Hand Pedal Bike} (13), \emph{Handball Player} (39), \emph{Handheld Game Console} (50), \emph{Handkerchief} (4), \emph{Hand-Powered Ferris Wheel} (5), \emph{Hardness Tester} (26), \emph{Hardware Box} (1), \emph{Hare} (35), \emph{Harlequin Rasboras} (11), \emph{Harlequin Shark} (5), \emph{Harlequin Shrimp} (31), \emph{Harlequin Sweetlips} (35), \emph{Harlequin Tuskfish} (35), \emph{Harmonica} (14), \emph{Harp} (17), \emph{Harp Guitar} (34), \emph{Harpoon} (25), \emph{Harpsichord} (31), \emph{Harvester} (31), \emph{Hat} (33), \emph{Hat Storage Box} (2), \emph{Hawkthorn} (1), \emph{Hazelnut} (9), \emph{Head} (83), \emph{Headband} (4), \emph{Heavy-Lift Helicopter} (56), \emph{Hedgehog} (32), \emph{Hedgehog Cactus} (10), \emph{Height Gauge} (108), \emph{Helicopter} (34), \emph{Helicopter Rotor} (32), \emph{Helmet} (186), \emph{Helmet Jellyfish} (18), \emph{Helter Skelter Slide} (1), \emph{Hemostasis Valve} (3), \emph{Heparin Lock} (2), \emph{Hericium Erinaceus} (33), \emph{Heroes Of The Storm} (79), \emph{Hibiscus} (31), \emph{High Heels} (4), \emph{High-Waisted Pants} (4), \emph{Hiking Boots} (33), \emph{Hill Prinia} (30), \emph{Himalayan Monal} (32), \emph{Hippopotamus} (35), \emph{Historical Reenactments With Blades} (30), \emph{Hk G28} (3), \emph{Hk Mk23 Mod 0} (4), \emph{Hk Usp45} (5), \emph{Hk Vp9} (4), \emph{Hk416} (3), \emph{Hockey Player} (24), \emph{Hog Badger} (32), \emph{Holacanthus Tricolor} (46), \emph{Honey Bottle} (12), \emph{Hooded Merganser} (44), \emph{Hoodie} (4), \emph{Hopper Balls} (10), \emph{Horizontal Bar} (9), \emph{Horned Puffin} (5), \emph{Horse} (49), \emph{Horse Mackerel} (8), \emph{Horsebean} (8), \emph{Hospital Sign} (20), \emph{Hot Air Balloon} (103), \emph{Hot Chocolate} (28), \emph{Housefly} (20), \emph{Hovercraft} (43), \emph{Howa Stalker} (3), \emph{Huber Needle} (2), \emph{Hudson H9} (10), \emph{Hula Hoop} (1), \emph{Hula Skirt Siphonophore} (8), \emph{Humpback Whale} (30), \emph{Humphead Parrotfish} (35), \emph{Hundred-Pace Viper} (4), \emph{Hurdy-Gurdy} (34), \emph{Hurling} (14), \emph{Hwando} (4), \emph{Hwandudaedo} (2), \emph{Hydration Backpack} (28), \emph{Hydraulic Breaker} (2), \emph{Hydrometer} (6), \emph{Hyena} (27), \emph{Hypodermic Needle} (3).}\\

\noindent\scriptsize{\centering{\textbf{\normalsize{I (62 classes)}}}\\\emph{Ia-2} (3), \emph{Ibex} (39), \emph{Ice Bag} (32), \emph{Ice Cream Box} (5), \emph{Ice Hockey Player} (3), \emph{Ice Rescue Suit} (24), \emph{Ice Speed Skating} (30), \emph{Ice Skater} (35), \emph{Icp} (11), \emph{Imi Desert Eagle} (10), \emph{Impact Driver} (19), \emph{Impatiens} (29), \emph{Imperial Shag} (23), \emph{Inca Tern} (10), \emph{Incense Burner} (27), \emph{Inclinometer} (6), \emph{Indian Ringneck Parrot} (31), \emph{Indochinese Green Magpie} (35), \emph{Indonesian Rupiah} (54), \emph{Inflatable Bounce Houses} (7), \emph{Inflatable Castle} (3), \emph{Inflatable Obstacle Course} (5), \emph{Inflatable Rescue Boat} (24), \emph{Infrared Thermometer} (63), \emph{Infusion Bags} (20), \emph{Infusion Pump} (3), \emph{Ink} (84), \emph{Inline Hockey} (30), \emph{Insect Repellent Bottle} (20), \emph{Instrument Tray} (16), \emph{Insulated Backpack} (29), \emph{Insulated Cup} (76), \emph{Insulin Pen} (3), \emph{Insulin Pump} (3), \emph{Interactive Fountains} (14), \emph{Interactive Light Walls} (9), \emph{Interactive Play Panel} (1), \emph{Interactive Sound Play} (3), \emph{Interceptor} (275), \emph{Intermediate Egret} (42), \emph{International Chess} (9), \emph{Intradermal Needle} (3), \emph{Intramuscular Needle} (3), \emph{Intraosseous Needle} (3), \emph{Iridescent Shark} (30), \emph{Iron Gate Wheel} (13), \emph{Iv Access Kit} (1), \emph{Iv Armboard} (3), \emph{Iv Ball} (3), \emph{Iv Clamp} (5), \emph{Iv Dressing} (3), \emph{Iv Drip Chamber} (3), \emph{Iv Flow Regulator} (9), \emph{Iv Injector} (1), \emph{Iv Pole} (2), \emph{Iv Pressure Bag} (3), \emph{Iv Push Syringe} (3), \emph{Iv Site Protector} (3), \emph{Iv Spike} (2), \emph{Iv Tubing} (3), \emph{Iv Warmer} (4), \emph{Ivory Gull} (5).}\\

\noindent\scriptsize{\centering{\textbf{\normalsize{J (27 classes)}}}\\\emph{Jack Dempsey Cichlid} (35), \emph{Jackal} (27), \emph{Jacket} (3), \emph{Jackfruit} (50), \emph{Jai Alai} (3), \emph{Jam Jar} (28), \emph{Jamaican Dollar} (35), \emph{Japanese Bobtail Cat} (3), \emph{Japanese Chin} (3), \emph{Japanese Keelback} (3), \emph{Japanese Macaque} (15), \emph{Japanese Murrelet} (3), \emph{Japanese Spitz} (3), \emph{Japanese White Eye} (35), \emph{Japanese Yen} (103), \emph{Javelin} (25), \emph{Jeans} (4), \emph{Jerboa} (2), \emph{Jerdon S Pitviper} (3), \emph{Jewelry Box} (23), \emph{Jigsaw Puzzle} (40), \emph{John Dory} (35), \emph{Judo Player} (35), \emph{Juggling Clubs} (15), \emph{Juice} (32), \emph{Jump Rope} (156), \emph{Jungle Gym} (3).}\\

\noindent\scriptsize{\centering{\textbf{\normalsize{K (30 classes)}}}\\\emph{K2} (3), \emph{Kangaroo} (80), \emph{Karate Athlete} (27), \emph{Kayak} (43), \emph{Keep Off Median Sign} (2), \emph{Kelp Gull} (10), \emph{Keyboard} (24), \emph{Kh2002} (1), \emph{Kids Playing In Parks} (25), \emph{Killer Whale} (35), \emph{Kimber Hunter Pro Desolve Blak} (3), \emph{Kin-Ball} (30), \emph{Kinetic Ball} (1), \emph{King Cobra} (2), \emph{Kinkajou} (28), \emph{Kissing Gourami} (33), \emph{Kitchen Scissors} (10), \emph{Kknd} (33), \emph{Knee Pads} (4), \emph{Knight} (25), \emph{Knitwear} (4), \emph{Knives} (10), \emph{Knobkierrie} (6), \emph{Koala} (29), \emph{Korfball} (30), \emph{Kpinga} (10), \emph{Kuhli Loach} (32), \emph{Kukri} (34), \emph{Kung Fu Fan Dance} (15), \emph{Kusarigama} (21).}\\

\noindent\scriptsize{\centering{\textbf{\normalsize{L (85 classes)}}}\\\emph{L85A2} (2), \emph{Labrador Retriever} (19), \emph{Labrisomus Nuchipinnis} (3), \emph{Labrisomus Philippii} (2), \emph{Lacewing} (3), \emph{Lacrosse} (10), \emph{Ladyfinger Cactus} (4), \emph{Laminate Trimmer} (1), \emph{Lantern} (29), \emph{Laptop} (50), \emph{Large-Billed Crow} (34), \emph{Large-Billed Tern} (10), \emph{Large-Tailed Nightjar} (34), \emph{Laser Distance Measurer} (57), \emph{Laser Level} (48), \emph{Latin Dancer} (70), \emph{Laundry Bag} (37), \emph{Lava Gull} (2), \emph{Lavender} (2), \emph{Lawn Bowling Set} (40), \emph{Lawnmower Blade} (78), \emph{Lawnmower Wheel} (32), \emph{Lce Hockey} (23), \emph{League Of Legends} (77), \emph{Least Auklet} (5), \emph{Lebel Mle 1886} (3), \emph{Leggings} (4), \emph{Lego Building Set} (138), \emph{Legwarmers} (4), \emph{Lemming} (10), \emph{Lemon} (32), \emph{Lemon Cactus} (3), \emph{Lemon Shark} (29), \emph{Lemur} (54), \emph{Lentil} (1), \emph{Leopard} (37), \emph{Leopard Bush Fish} (41), \emph{Lepomis Auratus} (31), \emph{Lepomis Gulosus} (9), \emph{Lepomis Macrochirus} (32), \emph{Lepomis Marginatus} (17), \emph{Lesser Frigatebird} (10), \emph{Lesser Spotted Woodpecker} (45), \emph{Lesser Panda} (33), \emph{Lhasa Apso} (3), \emph{Life Jacket} (46), \emph{Lifebuoy} (16), \emph{Lifting Hook Pulley} (20), \emph{Light Rail Train} (34), \emph{Lighter} (37), \emph{Lilac} (32), \emph{Linemans Pliers} (20), \emph{Liobagrus Reinii} (1), \emph{Lion Dance} (98), \emph{Lionfish} (29), \emph{Lipstick} (18), \emph{Litchi} (43), \emph{Little Auk} (2), \emph{Little Egret} (52), \emph{Little Forktail} (35), \emph{Little Gull} (4), \emph{Little Tern} (9), \emph{Little Wart Snake} (15), \emph{Lntravenous Catheter} (2), \emph{Logistics Delivery Vehicle} (35), \emph{Long Sword} (31), \emph{Longan} (41), \emph{Longear Sunfish} (28), \emph{Longicorn Beetle} (5), \emph{Longnose Hawkfish} (35), \emph{Long-Tailed Goral} (2), \emph{Long-Tailed Skua} (10), \emph{Loofah} (4), \emph{Loquat} (43), \emph{Lotus Boat} (12), \emph{Lucifer Titi Monkey} (5), \emph{Luer Adapter} (2), \emph{Luer Lock Cap} (7), \emph{Luer Slip Syringe} (2), \emph{Luger P08} (44), \emph{Luggage Cart Wheel} (39), \emph{Lunch Box} (24), \emph{Lutjanus Decussatus} (2), \emph{Lux Meter} (106), \emph{Lynx} (87).}\\

\noindent\scriptsize{\centering{\textbf{\normalsize{M (133 classes)}}}\\\emph{M1 Grand} (3), \emph{M107} (3), \emph{M110} (3), \emph{M16} (3), \emph{M1903} (3), \emph{M1911A1} (3), \emph{M1917} (2), \emph{M2 Heavy Machine Gun} (3), \emph{M25} (2), \emph{M3} (2), \emph{M40A1} (2), \emph{M45A1} (10), \emph{M4A1} (3), \emph{M700} (3), \emph{Macadamia Nut} (2), \emph{Mace Spinning} (14), \emph{Mackerel} (5), \emph{Magazine Lee-Metford} (2), \emph{Magic Mirror Maze} (2), \emph{Magic Track} (60), \emph{Magicians Performing Quick Hand Tricks} (3), \emph{Maglev Train} (34), \emph{Magnetic Sweeper} (3), \emph{Magnificent Frigatebird} (9), \emph{Magnifier} (1), \emph{Mail Truck} (10), \emph{Maine Coon Cat} (4), \emph{Mainland Serow} (8), \emph{Malayan Tapir} (49), \emph{Malaysian Ringgit} (24), \emph{Mallard} (21), \emph{Maltese} (13), \emph{Mammillaria} (35), \emph{Manatee} (1), \emph{Mandarin Duck} (35), \emph{Mandarin Rat Snake} (3), \emph{Mandarinfish} (33), \emph{Mandocello} (35), \emph{Mandolin} (3), \emph{Mango} (37), \emph{Mangosteen} (16), \emph{Mangual} (12), \emph{Mantis Shrimp} (8), \emph{Mantled Howler} (2), \emph{Many-Banded Krait} (3), \emph{Marathon Runner} (51), \emph{Marble Run Set} (28), \emph{Marbled Cat} (3), \emph{Marbled Hatchetfish} (35), \emph{Marbled Murrelet} (12), \emph{Marimba} (31), \emph{Mascara Wand} (18), \emph{Masked Booby} (10), \emph{Mason S Square} (2), \emph{Masonry Bit} (2), \emph{Matsutake} (84), \emph{Mattock} (33), \emph{Mauve Stinger} (34), \emph{Mbe} (11), \emph{Mccoskers Flasher Wrasse} (11), \emph{Measuring Cups And Spoons} (9), \emph{Measuring Spoon} (9), \emph{Meat Grinder Blade} (22), \emph{Meat Slicer} (6), \emph{Medical Cart Wheel} (57), \emph{Medicine Bottle} (42), \emph{Medieval Quarterstaff Combat} (27), \emph{Melocactus} (34), \emph{Merge Right Sign} (15), \emph{Metal Cutting Saw Blade} (78), \emph{Metal Lathe Chuck} (39), \emph{Meteor Hammer} (25), \emph{Microcontroller Tester} (19), \emph{Micrometer} (24), \emph{Micropterus Coosae} (22), \emph{Micropterus Salmoides} (32), \emph{Microscope} (34), \emph{Midi Skirt} (3), \emph{Military Transport Aircraft} (53), \emph{Millers Saki} (4), \emph{Mimic Octopus} (35), \emph{Mini Bowling Alley} (8), \emph{Mini Quadcopter} (32), \emph{Mini Roller Coaster} (3), \emph{Mini Soccer Field} (2), \emph{Miniature Schnauzer} (19), \emph{Miniature Village} (4), \emph{Minimalist Wallet} (31), \emph{Mink} (28), \emph{Mirror} (33), \emph{Mirrored Sunglasses} (5), \emph{Miter Saw} (20), \emph{Mixing Bowls} (7), \emph{Mixing Syringe} (2), \emph{Mk 14 Ebr} (3), \emph{Mk12} (2), \emph{Mk14} (3), \emph{Mk20} (3), \emph{Modern Martial Arts With Bo Staff} (20), \emph{Modular Backpack} (19), \emph{Moisture Meter} (29), \emph{Mole Rat} (13), \emph{Momentum Elite} (3), \emph{Mongolian Jird} (16), \emph{Monkey} (34), \emph{Monkey Terrier} (3), \emph{Monowheel} (31), \emph{Moon Jellyfish} (35), \emph{Moonlight Gourami} (30), \emph{Moose} (33), \emph{Morel} (34), \emph{Morus Capensis} (3), \emph{Mosasaurus} (38), \emph{Mosin-Nagant M1891} (1), \emph{Moth Bean} (5), \emph{Motorcycle} (14), \emph{Mountain Goat} (33), \emph{Mountain Hare} (33), \emph{Mp40} (3), \emph{Mp5} (3), \emph{Mp7} (3), \emph{Mpi Km} (3), \emph{Mpi Kms74} (3), \emph{Mrad} (3), \emph{Muffin Tin} (5), \emph{Multirole Fighter} (29), \emph{Mung Bean} (6), \emph{Music Box} (30), \emph{Music Express} (12), \emph{Musk Deer} (10), \emph{Musk Ox} (33), \emph{Muskrat} (28), \emph{Mustard Bottle} (50).}\\

\noindent\scriptsize{\centering{\textbf{\normalsize{N (32 classes)}}}\\\emph{Nail Clippers} (73), \emph{Nandao} (30), \emph{Napo Saki} (3), \emph{Narrow Bridge Sign} (4), \emph{Nassau Grouper} (30), \emph{Nature Observation Deck} (2), \emph{Navanax} (4), \emph{Neck Warmer} (3), \emph{Needle Counter} (3), \emph{Needle Disposal Container} (3), \emph{Needle Holder} (3), \emph{Needle Nose Pliers} (20), \emph{Needleless Connector} (3), \emph{Nepalese Rupee} (5), \emph{New Zealand Dollar} (49), \emph{Newt} (50), \emph{Ney} (44), \emph{Night Blooming Cereus} (32), \emph{Niviventer Fulvescens} (3), \emph{No Bicycles Sign} (1), \emph{No Parking Fire Lane Sign} (25), \emph{No Parking Sign} (24), \emph{No Passing Zone Sign} (2), \emph{No Turn On Red Sign} (32), \emph{Northern Gannet} (9), \emph{Northern Lapwing} (39), \emph{Norwegian Krone} (25), \emph{Notebook} (31), \emph{Notebook Packing Box} (36), \emph{Numbat} (42), \emph{Nunchucks Play} (26), \emph{Nyckelharpa} (33).}\\

\noindent\scriptsize{\centering{\textbf{\normalsize{O (29 classes)}}}\\\emph{Oats} (2), \emph{Oboe} (42), \emph{Observation Aircraft} (35), \emph{Ocean Sunfish} (55), \emph{Ocelot} (1), \emph{Octopus Oculifer} (5), \emph{Odonus Niger} (50), \emph{Office Chair Base Wheel} (44), \emph{Off-Road Vehicle} (55), \emph{Oil Tanker} (37), \emph{Olive Sea Turtle} (34), \emph{Olives} (11), \emph{One-Shoulder Top} (3), \emph{Onion} (4), \emph{Opalescent Nudibranch} (35), \emph{Opossum} (27), \emph{Orange} (47), \emph{Oriental Dollarbird} (25), \emph{Oriental Small-Clawed Otter} (9), \emph{Oriental Turtle-Dove} (54), \emph{Ornate Octopus} (33), \emph{Oscillating Multi-Tool} (19), \emph{Osmanthus Fragrans} (22), \emph{Ostrich} (35), \emph{Otter} (51), \emph{Oven} (25), \emph{Oversized Tote} (39), \emph{Oxyeleotris Marmorata} (12), \emph{Oyster Mushroom} (30).}\\

\noindent\scriptsize{\centering{\textbf{\normalsize{P (155 classes)}}}\\\emph{Pacific Gull} (10), \emph{Pacific Spiny Lumpsucker} (35), \emph{Packing Cube} (38), \emph{Pad} (46), \emph{Paddle Go-Round} (10), \emph{Paddle Wheeler Ride} (5), \emph{Paddleball} (30), \emph{Paddlefish} (33), \emph{Paddy} (2), \emph{Paint Sprayer} (3), \emph{Painted Frogfish} (35), \emph{Painter} (93), \emph{Pajama Cardinalfish} (25), \emph{Pajamas} (3), \emph{Pallass Fish Eagle} (35), \emph{Pallass Rosefinch} (40), \emph{Panamanian Balboa} (74), \emph{Panel Saw} (20), \emph{Pannier Bag} (50), \emph{Pants} (3), \emph{Paper Airplane} (6), \emph{Paper Clip} (6), \emph{Paper Cutter} (15), \emph{Paper Nautilus} (37), \emph{Papillon} (3), \emph{Parablennius Gattorugine} (26), \emph{Parachanna Africana} (40), \emph{Paracheilinus Bellae} (10), \emph{Paracheilinus Cyaneus} (3), \emph{Paracheilinus Hemitaeniatus} (4), \emph{Paracheilinus Octotaenia} (19), \emph{Paracheilinus Piscilineatus} (16), \emph{Paracheilinus Rubricaudalis} (6), \emph{Paraclinus Spectator} (1), \emph{Paradise Fish} (49), \emph{Pareques Acuminatus} (19), \emph{Parkour Practitioners} (1), \emph{Parodia Cactus} (11), \emph{Passenger Aircraft} (65), \emph{Pasta Server} (8), \emph{Pata} (3), \emph{Pca Pump} (10), \emph{Pea} (6), \emph{Peacock} (8), \emph{Peacock Grouper} (14), \emph{Peacock Gudgeon} (34), \emph{Peanutt} (9), \emph{Pear} (41), \emph{Pecan} (3), \emph{Pedal Karts} (3), \emph{Peeler} (6), \emph{Pen Needle} (3), \emph{Pencil} (50), \emph{Pencil Case} (27), \emph{Pepper} (4), \emph{Pepper Mill} (9), \emph{Percnon Gibbesi} (11), \emph{Persian Cat} (2), \emph{Peruvian Apple Cactus} (10), \emph{Peruvian Booby} (8), \emph{Peruvian Night Monkey} (4), \emph{Peruvian Sol} (5), \emph{Peyote} (6), \emph{Ph Meter} (99), \emph{Philippine Tarsier} (3), \emph{Phone} (49), \emph{Phone Case} (117), \emph{Phone Charger} (28), \emph{Photochromic Lenses} (1), \emph{Pianist} (35), \emph{Picasso Triggerfish} (34), \emph{Picc Line} (3), \emph{Piccolo} (32), \emph{Pickup Truck} (27), \emph{Picnic Basket} (26), \emph{Pied Tamarin} (3), \emph{Piglet Squid} (20), \emph{Pig-Tailed Macaque} (34), \emph{Pikachu Nudibranch} (35), \emph{Pillbug} (5), \emph{Pilot Boat} (57), \emph{Pinafore} (3), \emph{Pine Marten} (10), \emph{Pine Nut} (34), \emph{Pink-Browed Rosefinch} (20), \emph{Pipa} (35), \emph{Pipe Stand} (5), \emph{Pipe Vise} (18), \emph{Pipe Wrench} (15), \emph{Pistachio} (1), \emph{Pizza Stone} (8), \emph{Plaid Shirt} (3), \emph{Plains Zebra} (9), \emph{Plant Box} (26), \emph{Plaster} (21), \emph{Plate} (29), \emph{Platform Shoes} (97), \emph{Platy Fish} (35), \emph{Platypus} (26), \emph{Playground Swing Set} (30), \emph{Playing Cards} (5), \emph{Plug Strip} (59), \emph{Plum Ak74} (3), \emph{Plumb Bob} (8), \emph{Plum-Headed Parakeet} (33), \emph{Pneumatic Drill} (1), \emph{Pod} (4), \emph{Pokemon} (45), \emph{Polarized Sunglasses} (2), \emph{Police Baton Training} (16), \emph{Polleni Grouper} (8), \emph{Polo} (15), \emph{Polynemus Paradiseus} (15), \emph{Pomarine Skua} (10), \emph{Pomegranate} (89), \emph{Pomeranian} (1), \emph{Poodle} (20), \emph{Popcorn Box} (35), \emph{Poppies} (6), \emph{Porpoise} (9), \emph{Port Jackson Shark} (37), \emph{Portable Buoyant Device} (88), \emph{Portable Spotlight} (20), \emph{Portly Spider Crab} (11), \emph{Potato} (3), \emph{Potato Masher} (8), \emph{Potato Ricer} (7), \emph{Pottery Figurine} (218), \emph{Power Broom} (18), \emph{Power Float} (3), \emph{Power Screed} (20), \emph{Ppk} (37), \emph{Ppq} (30), \emph{Prefilled Syringe} (3), \emph{Pressure Cooker} (5), \emph{Pressure Gauge} (35), \emph{Prince Of Persia} (4), \emph{Printed Dress} (3), \emph{Printer Paper} (30), \emph{Printing Press Plate Cylinder} (60), \emph{Proboscis Monkey} (30), \emph{Protective Clothing} (4), \emph{Protist} (31), \emph{Prototype Racing Car} (48), \emph{Protractor} (95), \emph{Pudao} (10), \emph{Pufferfish} (32), \emph{Pull-Up Bars} (9), \emph{Pumpkin} (4), \emph{Pumpkin Seedt} (9), \emph{Putty Knife} (18), \emph{Pygmy Cormorant} (10), \emph{Pygmy Gourami} (36), \emph{Pygmy Hog} (28), \emph{Pygmy Squid} (16).}\\

\noindent\scriptsize{\centering{\textbf{\normalsize{Q (4 classes)}}}\\\emph{Qanun} (35), \emph{Qipao Cheongsam} (4), \emph{Quail Egg} (26), \emph{Quinoa} (6).}\\

\noindent\scriptsize{\centering{\textbf{\normalsize{R (107 classes)}}}\\\emph{Raccoon} (34), \emph{Racing Car} (34), \emph{Radish} (2), \emph{Ragdoll Cat} (3), \emph{Railroad Crossing Ahead Sign} (59), \emph{Rain Boots} (9), \emph{Rainbow Shark} (34), \emph{Rainbow Shirt} (2), \emph{Rapier} (169), \emph{Razor} (29), \emph{Razorbill} (3), \emph{Rc Boat} (35), \emph{Reading Glasses} (11), \emph{Rebab} (33), \emph{Reciprocating Saw} (20), \emph{Reconstitution Syringe} (3), \emph{Record Player Turntable} (42), \emph{Red Arrow Raw 15} (3), \emph{Red Bean} (6), \emph{Red Damselfly} (19), \emph{Red Deer} (25), \emph{Red Firefish} (35), \emph{Red Large-Toothed Snake} (3), \emph{Red River Hog} (53), \emph{Red Stumpnose} (13), \emph{Red Velvetfish} (12), \emph{Red Fox} (30), \emph{Red-Billed Blue Magpie} (91), \emph{Red-Billed Leiothrix} (15), \emph{Red-Browed Finch} (65), \emph{Red-Crowned Crane} (40), \emph{Red-Faced Spider Monkey} (4), \emph{Redflank Bloodfin} (3), \emph{Red-Flanked Bluetail} (35), \emph{Red-Lipped Batfish} (35), \emph{Red-Tailed Black Shark} (10), \emph{Red-Tailed Pipe Snake} (3), \emph{Red-Wattled Lapwing} (32), \emph{Red-Winged Blackbird} (34), \emph{Reeves S Muntjac} (31), \emph{Reindeer} (61), \emph{Remington Msr} (2), \emph{Remote Control} (90), \emph{Remote Control Airplane} (196), \emph{Reporter} (13), \emph{Rescue Basket Stretcher} (11), \emph{Rescue Board} (52), \emph{Rescue Breathing Equipment} (1), \emph{Rescue Buoy} (3), \emph{Rescue Buoy With Lights} (5), \emph{Rescue Sled} (6), \emph{Rescue Snap Hook} (80), \emph{Rescue Strobe} (16), \emph{Rescue Throw Line} (20), \emph{Rescue Tube} (12), \emph{Rescue Whistle} (11), \emph{Rhinoceros} (35), \emph{Rhinoceros Auklet} (5), \emph{Rhythmic Gymnastics With Clubs} (30), \emph{Ribbon Seal} (48), \emph{Richards Pipit} (49), \emph{Rickshaw} (45), \emph{Ring} (107), \emph{Ring-Tailed Lemur} (31), \emph{River Tern} (10), \emph{Riverboat} (20), \emph{Robe} (3), \emph{Robot} (22), \emph{Robot Dog} (51), \emph{Rock Shag} (4), \emph{Rocket} (140), \emph{Rocket Plane} (15), \emph{Rocking Horse} (60), \emph{Roe Deer} (24), \emph{Roller Coaster} (6), \emph{Roller Hockey} (4), \emph{Roller Hockey Puck} (50), \emph{Roller Skate} (8), \emph{Roller Skating} (11), \emph{Roller Speed Skating} (30), \emph{Rolling Barrels} (3), \emph{Rolling Pin} (10), \emph{Rolling Tool Tote} (3), \emph{Rope Bridge} (1), \emph{Roseline Shark} (22), \emph{Rotary Cutter Blade} (70), \emph{Rotary Hammer} (18), \emph{Rotating Climber} (5), \emph{Rough Road Sign} (5), \emph{Round Glasses} (3), \emph{Roundabout Sign} (12), \emph{Rower} (25), \emph{Rowing} (48), \emph{Royal Tern} (9), \emph{Rpk} (3), \emph{Rpk74} (3), \emph{Rpk74M} (3), \emph{Ruan} (32), \emph{Ruby-Throated Hummingbird} (35), \emph{Ruffe Fish} (6), \emph{Ruger Gp100} (37), \emph{Ruger P90} (32), \emph{Ruger Precision Rifle} (3), \emph{Ruger Sr9C} (60), \emph{Russian Blue Cat} (5), \emph{Rv} (31), \emph{Ryegrass} (5).}\\

\noindent\scriptsize{\centering{\textbf{\normalsize{S (233 classes)}}}\\\emph{Safety Glasses} (2), \emph{Safety Needle} (2), \emph{Saiga Antelope} (8), \emph{Sailboat} (31), \emph{Sailfish} (8), \emph{Sailing Dinghy} (40), \emph{Salad Dressing Bottle} (6), \emph{Salvia Splendens} (33), \emph{Samoyed} (13), \emph{Sand And Water Table} (6), \emph{Sand Diggers} (10), \emph{Sandals} (38), \emph{Sandbox Toys} (3), \emph{Sanxian} (19), \emph{Sar21} (1), \emph{Sarangi} (26), \emph{Sarrusophone} (1), \emph{Saucepan} (10), \emph{Savage Impulse} (3), \emph{Sawshark} (17), \emph{Saxophone} (34), \emph{Scalpel} (17), \emph{Scar} (3), \emph{Scarf} (94), \emph{Scarlet Finch} (41), \emph{Sccy Cpx-1} (10), \emph{Schmidt-Rubin M1889} (2), \emph{School Bus} (32), \emph{School Zone Sign} (30), \emph{School Zone Speed Limit Sign} (39), \emph{Scissor Lift} (13), \emph{Scolopsis Bilineata} (12), \emph{Scooter} (8), \emph{Scottish Fold Cat} (4), \emph{Screw Gun} (9), \emph{Scrippss Murrelet} (1), \emph{Scrubber} (16), \emph{Sea Angel} (22), \emph{Seahorse} (34), \emph{Security Guard} (13), \emph{Semi-Rimless Glasses} (2), \emph{Sepak Raga} (10), \emph{Sepak Takraw} (30), \emph{Sergeant Major Fish} (44), \emph{Severum Cichlid} (35), \emph{Sewer Cleaning Truck} (27), \emph{Sewing Thread} (50), \emph{Shadow Play Area} (3), \emph{Shakuhachi} (24), \emph{Shaolin Stick Fighting} (30), \emph{Sharp Turn Sign} (1), \emph{Shawl} (7), \emph{Sheep} (37), \emph{Shell Game Cup} (12), \emph{Shiba Inu} (12), \emph{Shield Bug} (5), \emph{Shillelagh} (34), \emph{Ship Is Wheel} (39), \emph{Shirt} (4), \emph{Shoe Box} (32), \emph{Shoes} (97), \emph{Shogi} (5), \emph{Shopping Cart Wheel} (38), \emph{Shorts} (4), \emph{Shot Put} (4), \emph{Shoulder Drop-Off Sign} (1), \emph{Shrew} (12), \emph{Shuttlecock} (30), \emph{Siamese Cat} (4), \emph{Siberian Husky} (15), \emph{Siberian Rubythroat} (49), \emph{Siberian Thrush} (4), \emph{Siberian Weasel} (34), \emph{Sig P220S} (3), \emph{Sig P226} (11), \emph{Sig Sauer M400 Tread} (3), \emph{Sig-Sauer P230} (7), \emph{Sika Deer} (30), \emph{Silver Dollar Fish} (12), \emph{Silver Torch Cactus} (4), \emph{Silvery Marmoset} (3), \emph{Singer} (64), \emph{Sitar} (35), \emph{Site Light} (3), \emph{Skate} (22), \emph{Skateboard} (6), \emph{Skateboard Ramp} (18), \emph{Skateboard Wheel} (48), \emph{Skateboarder} (69), \emph{Ski Pole} (98), \emph{Skillet} (9), \emph{Skirt} (4), \emph{Skydiving Aircraft} (26), \emph{Sleeping Bag} (60), \emph{Sling Bag} (1), \emph{Sling Seat Swing} (8), \emph{Slipform Paver} (8), \emph{Slipper Lobster} (25), \emph{Slippers} (4), \emph{Slippery Road Sign} (2), \emph{Slot Car} (56), \emph{Slot Machine Wheel} (20), \emph{Sloth Bear} (34), \emph{Slow Children Playing Sign} (14), \emph{Slow Loris} (10), \emph{Slow Signal Sign} (7), \emph{Slr Camera Len} (45), \emph{Snailfish} (19), \emph{Sneaker} (30), \emph{Snooker} (35), \emph{Snow Leopard} (34), \emph{Snowboard} (29), \emph{Snowmobile} (34), \emph{Snowmobile Track Wheel} (58), \emph{Snowplow} (35), \emph{Snowshoe Hare} (35), \emph{Snowy Owl} (31), \emph{Soap Box} (29), \emph{Soccer Player} (11), \emph{Socks} (4), \emph{Solar-Powered Boat} (31), \emph{Soldier} (59), \emph{Sooty Gull} (4), \emph{Sound Velocity Meter} (11), \emph{Sousaphone} (29), \emph{South African Rand} (85), \emph{South Korean Won} (36), \emph{South Polar Skua} (1), \emph{Soy Milk} (22), \emph{Soy Sauce Bottle} (9), \emph{Soybean} (4), \emph{Spanish Dancer} (35), \emph{Spatula} (9), \emph{Spear Thrower} (8), \emph{Spectacled Caiman} (34), \emph{Spectacled Guillemot} (5), \emph{Spectrophotometer} (28), \emph{Speed Bump Sign} (5), \emph{Spell} (55), \emph{Sphygmomanometer} (25), \emph{Sphynx Cat} (4), \emph{Spider} (5), \emph{Spin Zone} (9), \emph{Spinal Needle} (3), \emph{Spinning Platforms} (9), \emph{Spinning Top} (3), \emph{Spiral Climber} (2), \emph{Spiral Spinners} (1), \emph{Spiral Vegetable Slicer} (7), \emph{Spirit Level} (44), \emph{Splash Pads} (2), \emph{Sponge Crab} (43), \emph{Spoon} (34), \emph{Sports Car} (30), \emph{Spotted Hatchetfish} (3), \emph{Spotted Knifejaw} (25), \emph{Spotted-Tail Quoll} (32), \emph{Spray Paint Wheel} (70), \emph{Spring Rider} (15), \emph{Springfield Armory Waypoint} (3), \emph{Springfield Xdm-10} (20), \emph{Square Dancer} (29), \emph{Squash} (48), \emph{Squash Racket} (30), \emph{Squirrel} (6), \emph{Squirrelfish} (27), \emph{Sr-25} (3), \emph{Ss2} (1), \emph{Staple Gun} (10), \emph{Stapler} (28), \emph{Starcraft} (33), \emph{State Highway Sign} (5), \emph{Statues Of Buddha And Deities} (3), \emph{Steamer} (23), \emph{Steamers} (6), \emph{Steel Drums} (31), \emph{Steel Toe Boots} (34), \emph{Steel Trowel} (2), \emph{Steep Hill Sign} (5), \emph{Steering Wheel} (48), \emph{Stejneger S Pit Viper} (3), \emph{Stethoscope} (12), \emph{Stick Fighting Arts} (30), \emph{Sticky Note} (1), \emph{Stingrays} (24), \emph{Stock Pot} (10), \emph{Stopcock} (5), \emph{Strainer} (9), \emph{Strawberry} (39), \emph{Street Dancer} (69), \emph{Street Pedestrian} (66), \emph{Street Sweeper} (46), \emph{Striped Boarfish} (35), \emph{Striped Burrfish} (35), \emph{Striped Dolphin} (23), \emph{Striped Hyena} (31), \emph{Striped Pyjama Squid} (30), \emph{Striped Shirt} (3), \emph{Striped Snakehead} (9), \emph{Stroller Wheel} (55), \emph{Suit} (3), \emph{Sun Bear} (74), \emph{Sun Hat} (51), \emph{Sunflower} (48), \emph{Sunflower Seedt} (5), \emph{Sunglasses} (6), \emph{Super Mario} (35), \emph{Superagui Lion Tamarin} (3), \emph{Surfboard} (27), \emph{Surfer} (32), \emph{Swab} (15), \emph{Swallows} (46), \emph{Swallow-Tailed Gull} (4), \emph{Sweater} (3), \emph{Swedish Krona} (4), \emph{Sweeper Truck} (90), \emph{Sweeping Robot} (55), \emph{Sweet Potato} (4), \emph{Swimwear} (4), \emph{Swing Gliders} (1), \emph{Swing The Ball} (24), \emph{Swiss Franc} (20), \emph{Sword And Fairy} (34), \emph{Swordfish} (8), \emph{Swordtail Fish} (34), \emph{Synodontis Grandiops} (21), \emph{Synodontis Multipunctatus} (47), \emph{Synthesizer} (30), \emph{Syrian Hamster} (33), \emph{Syringe} (38), \emph{Syringe Cap} (2), \emph{Syringe Filter} (2), \emph{Syringe Pump} (2).}\\

\noindent\scriptsize{\centering{\textbf{\normalsize{T (114 classes)}}}\\\emph{T Intersection Sign} (6), \emph{Table Tennis} (1), \emph{Table Tennis Bats} (19), \emph{Table Tennis Player} (39), \emph{Tac-50} (3), \emph{Taiko Drum} (29), \emph{Takin} (35), \emph{Tamarind Fruit} (4), \emph{Tango Dancer} (30), \emph{Tank} (55), \emph{Tanto} (13), \emph{Tape Measure} (34), \emph{Tar} (7), \emph{Tar-21} (2), \emph{Tasmanian Devil} (31), \emph{T-Bevel} (3), \emph{Tea} (21), \emph{Tea Tree Mushroom} (2), \emph{Teeter-Totter} (20), \emph{Tekko Kagi} (11), \emph{Telescope} (11), \emph{Telescopic Handler} (3), \emph{Television} (28), \emph{Tem} (5), \emph{Temmincks Courser} (9), \emph{Temporary Road Closed Sign} (2), \emph{Tennis} (30), \emph{Tennis Racket} (30), \emph{Tensile Testing Machine} (4), \emph{Terapon Jarbua} (42), \emph{Testudinidae} (29), \emph{Texas Prickly Pear Cactus} (5), \emph{Thalasseus Sandvicensis} (8), \emph{The Witcher} (26), \emph{Theorbo} (16), \emph{Thermometer} (14), \emph{Thickness Gauge} (100), \emph{Thimble Cactus} (24), \emph{Thorny Devil} (30), \emph{Thread Gauge} (33), \emph{Three-Section Staff} (35), \emph{Threestripe Gourami} (20), \emph{Thresher Shark} (34), \emph{Throw Bag} (73), \emph{Throwing Axe} (34), \emph{Throwing Knife Demonstrations} (29), \emph{Tiara} (105), \emph{Tibetan Mastiff} (19), \emph{Tibetan Antelope} (35), \emph{Tiger} (27), \emph{Tiger Barb} (11), \emph{Tiger Hook Swords} (33), \emph{Tiger Shark} (35), \emph{Tile Spacers} (2), \emph{Tilting Platforms} (1), \emph{Timer} (1), \emph{Timpani} (28), \emph{Tin Whistle} (29), \emph{Tire Swing} (36), \emph{Tissue Box} (34), \emph{Toddler Play Area} (2), \emph{Toilet Brush} (10), \emph{Toll Booth Sign} (1), \emph{Tom Thumb Cactus} (3), \emph{Tomahawk} (31), \emph{Tomato} (2), \emph{Tomato Slicer} (4), \emph{Tongs} (8), \emph{Toothpaste Bottle} (4), \emph{Torch Cactus} (33), \emph{Torque Wrench} (44), \emph{Tortoise Beetle} (5), \emph{Tote Bag} (33), \emph{Touchscreen Gloves} (4), \emph{Tourist Guide} (9), \emph{Toy Box} (5), \emph{Toy Car Wheel} (31), \emph{Toy Piano} (29), \emph{Track And Field Athlete} (30), \emph{Trackless Train} (3), \emph{Tracksuit} (4), \emph{Traditional Chinese Sword Dance} (24), \emph{Traditions Pursuit Xt} (3), \emph{Train} (31), \emph{Tray} (3), \emph{Treefish} (10), \emph{Trench Club} (27), \emph{Trench Coat} (3), \emph{Trencher} (3), \emph{Triangle} (35), \emph{Triangle Board} (5), \emph{Trichocereus} (10), \emph{Tripod Stand} (20), \emph{Trocar Needle} (2), \emph{Trolley} (30), \emph{Trolley Bag} (41), \emph{Trombone} (48), \emph{Truck Racing Truck} (48), \emph{Trumpet} (51), \emph{T-Shirt} (3), \emph{T-Shirt Dress} (2), \emph{Tuba} (34), \emph{Tufted Capuchin} (3), \emph{Tufted Deer} (35), \emph{Tufted Puffin} (4), \emph{Tugboat Ride} (3), \emph{Tumble Drums} (7), \emph{Tumble Tracks} (6), \emph{Tuna Crab} (24), \emph{Tunnel Maze} (2), \emph{Turkish Lira} (2), \emph{Turks Cap Cactus} (4), \emph{Turtleneck} (3), \emph{Tyrannosaurus Rex} (28).}\\

\noindent\scriptsize{\centering{\textbf{\normalsize{U (11 classes)}}}\\\emph{Uav} (48), \emph{Ukulele} (49), \emph{Umbrella} (31), \emph{Underwear} (8), \emph{Uniform} (4), \emph{United Arab Emirates Dirham} (2), \emph{Upside-Down Jellyfish} (35), \emph{Us Dollars} (3), \emph{Utility Knife} (10), \emph{Uv Sterilizer Vacuum} (26), \emph{Uv-Vis} (3).}\\

\noindent\scriptsize{\centering{\textbf{\normalsize{V (18 classes)}}}\\\emph{Vacuum Cleaner} (40), \emph{Vacuum Cleaner Brush Roller} (99), \emph{Vacuum Cleaner Crevice Tool Wheel} (18), \emph{Vampire Squid} (25), \emph{Vanilla Bean} (7), \emph{Vernier Caliper} (107), \emph{Vest} (8), \emph{Vial Adapter} (1), \emph{Vibration Meter} (34), \emph{Vietnamese Leaf Monkey} (8), \emph{Vintage Aircraft} (35), \emph{Vintage Car} (40), \emph{Violinist} (12), \emph{Virtual Reality Headset} (3), \emph{Virtual Reality Rides} (4), \emph{Viscometer} (50), \emph{Volleyball} (30), \emph{Volleyball Player} (49).}\\

\noindent\scriptsize{\centering{\textbf{\normalsize{W (60 classes)}}}\\\emph{Waffle Wheel} (110), \emph{Wagon Wheel} (1), \emph{Waiter} (35), \emph{Walkie Talkies} (31), \emph{Wall Decoration Painting} (12), \emph{Wall Scanner} (20), \emph{Wallaby} (34), \emph{Wallball} (30), \emph{Wallet} (30), \emph{Walnut} (3), \emph{Walrus} (28), \emph{Waltz Dancer} (23), \emph{Waspfish} (43), \emph{Water Bottle} (2), \emph{Water Filter Bottle} (29), \emph{Water Gun} (28), \emph{Water Painting Wall} (3), \emph{Water Play Table} (3), \emph{Water Polo} (18), \emph{Water Shooters} (10), \emph{Water Walking Balls} (10), \emph{Water Wheel} (34), \emph{Watering Truck} (35), \emph{Wave Slide} (1), \emph{Weddell Seal} (32), \emph{Wedges} (83), \emph{Weekender} (47), \emph{Weever Fish} (28), \emph{Weigthlifter} (34), \emph{Western Pygmy Marmoset} (2), \emph{Wetmorella Nigropinnata} (16), \emph{Whale Shark} (14), \emph{Wheat} (3), \emph{Wheelchair} (209), \emph{Wheelchair Wheel} (50), \emph{Whirling Whirlpools} (4), \emph{Whisk} (8), \emph{White-Cheeked Spider Monkey} (5), \emph{White-Faced Saki} (3), \emph{White-Fronted Tern} (9), \emph{White-Headed Marmoset} (3), \emph{White-Lipped Deer} (15), \emph{White-Lipped Tamarin} (3), \emph{White-Naped Crane} (43), \emph{White-Nosed Saki} (2), \emph{White-Tailed Rubythroat} (35), \emph{White-Throated Rock Thrush} (10), \emph{Whooper Swan} (50), \emph{Wide-Brimmed Hat} (3), \emph{Wieds Marmoset} (3), \emph{Wild Boar} (33), \emph{Wind Speed Gauge} (4), \emph{Wine Glass} (49), \emph{Wire Stripper} (3), \emph{Wok} (40), \emph{Wolverine} (25), \emph{Workwear} (7), \emph{Wristwatch Wheel} (9), \emph{Writing Brush} (29), \emph{Wurlitzer Piano} (32).}\\

\noindent\scriptsize{\centering{\textbf{\normalsize{X (3 classes)}}}\\\emph{Xema Sabini} (4), \emph{Xm109} (1), \emph{Xylophone} (32).}\\

\noindent\scriptsize{\centering{\textbf{\normalsize{Y (17 classes)}}}\\\emph{Yangqin} (31), \emph{Yellow Crested Weedfish} (14), \emph{Yellow Peach} (23), \emph{Yellow Warbler} (34), \emph{Yellow-Bellied Tit} (34), \emph{Yellow-Billed Magpie} (33), \emph{Yellow-Cheeked Tit} (45), \emph{Yellowfin Flasher Wrasse} (8), \emph{Yellowmargin Triggerfish} (28), \emph{Yellow-Rumped Warbler} (31), \emph{Yellowtail Damselfish} (35), \emph{Yellow-Throated Marten} (35), \emph{Yellow-Throated Warbler} (5), \emph{Yoga Mat Bag} (42), \emph{Yoga Wheel} (34), \emph{Yo-Yo} (30), \emph{Y-Site Injection Port} (1).}\\

\noindent\scriptsize{\centering{\textbf{\normalsize{Z (9 classes)}}}\\\emph{Zebra} (33), \emph{Zebra Danio} (34), \emph{Zebra Finch} (45), \emph{Zebra Loach} (47), \emph{Zebra Pleco} (45), \emph{Zebra Shark} (40), \emph{Zhanmadao} (3), \emph{Zhongshan Suit} (3), \emph{Zither} (32).}\\

\begin{figure}[!t]
	\centering
	\includegraphics[width=\linewidth]{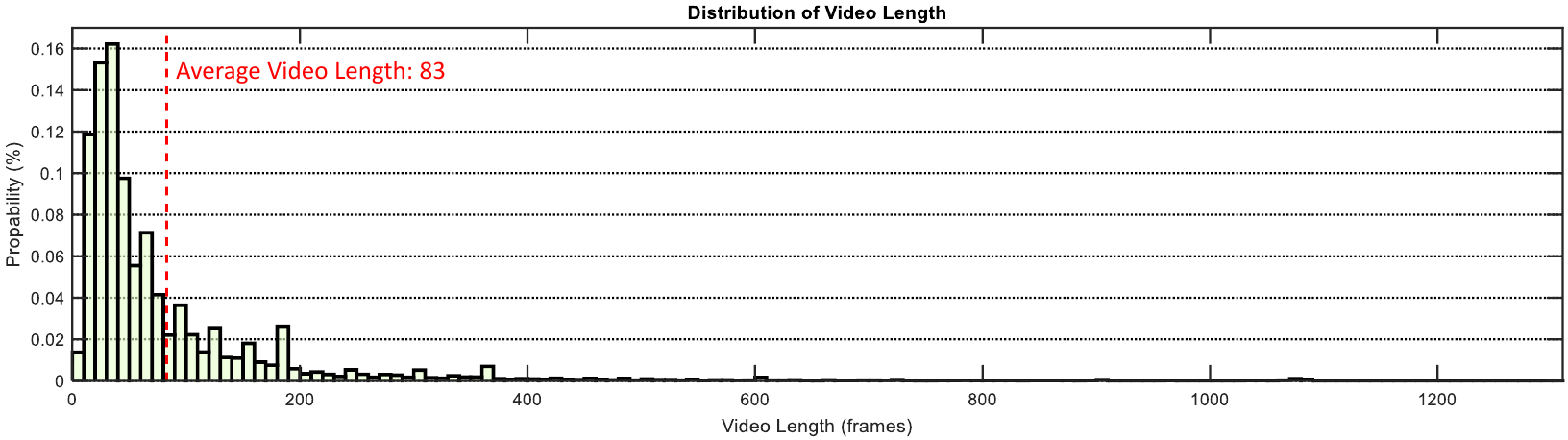}
	\caption{Distribution of sequence length on VastTrack. The average video length of our VastTrack is 83 frames. Please notice that, VastTrack is focused on short-term tracking by offering abundant classes and videos, but could also be used for training long-term temporal trackers as discussed in the main text.}
	\label{fig:seqlength}
\end{figure}

\begin{figure}[!t]
	\centering
	\includegraphics[width=\linewidth]{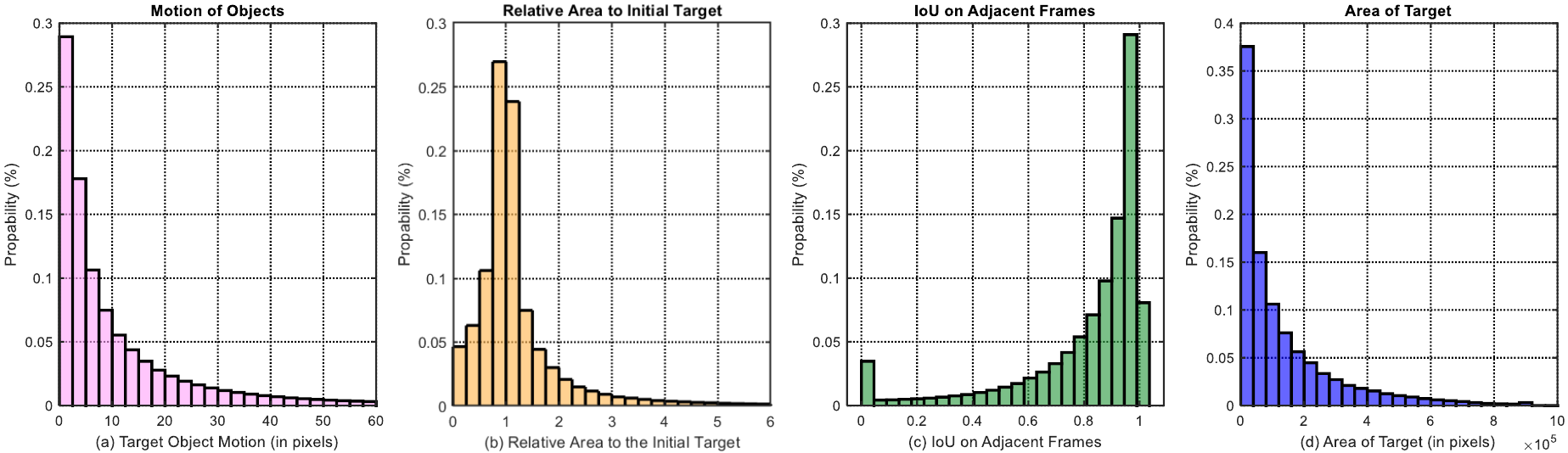}
	\caption{Statistics of annotations on object motion (image (a)), relative area compared to the initial object (image (b)), IoU of targets in adjacent frames (image (c)), and size of targets (image (d)).}
	\label{fig:sta}
\end{figure}

\begin{figure}[!t]
	\centering
	\includegraphics[width=0.19\linewidth]{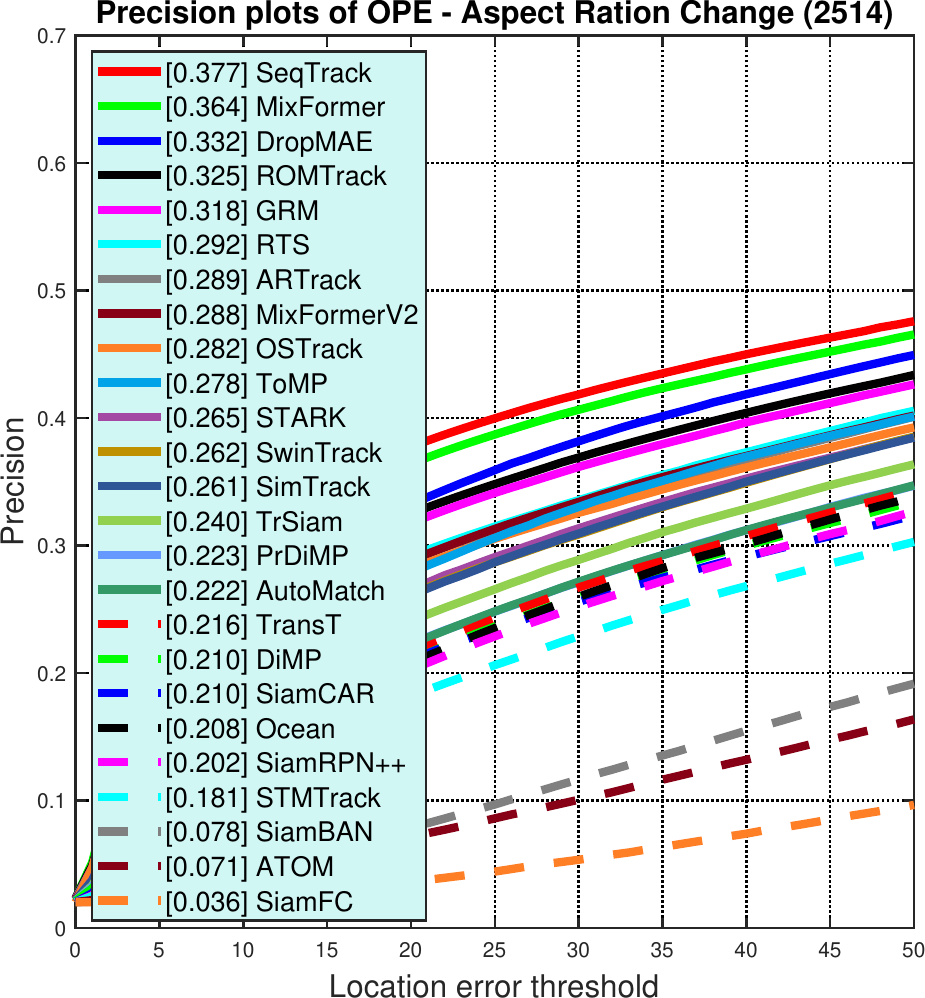}
        \includegraphics[width=0.19\linewidth]{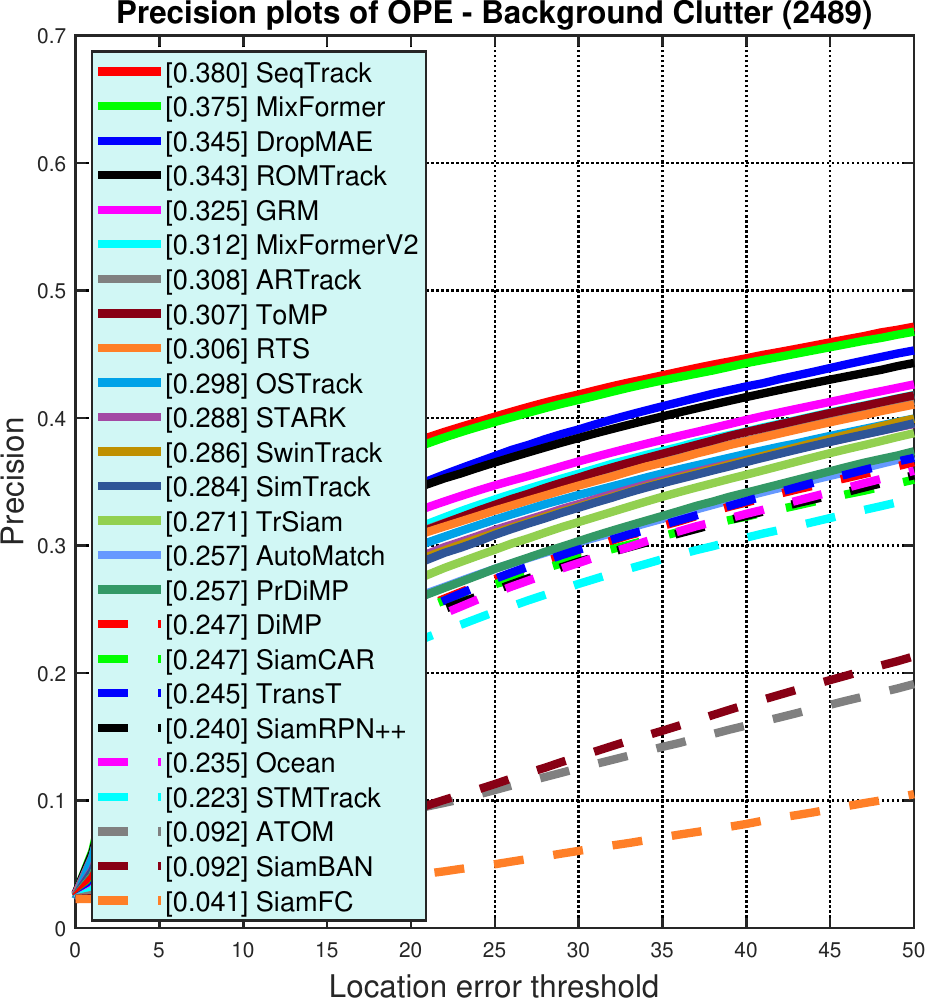}
	\includegraphics[width=0.19\linewidth]{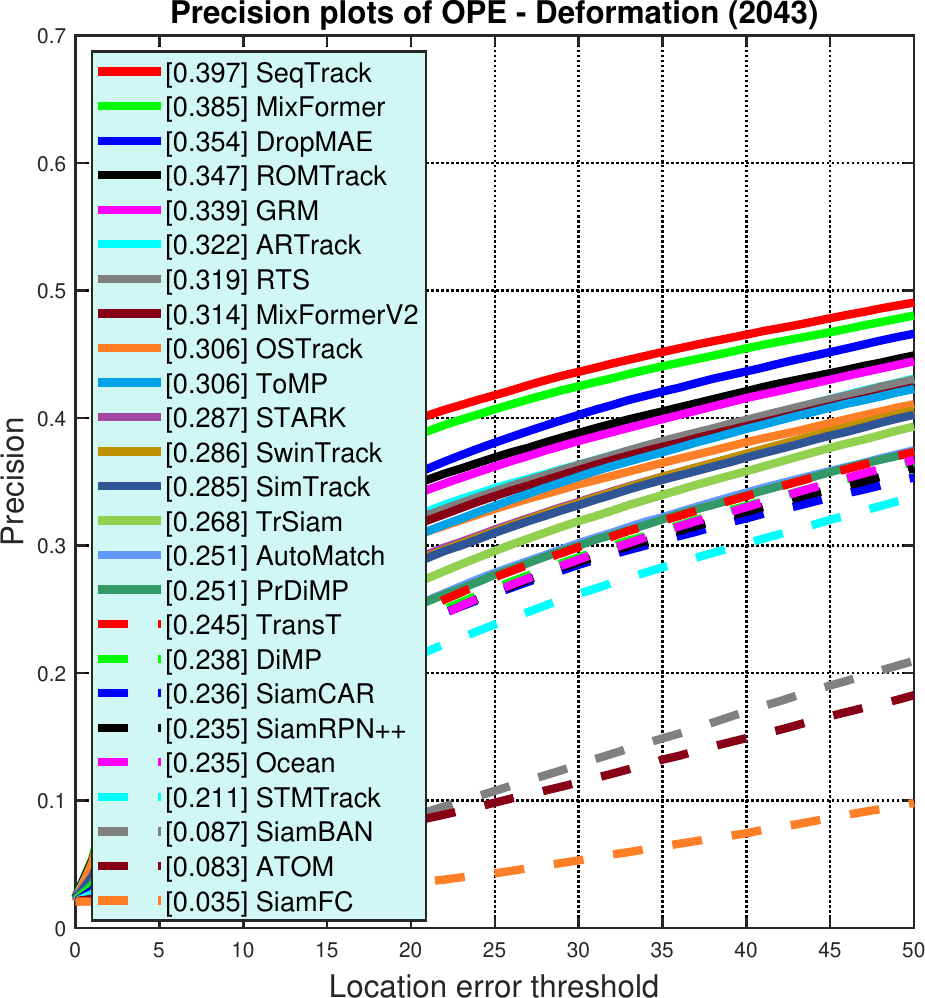}
        \includegraphics[width=0.19\linewidth]{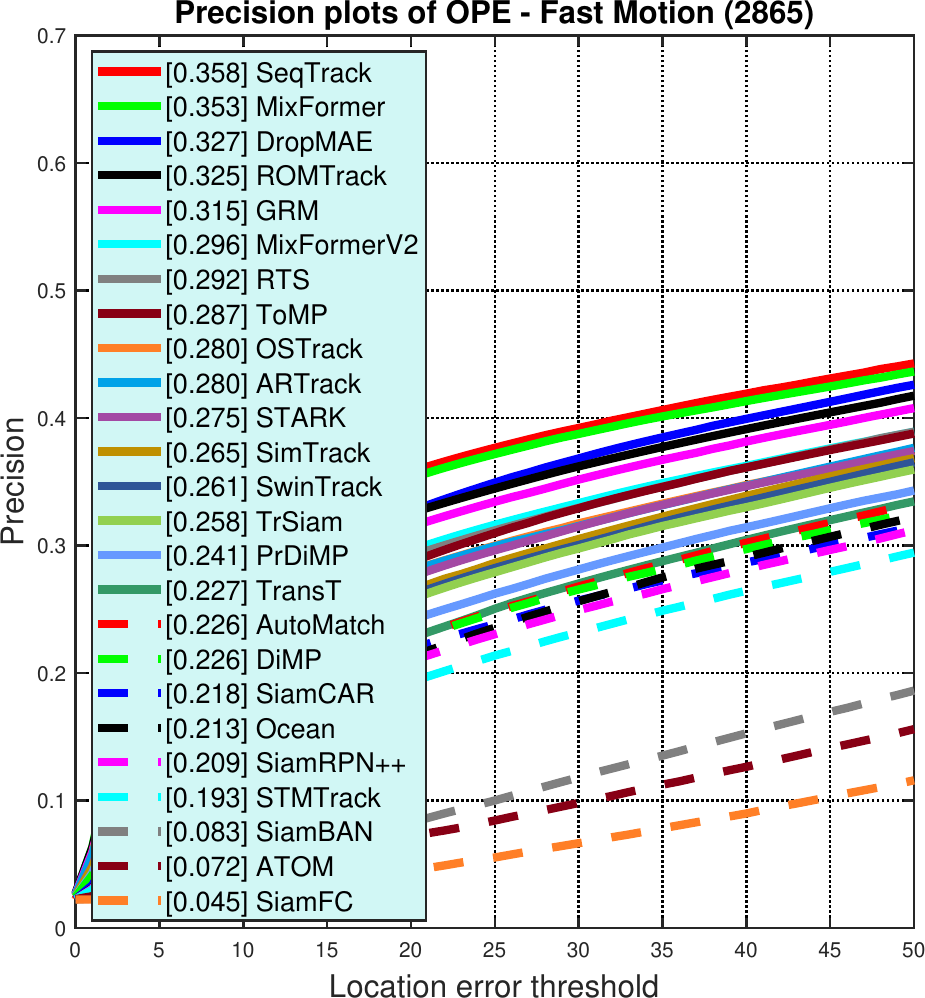}
	\includegraphics[width=0.19\linewidth]{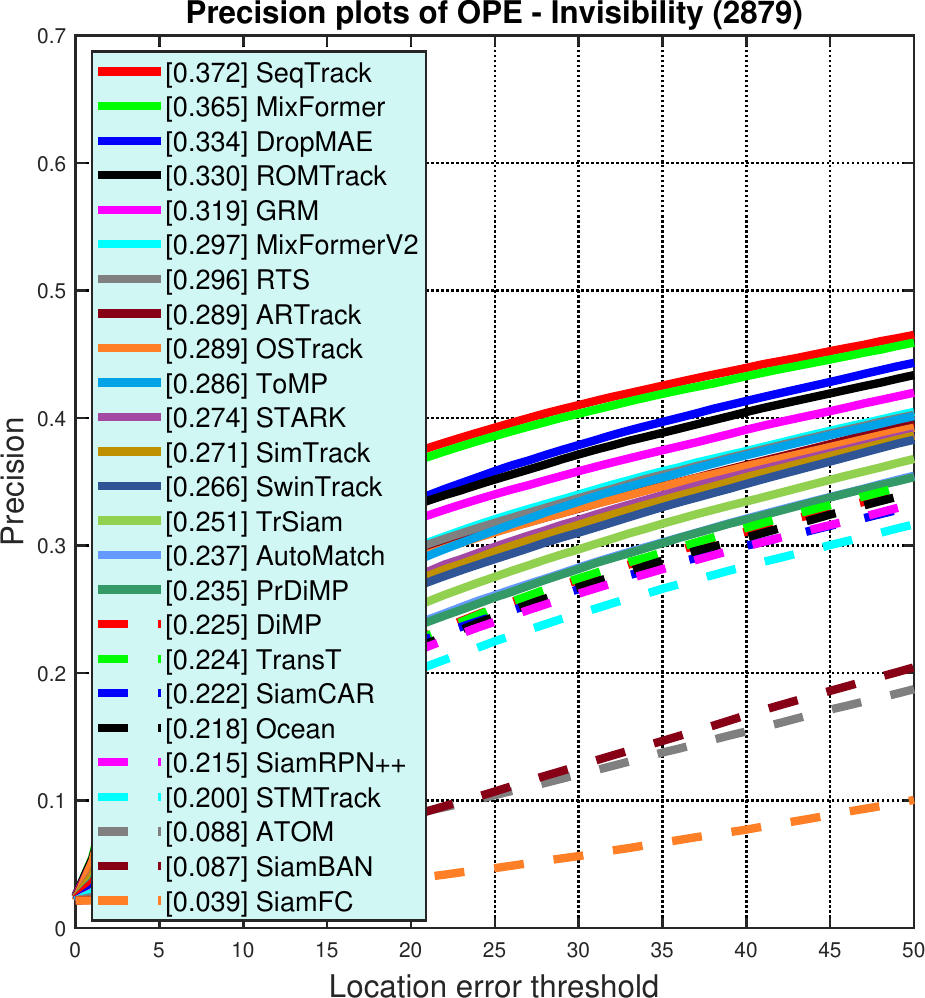}\\
        \includegraphics[width=0.19\linewidth]{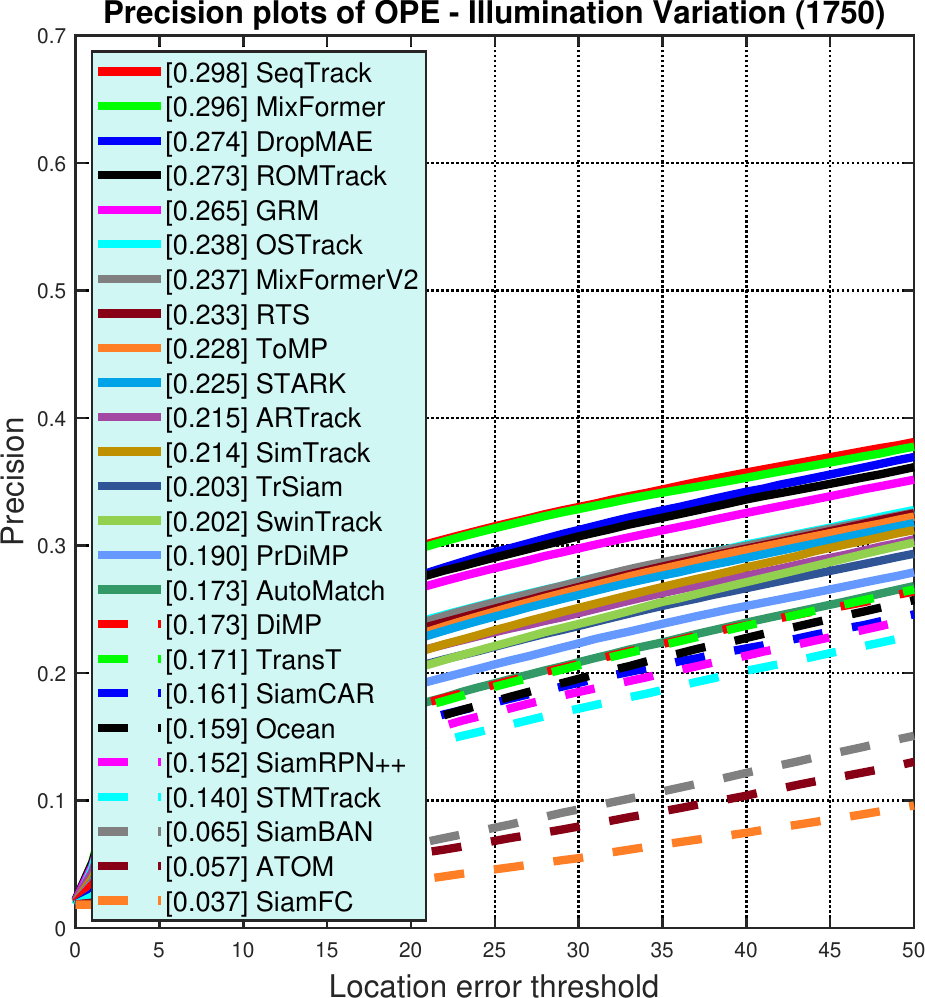}
	\includegraphics[width=0.19\linewidth]{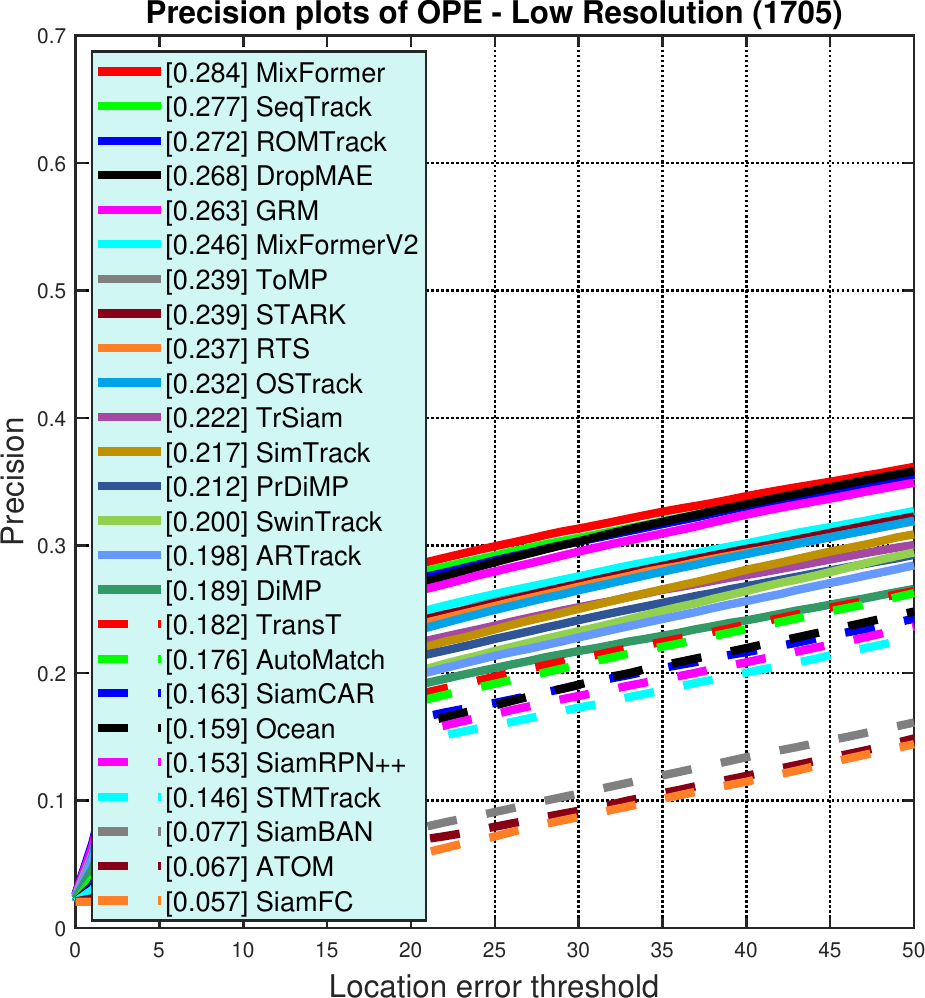}
        \includegraphics[width=0.19\linewidth]{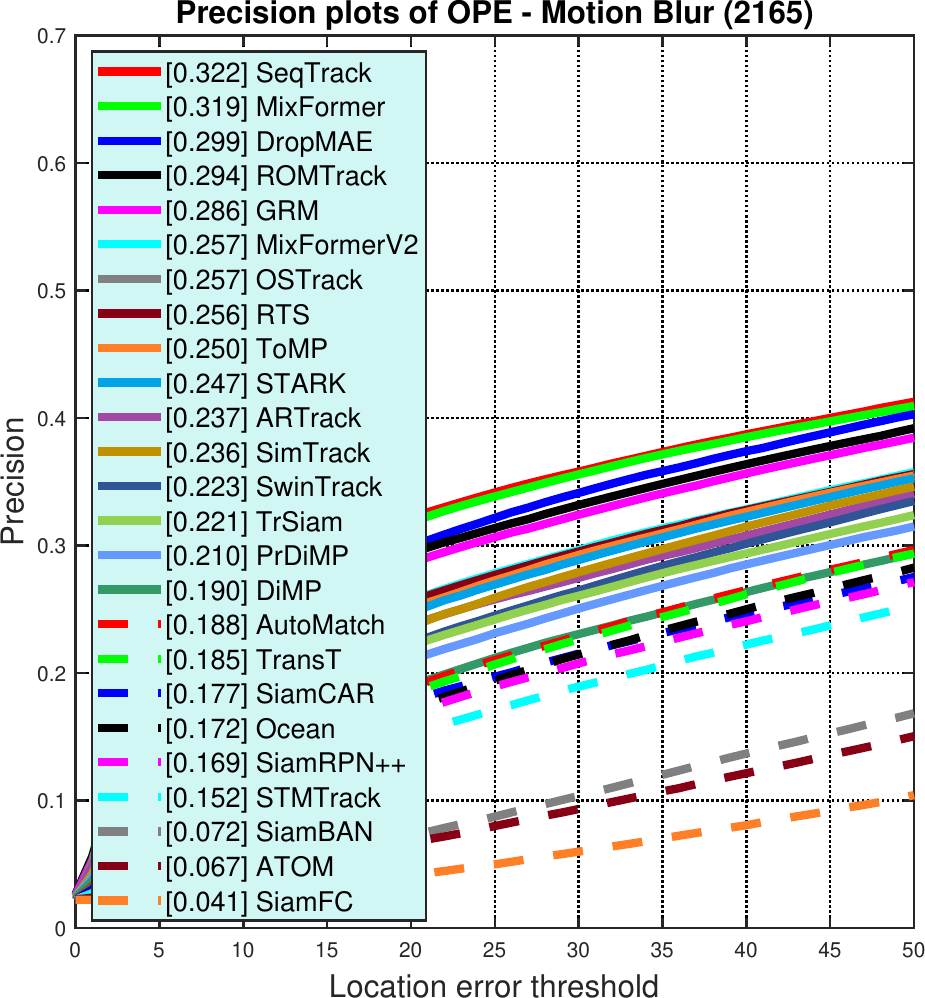}
	\includegraphics[width=0.19\linewidth]{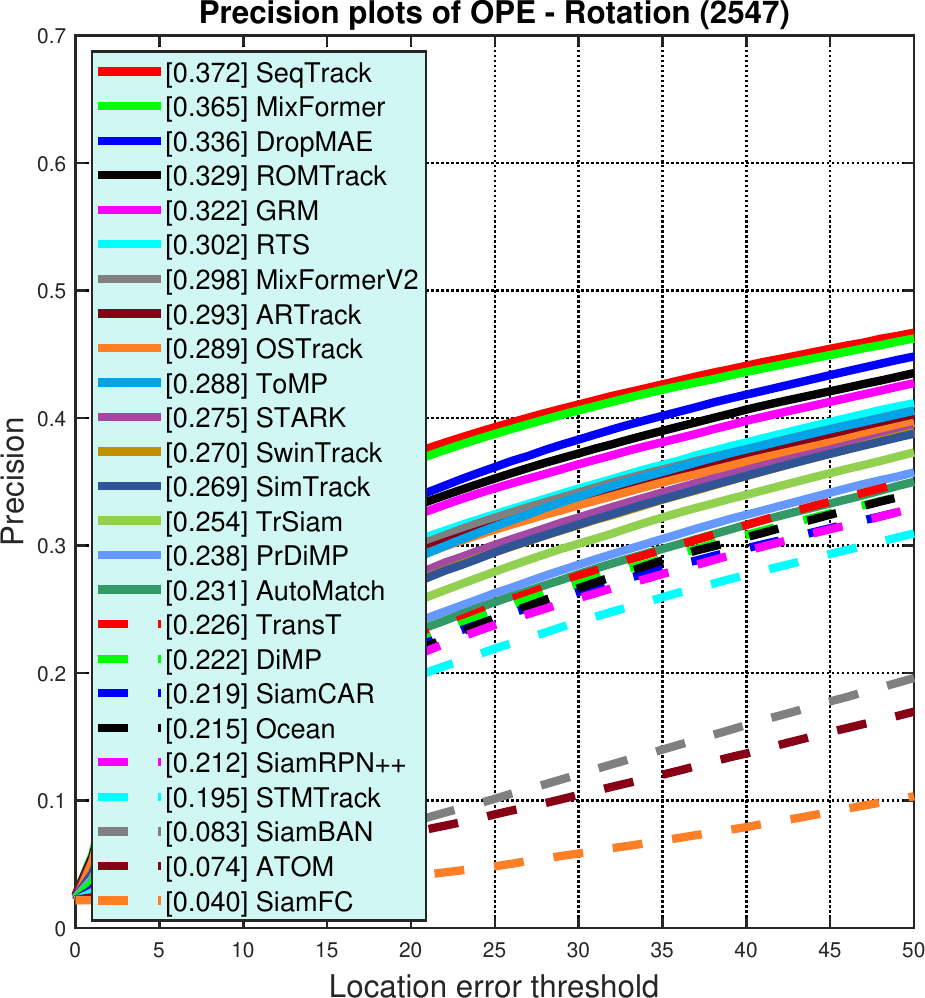}
        \includegraphics[width=0.19\linewidth]{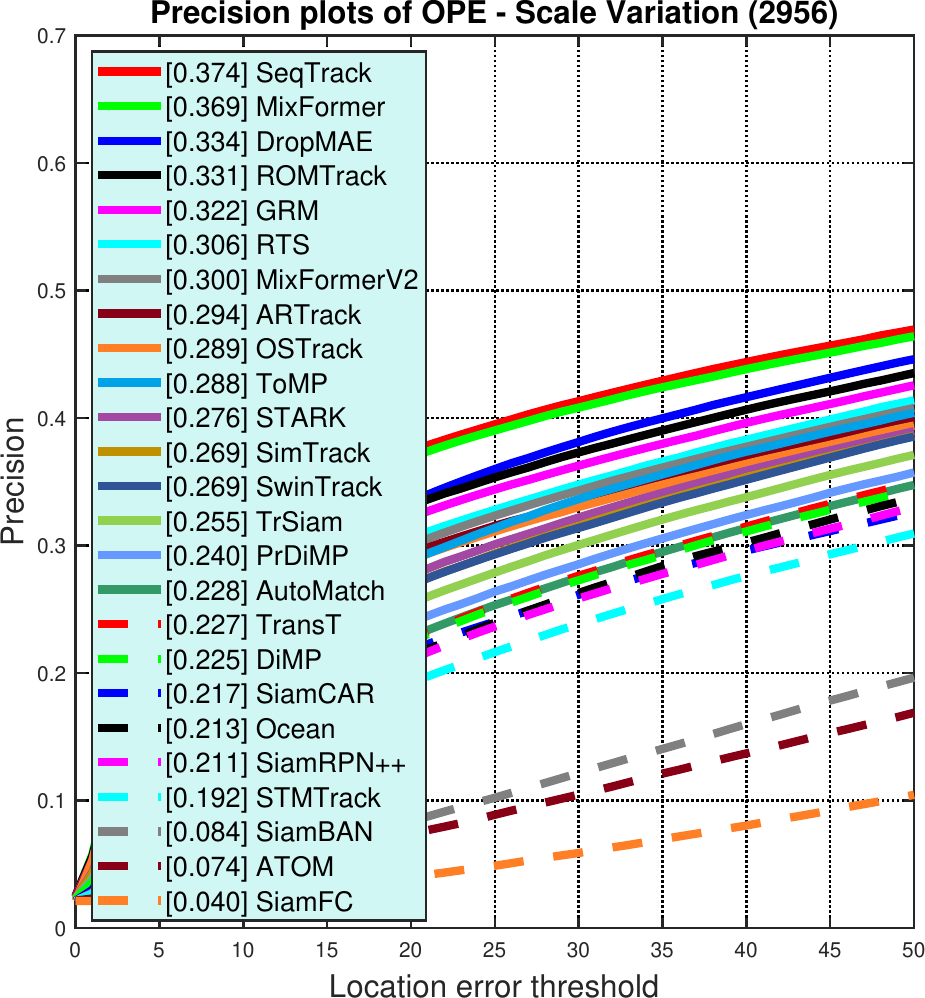}
	\caption{Performance of trackers on ten attributes using precision (PRE).}
	\label{fig:full_att_pre}
\end{figure}

\subsection*{S2 \;  Statistics of Sequence Length and Annotation Boxes}

\normalsize{To} better understand the features of VastTrack, we further show representative statistics of sequence length and annotation boxes. Fig.~\ref{fig:seqlength} shows the distribution of sequence length on VastTrack. Notice that, VastTrack has an average video length of 83 frames, and it is mainly focused on short-term tracking by offering abundant categories and sequences, but can also be used for long-term tracking. 

In Fig.~\ref{fig:sta}, we demonstrate the distributions of target object motion, relative area to the initial object, Intersection over Union (IoU) between the target objects in adjacent frames, and the size of object. From these statistics, we can see that objects moves fast and varies rapidly in the videos of VastTrack.

\begin{figure}[!t]
	\centering
	\includegraphics[width=0.19\linewidth]{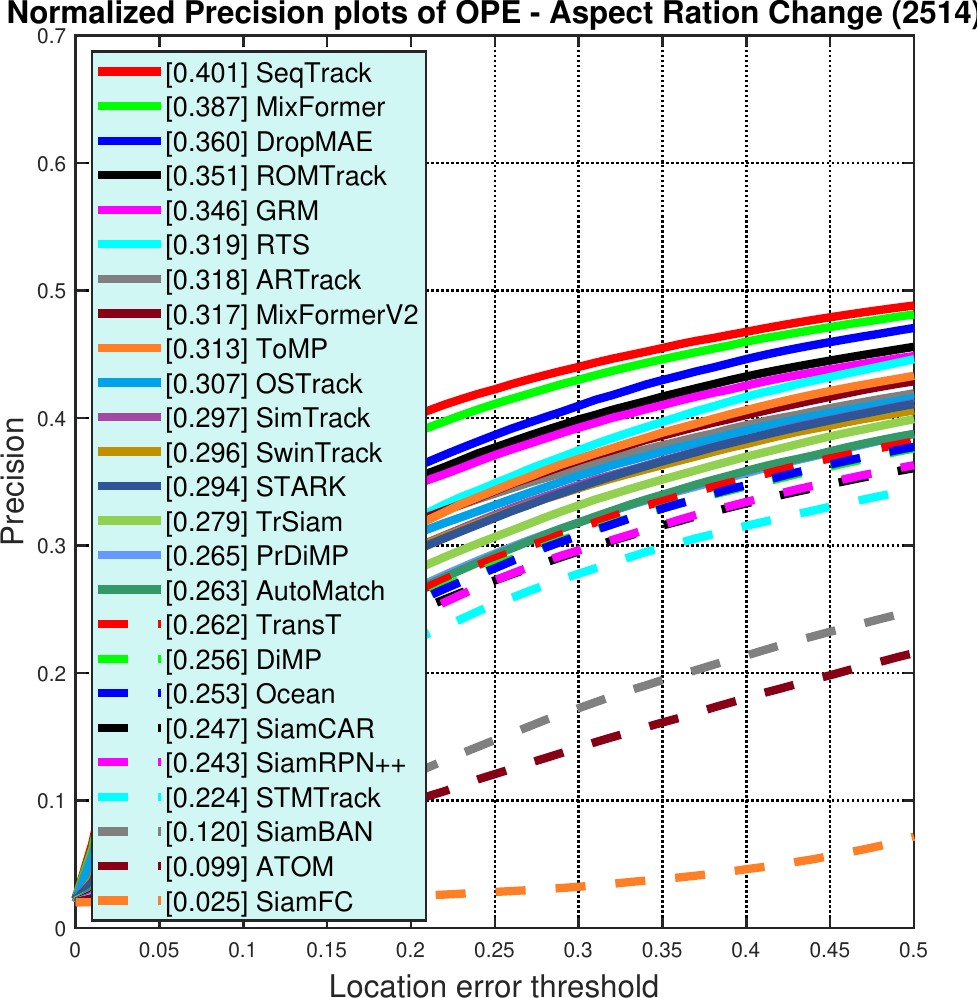}
        \includegraphics[width=0.19\linewidth]{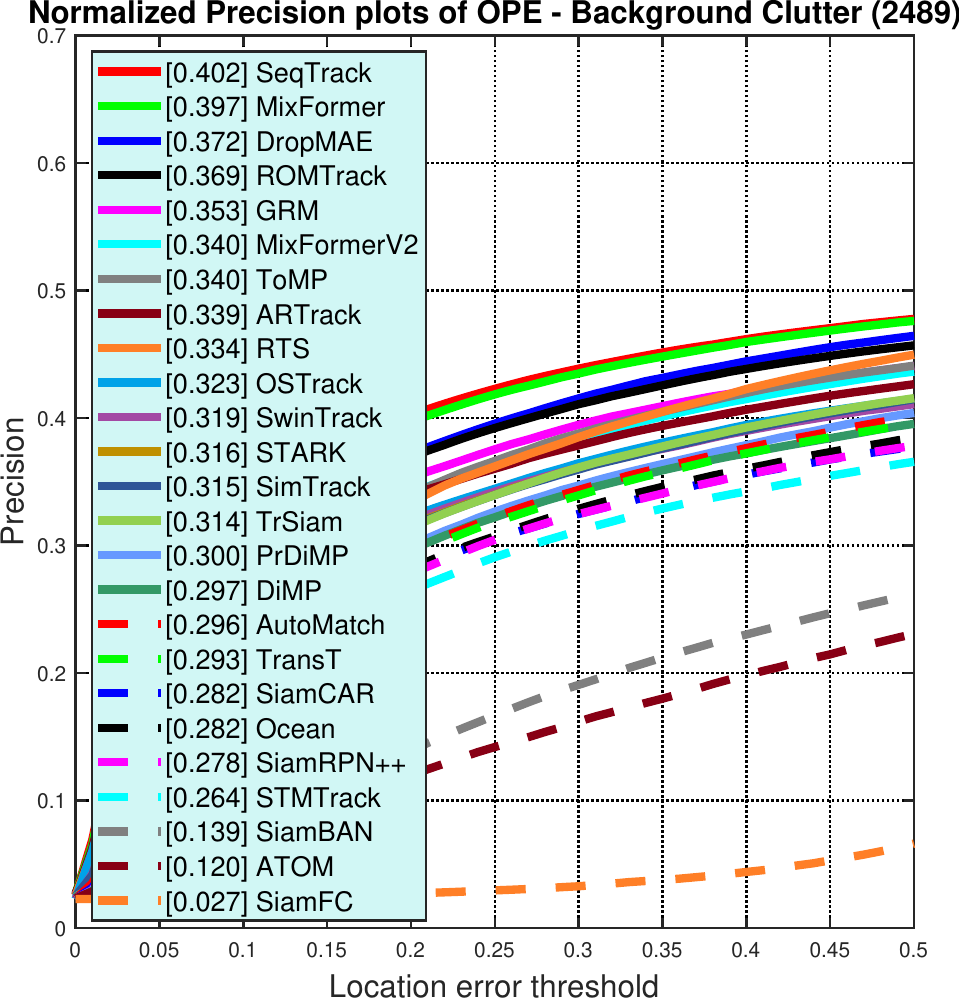}
	\includegraphics[width=0.19\linewidth]{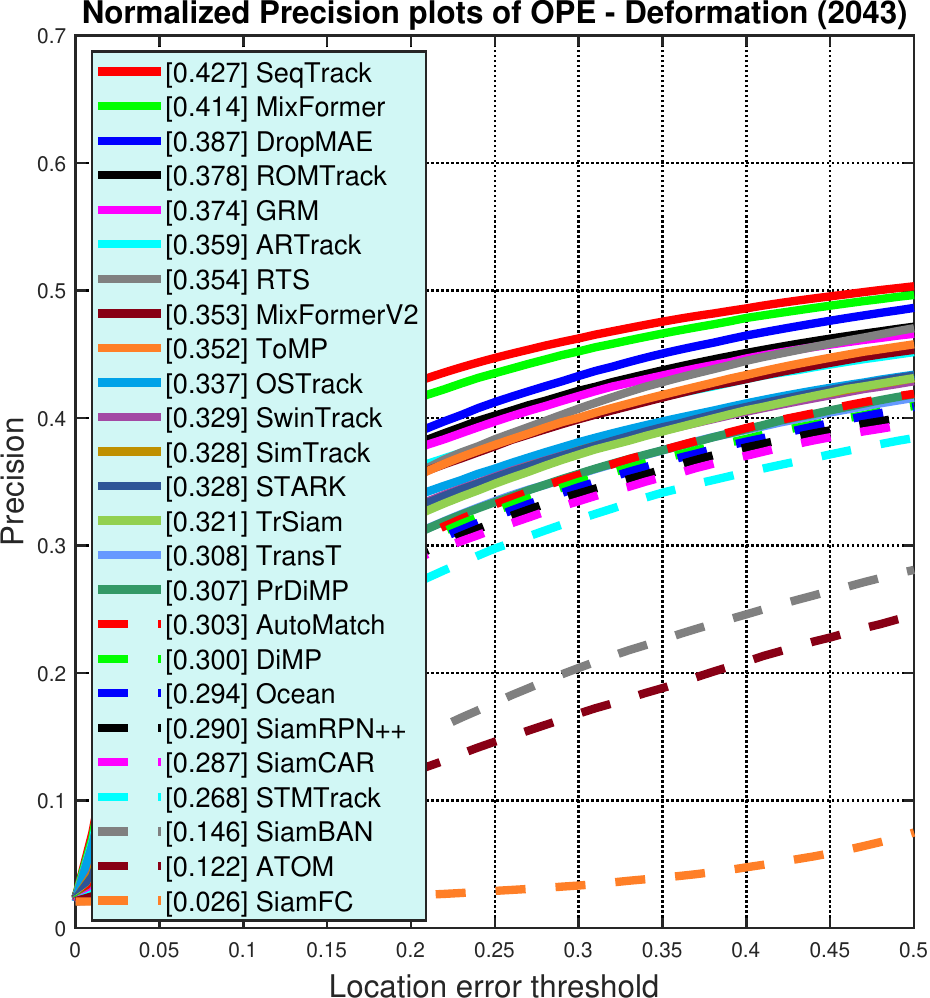}
        \includegraphics[width=0.19\linewidth]{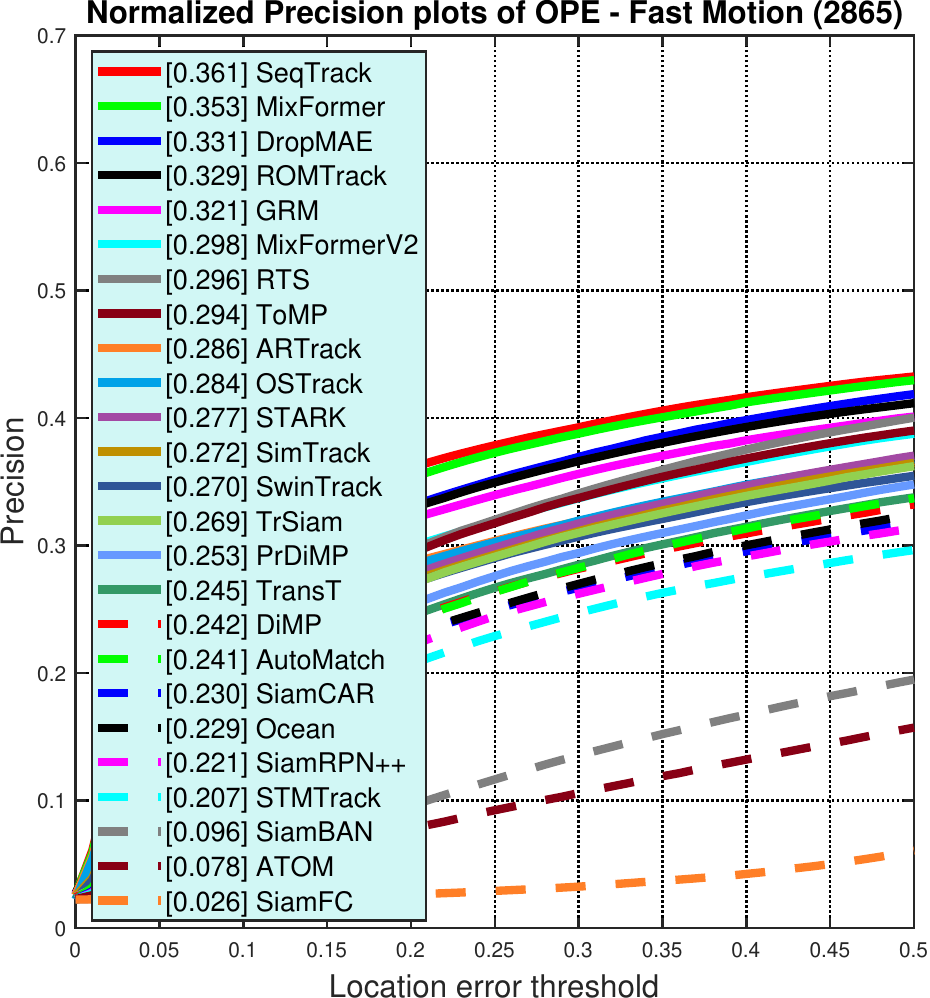}
	\includegraphics[width=0.19\linewidth]{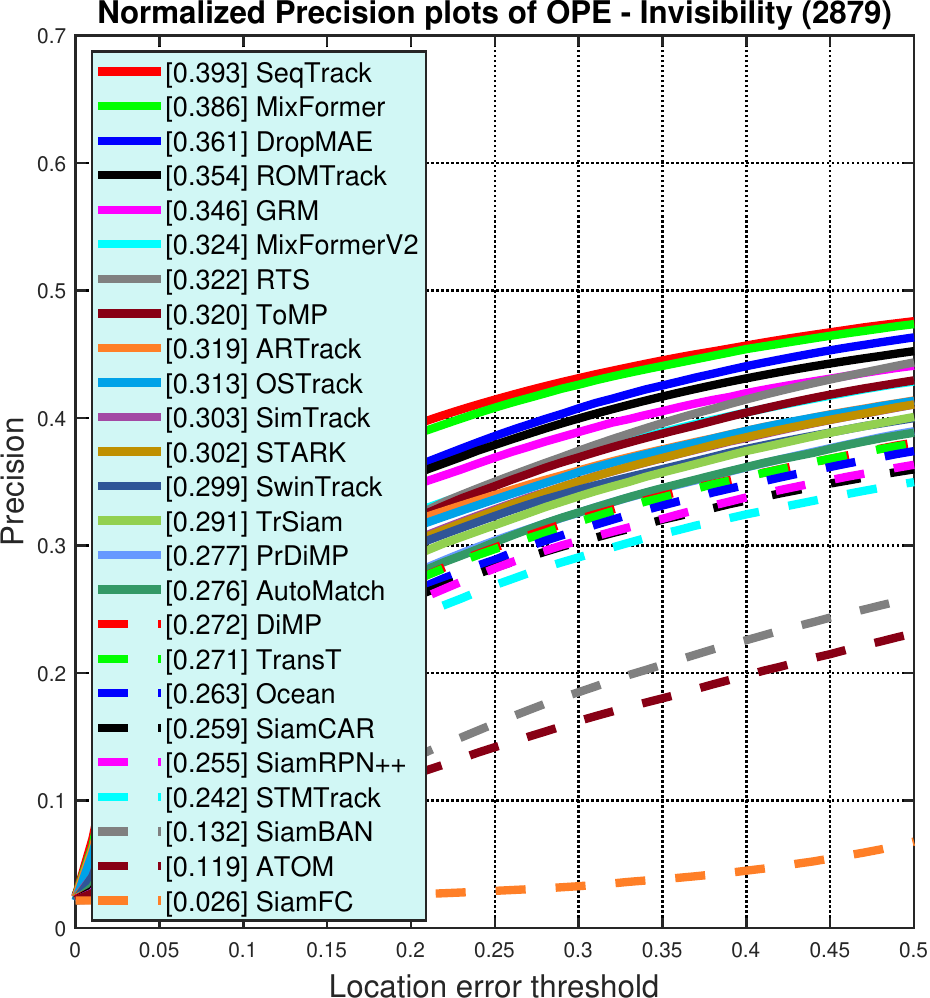}\\
        \includegraphics[width=0.19\linewidth]{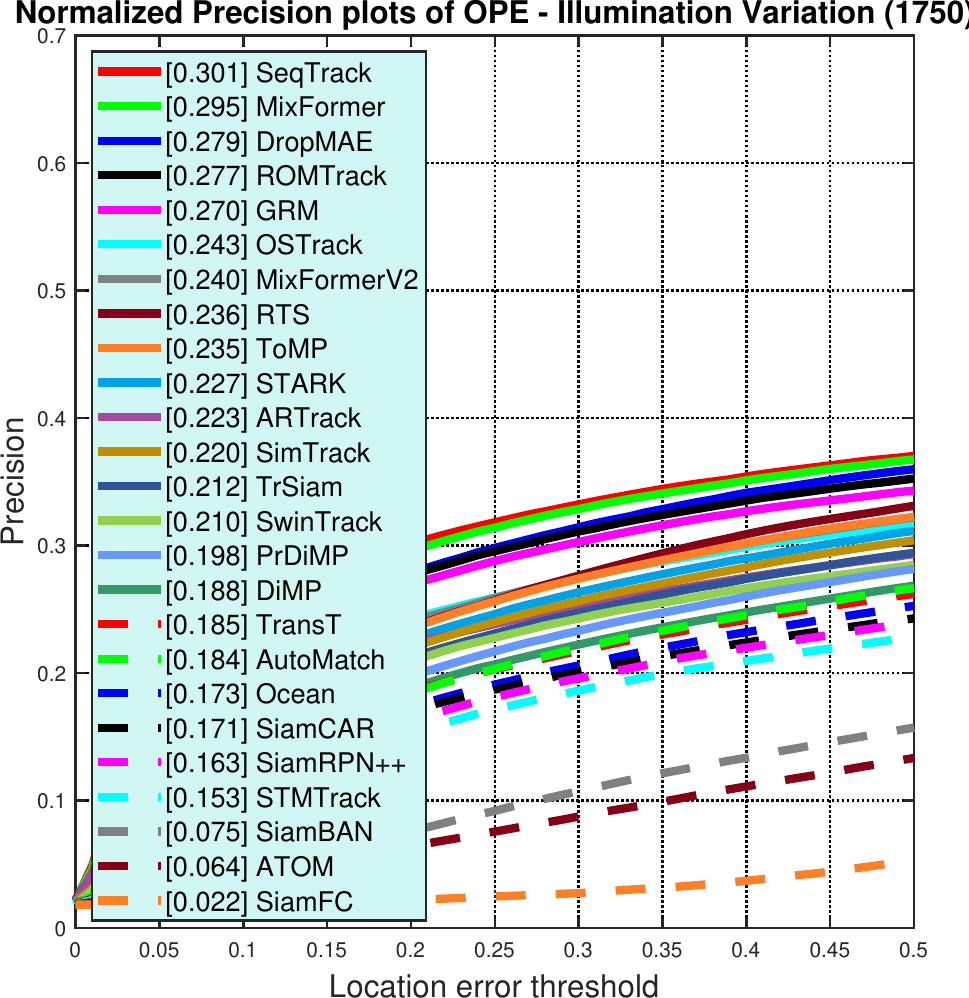}
	\includegraphics[width=0.19\linewidth]{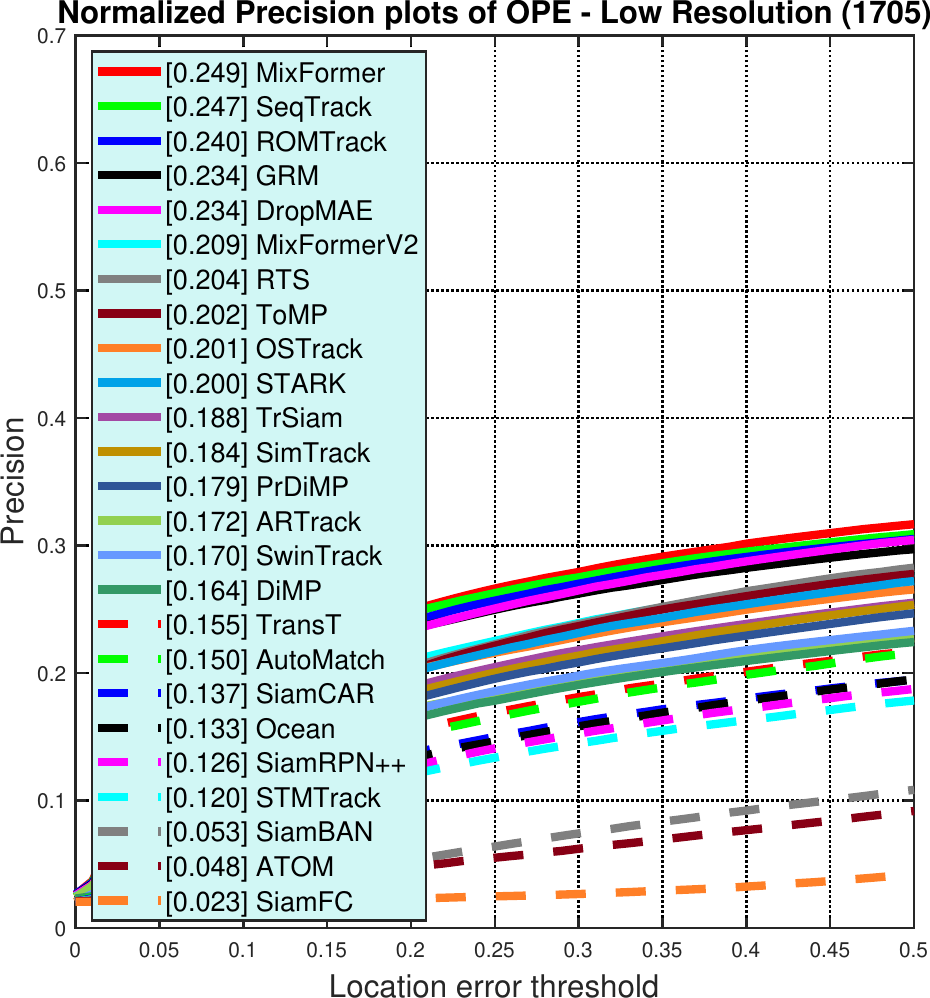}
        \includegraphics[width=0.19\linewidth]{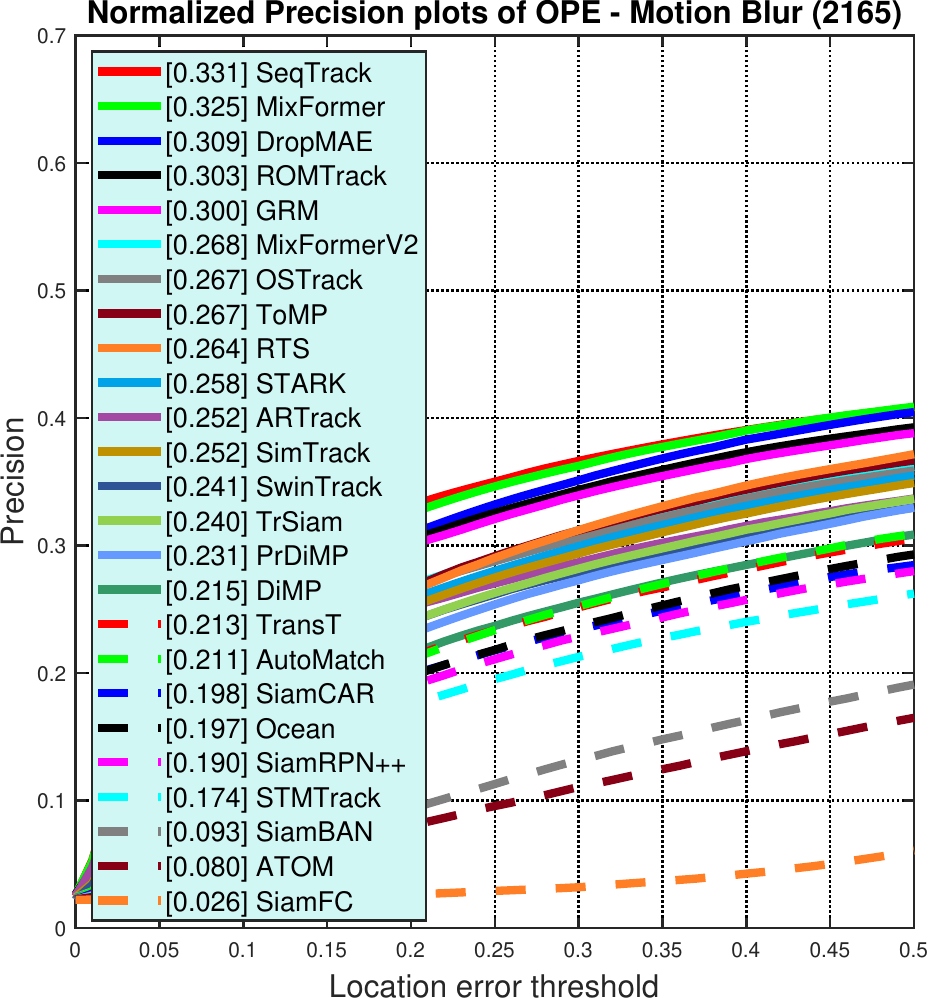}
	\includegraphics[width=0.19\linewidth]{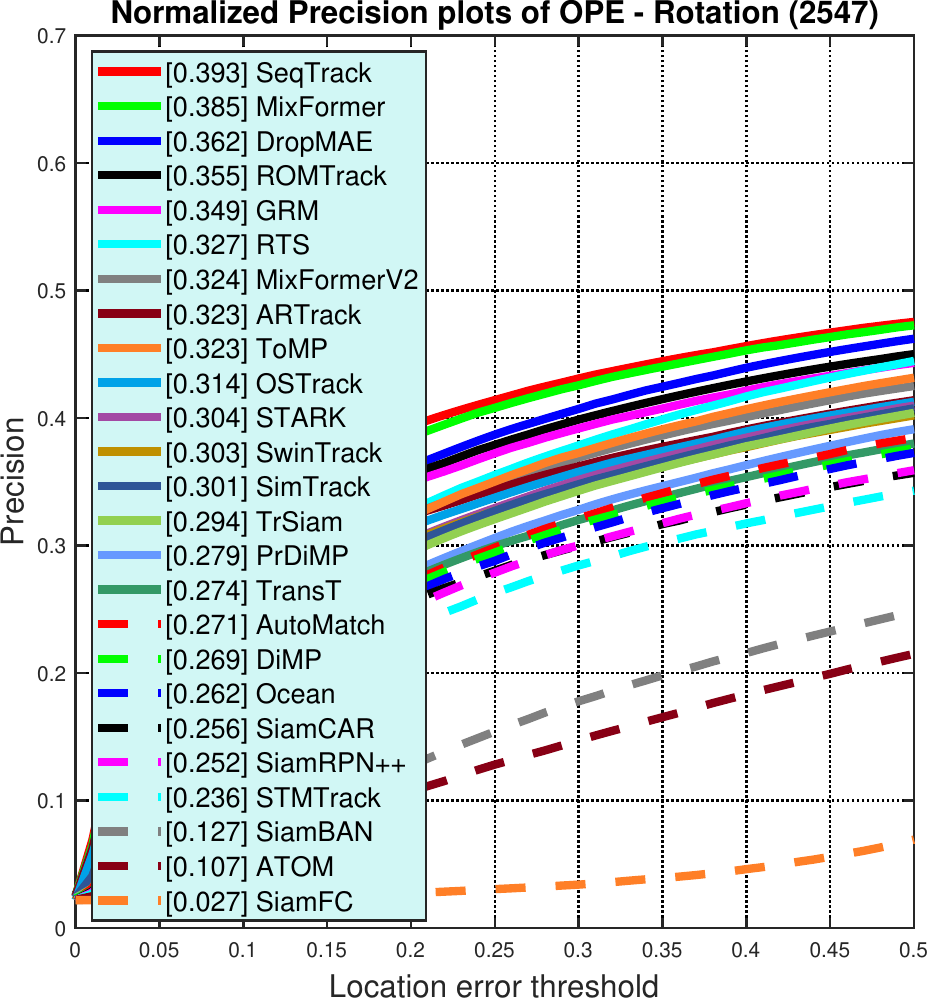}
        \includegraphics[width=0.19\linewidth]{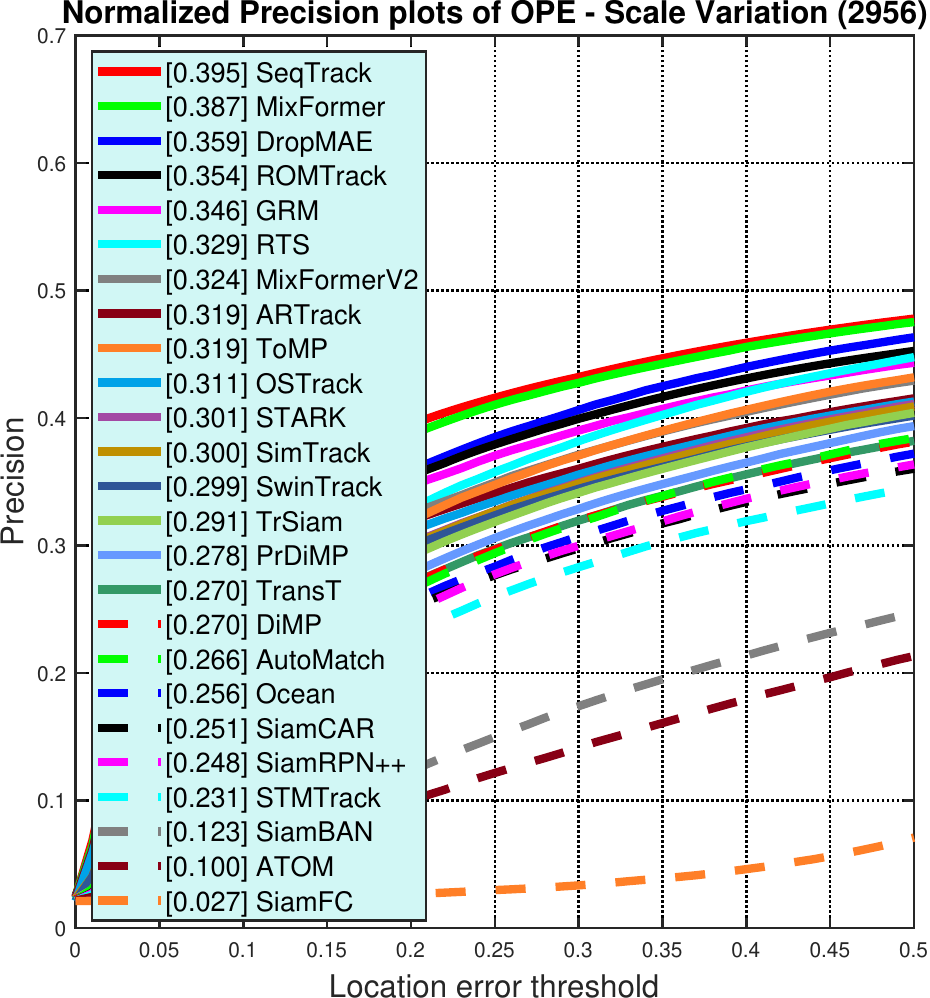}
	\caption{Performance of trackers on ten attributes using normalized precision (NPRE).}
	\label{fig:full_att_npre}
\end{figure}

\begin{figure}[!t]
	\centering
	\includegraphics[width=0.19\linewidth]{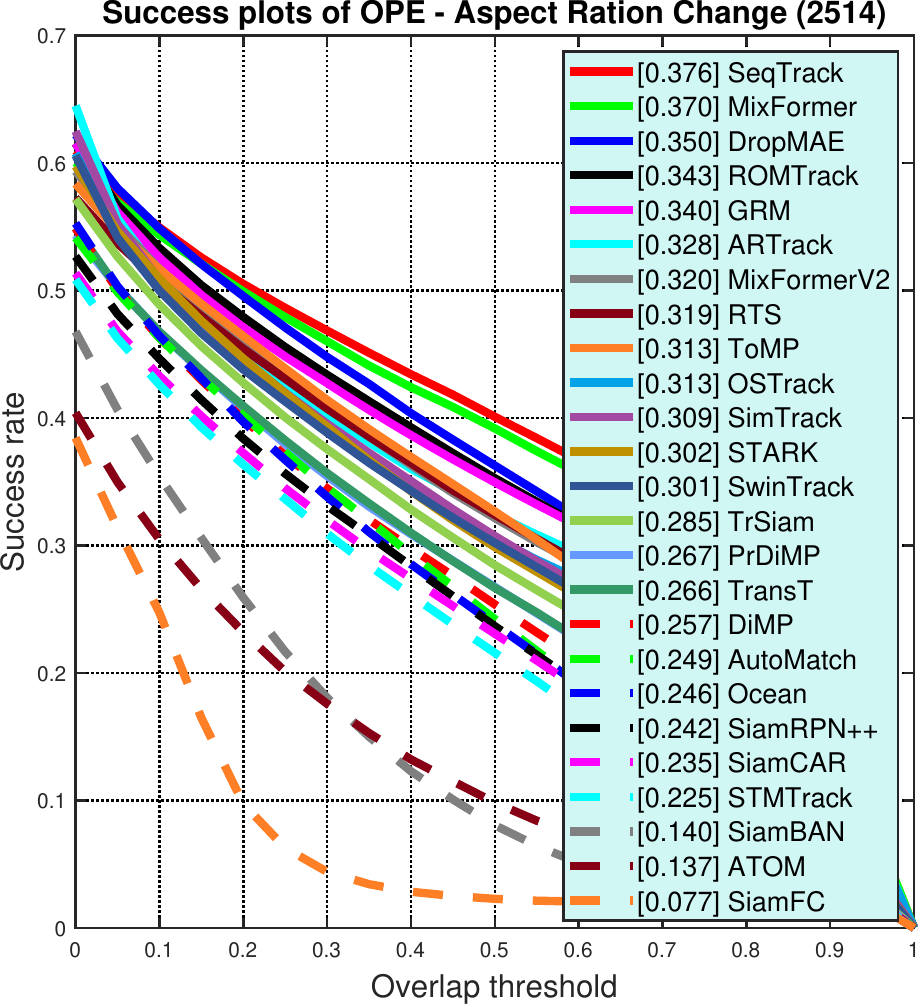}
        \includegraphics[width=0.19\linewidth]{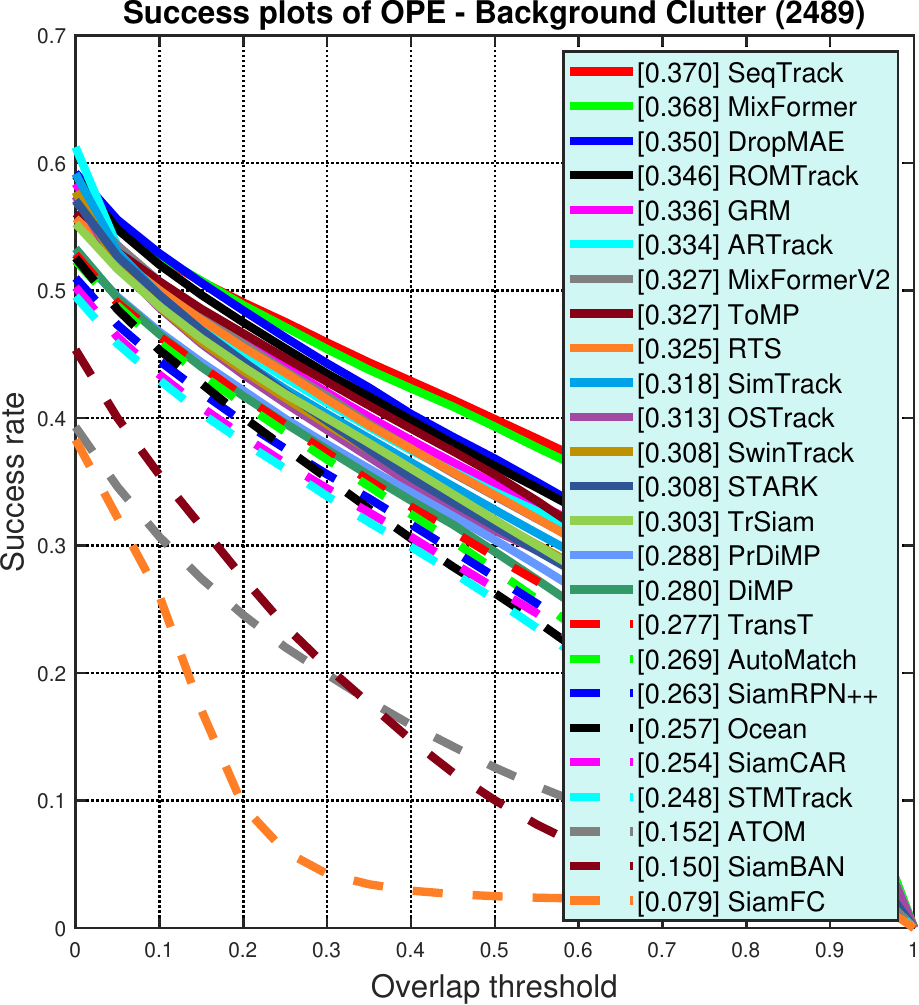}
	\includegraphics[width=0.19\linewidth]{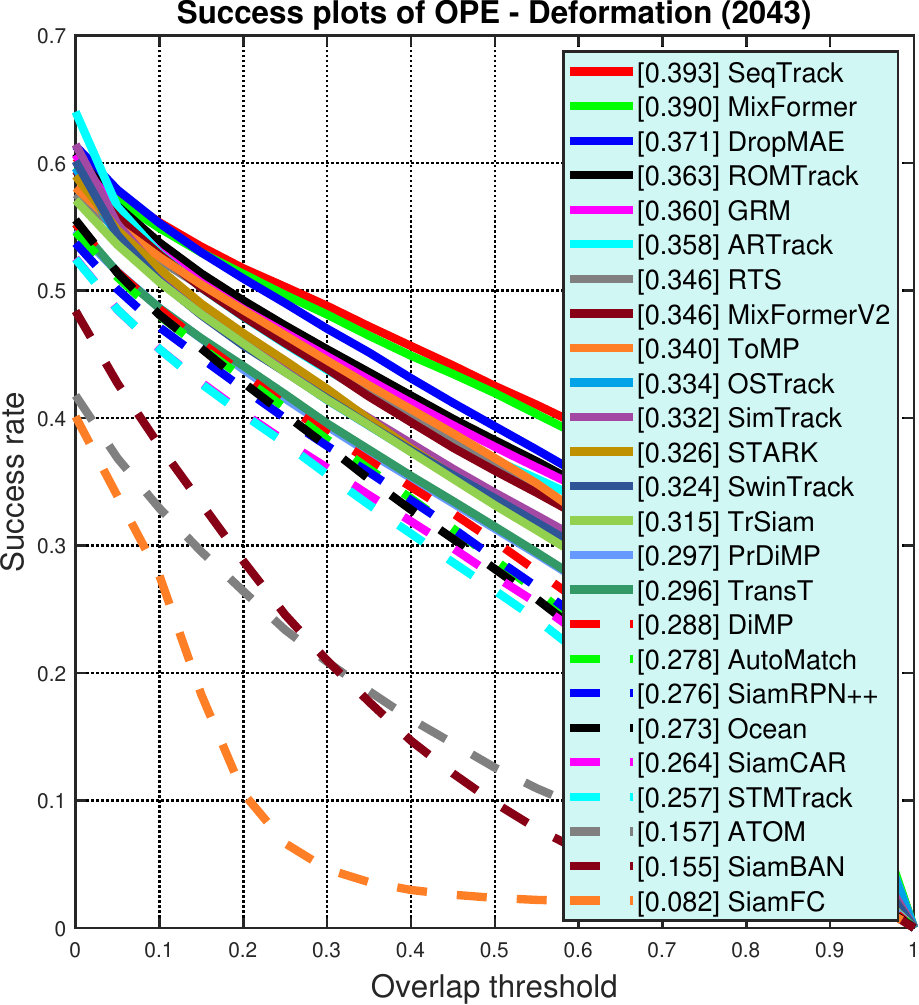}
        \includegraphics[width=0.19\linewidth]{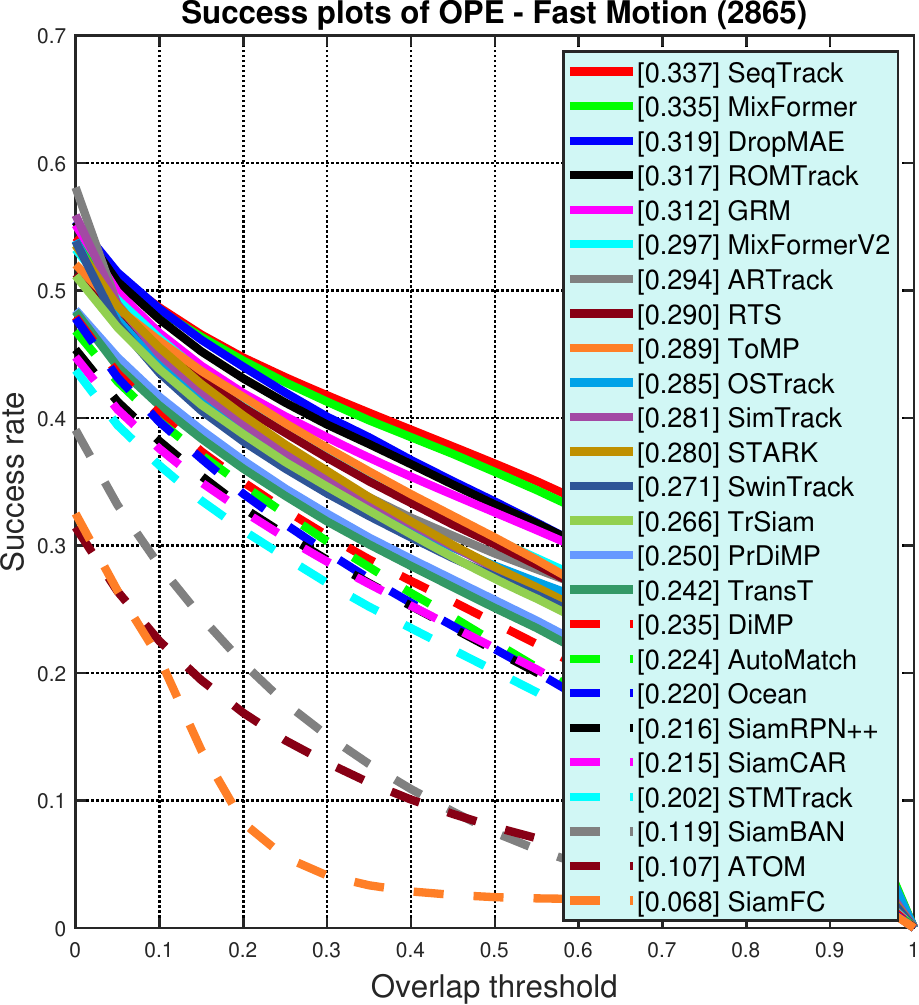}
	\includegraphics[width=0.19\linewidth]{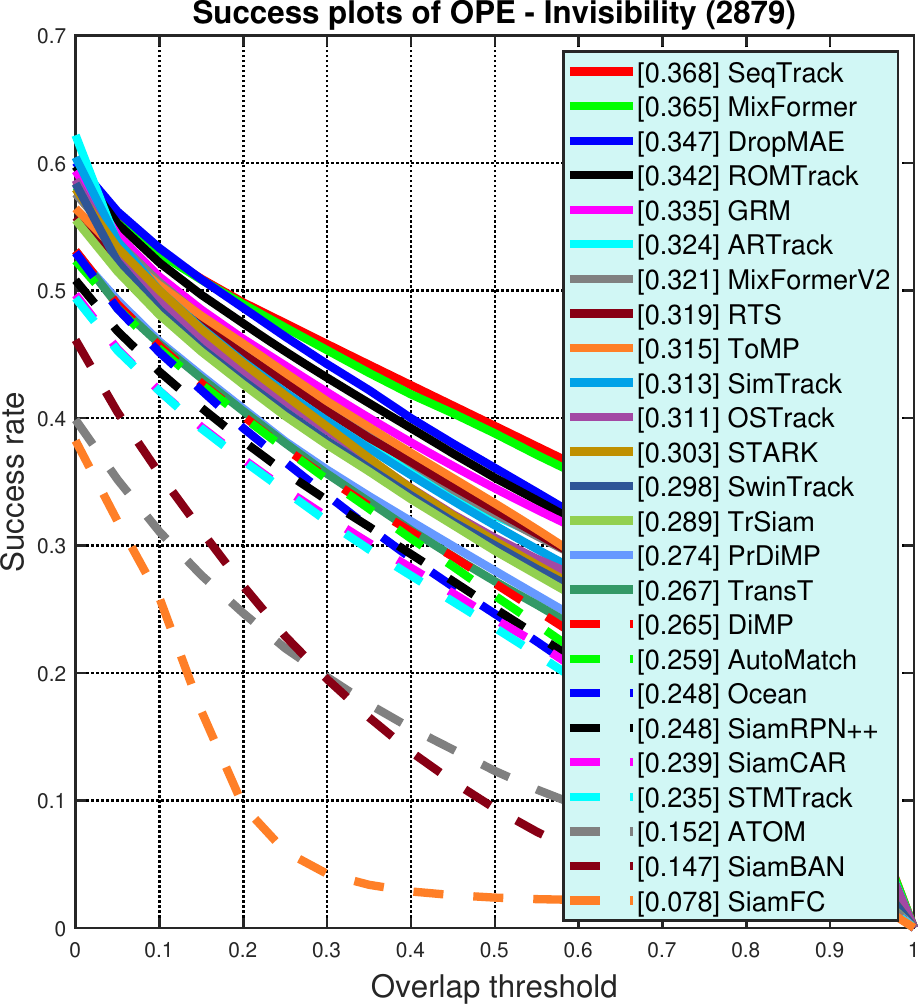}\\
        \includegraphics[width=0.19\linewidth]{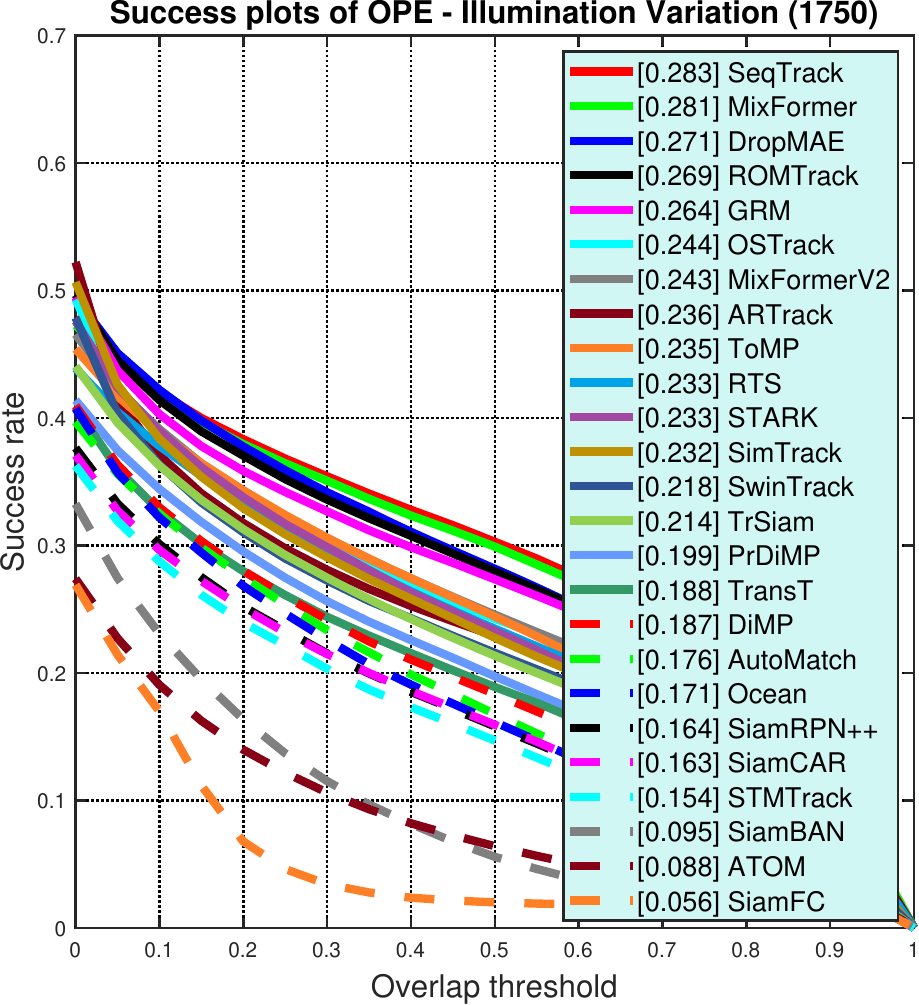}
	\includegraphics[width=0.19\linewidth]{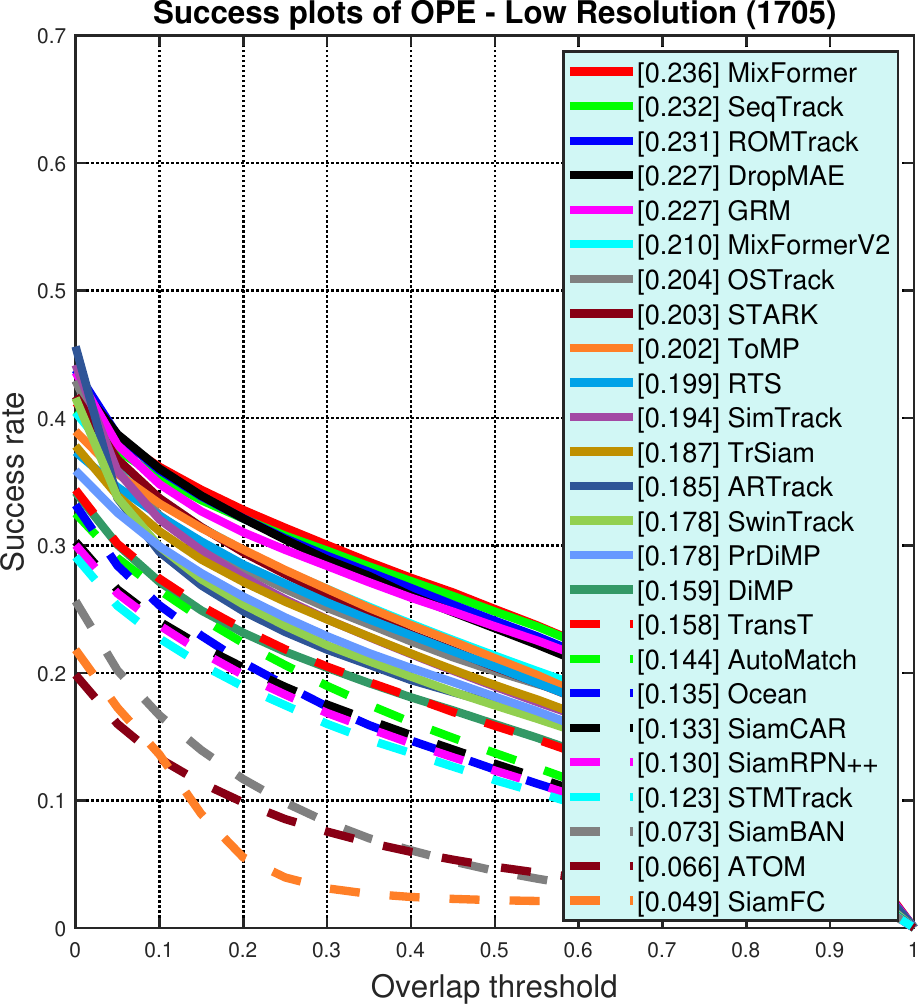}
        \includegraphics[width=0.19\linewidth]{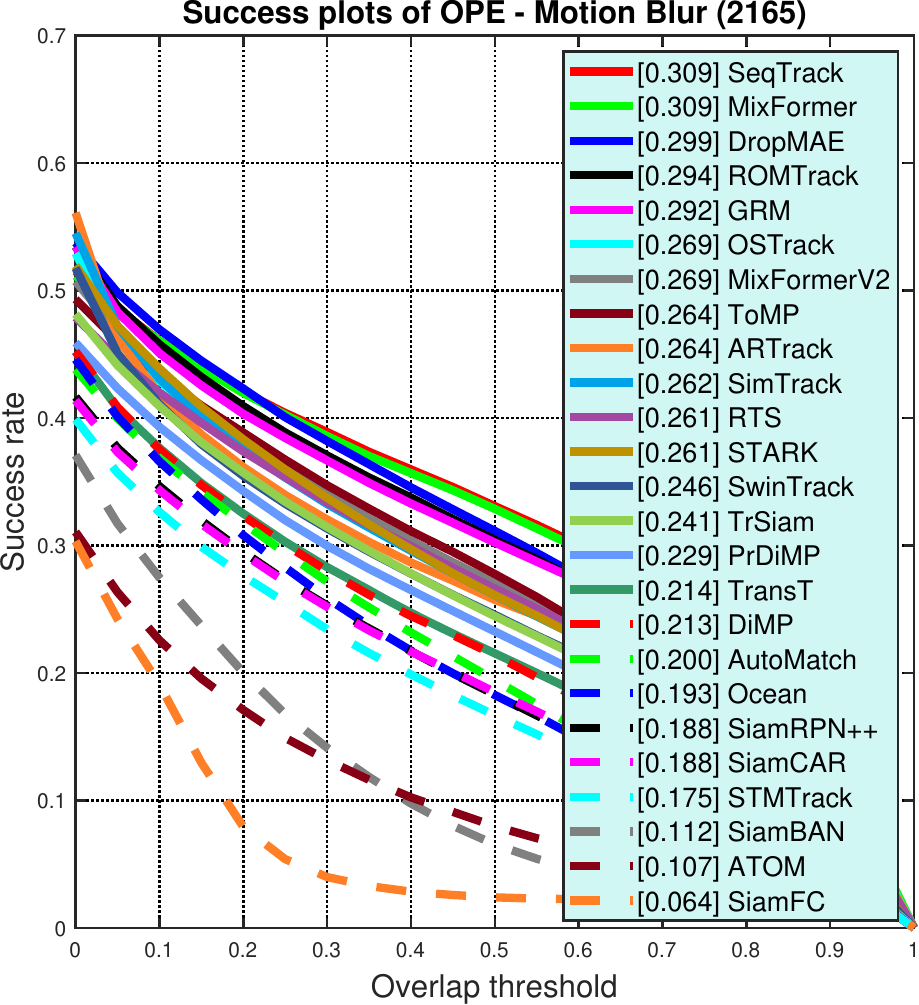}
	\includegraphics[width=0.19\linewidth]{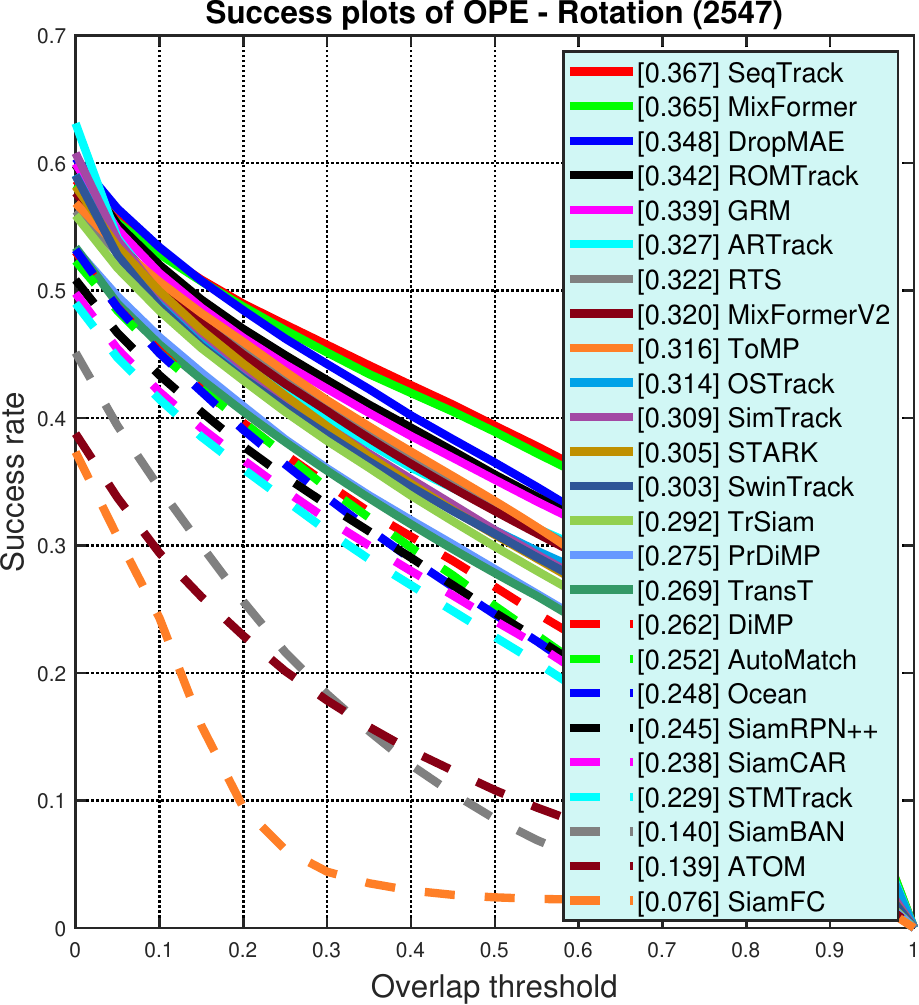}
        \includegraphics[width=0.19\linewidth]{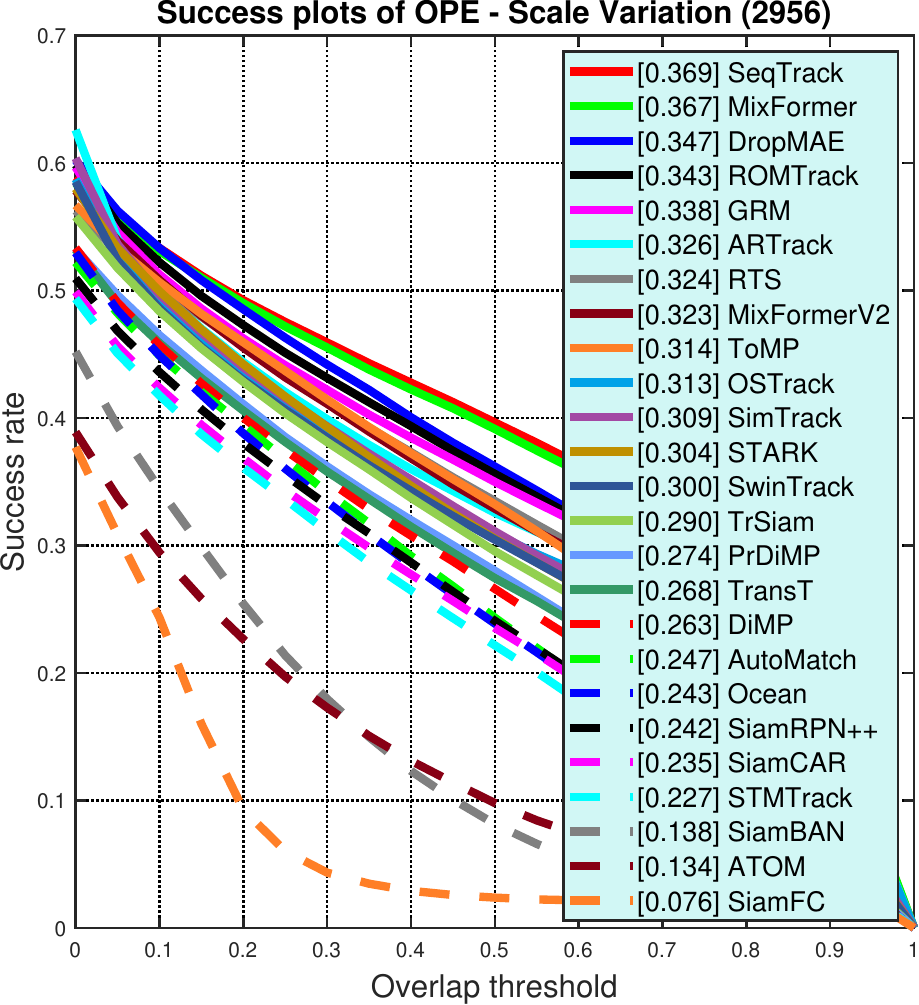}
	\caption{Performance of trackers on ten attributes using success (SUC).}
	\label{fig:full_att_suc}
\end{figure}

\subsection*{S3 \;  Full Results of Attribute-based Evaluation}

We present evaluation for all trackers under ten attributes, including f ARC, BC, DEF, FM, INV, IV, LR,
MB, ROT, and SV. Specifically, Fig.~\ref{fig:full_att_pre}, Fig.~\ref{fig:full_att_npre}, and Fig.~\ref{fig:full_att_suc} respectively show the attribute-based evaluations using PRE, NPRE, and SUC. From these evaluation results, we can observe that, existing state-of-the-art trackers heavily suffer from various challenges in the videos. To achieve general tracking, more efforts are needed to improve their robustness.

\subsection*{S4 \;  Additional Discussion}

We briefly discuss performance of rare and common classes (rare/common classes defined as those unseen/seen in LaSOT and TrackingNet). We conduct analysis based on average SUC (called avgSUC). We find that although rare classes are generally harder to track. For example, the avgSUC of \emph{Air Hockey} (rare) is 0.052 (other examples like \emph{Knife} etc), while avgSUC of \emph{Actor} (common) is 0.691. Yet, this is not absolute. For some rare class like \emph{Aardwolf}, its avgSUC is 0.689, higher than performance of some common classes. We argue that this is determined by a specific category and the environment where the target is located.

%
%
\bibliographystyle{splncs04}
\bibliography{main}
\end{document}